\documentclass[acmtog, screen, nonacm]{acmart} 

\usepackage{booktabs} 

\usepackage[ruled]{algorithm2e} 
\usepackage{multirow}
\usepackage{tkz-euclide}
\usepackage{subcaption}

\begin{document}
\title{360Roam: Real-Time Indoor Roaming Using Geometry-Aware ${360^\circ}$ Radiance Fields}

\author{Huajian Huang}
\email{hhuangbg@connect.ust.hk}
\author{Yingshu Chen}
\email{yingshu2008@gmail.com}
\author{Tianjia Zhang}
\email{tzhangbl@connect.ust.hk}
\author{Sai-Kit Yeung}
\email{saikit@ust.hk}
\affiliation{%
	\institution{Hong Kong University of Science and Technology}
	\department{Department of Computer Science and Engineering}
	\city{Hong Kong}
	\country{China}
}
\begin{teaserfigure}
	\centering
	\includegraphics[width=\linewidth]{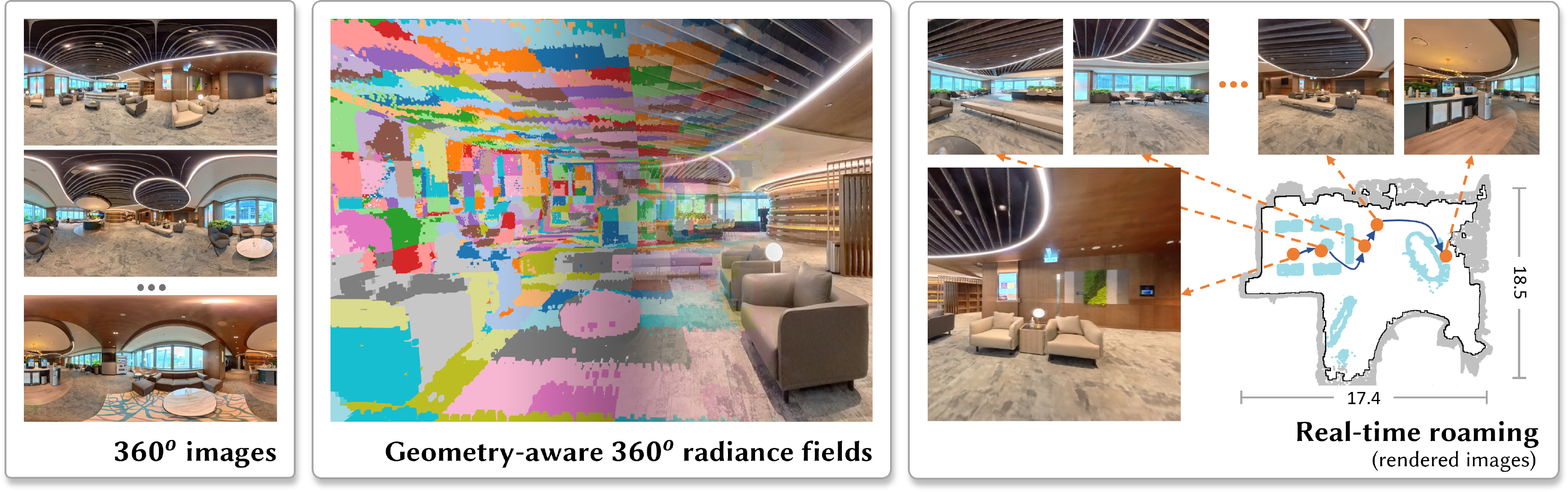}
	\caption{360Roam is an immersive 6-DoF roaming system designed for indoor scenarios. It utilizes $360^\circ$ images as input to acquire geometry-aware omnidirectional radiance fields, which comprise a multitude of adaptive-assigned neural perceptrons (represented by various colors in the middle figure). The system is capable of rendering real-time novel views on a single GTX1080 GPU. Furthermore, it incorporates a floorplan output of the observed scene to enhance immersive roaming experience.
	}\label{fig:teaser}
\end{teaserfigure}

\begin{abstract}

Virtual tour among sparse 360$^\circ$ images is widely used while hindering smooth and immersive roaming experiences.
The emergence of Neural Radiance Field (NeRF) has showcased significant progress in synthesizing novel views, unlocking the potential for immersive scene exploration. Nevertheless, previous NeRF works primarily focused on object-centric scenarios, resulting in noticeable performance degradation when applied to outward-facing and large-scale scenes due to limitations in scene parameterization.
To achieve seamless and real-time indoor roaming, we propose a novel approach using geometry-aware radiance fields with adaptively assigned local radiance fields. Initially, we employ multiple 360$^\circ$ images of an indoor scene to progressively reconstruct explicit geometry in the form of a probabilistic occupancy map, derived from a global omnidirectional radiance field. Subsequently, we assign local radiance fields through an adaptive divide-and-conquer strategy based on the recovered geometry. By incorporating geometry-aware sampling and decomposition of the global radiance field, our system effectively utilizes positional encoding and compact neural networks to enhance rendering quality and speed. Additionally, the extracted floorplan of the scene aids in providing visual guidance, contributing to a realistic roaming experience.
To demonstrate the effectiveness of our system, we curated a diverse dataset of 360$^\circ$ images encompassing various real-life scenes, on which we conducted extensive experiments. Quantitative and qualitative comparisons against baseline approaches illustrated the superior performance of our system in large-scale indoor scene roaming.

\end{abstract}

%
%

\ccsdesc[500]{Computing methodologies~Computational photography}
\ccsdesc[500]{Computing methodologies~Rendering}
\ccsdesc[500]{Computing methodologies~Virtual reality}

%

\keywords{Novel view synthesis, Neural rendering, Photorealistic imagery}

\maketitle

\section{Introduction}
Virtual tours of indoor scenes have become prevailing with advanced capturing devices, e.g., 360$^\circ$ cameras. But teleporting among discontinuous panoramic views cannot provide an immersive experience. Instead, we turn to neural rendering, i.e., the neural radiance field (NeRF)~\cite{mildenhall2020nerf} for continuous novel view synthesis learning from 360 images, and propose an immersive virtual roaming framework with real-time rendering performance in large-scale scenes. 
NeRF represents the structure and appearance of captured objects as an implicit neural network and utilizes multi-layer perceptrons to regress density and view-dependent color per ray. Through dense sampling along each ray, it can render photorealistic results even when encountering challenging rendering elements in real scenes such as view-dependent effects and transparent or semi-transparent objects. The follow-up NeRF-based methods~\cite{zhang2020nerf, liu2020neural,garbin2021fastnerf, barron2021mip, pumarola2021d,barron2022mip360, muller2022instant,chen2022tensorf} have made great efforts and modifications to increase rendering quality and speed. However, most of them focus on object-centric or small-scale scenes where the placements of cameras are constrained during the training and inference phases. Their performance degrades in large-scale scenes and the synthesized images become blurry and lack high-frequency information, due to insufficient representational capacity of scene parameterization.


In this work, we explore neural radiance fields for real-time roaming in complex indoor scenes. To achieve speedy rendering for immersive applications, we avoid increasing the computational complexity, e.g., utilizing a higher dimension of positional encoding and larger neural networks, 
Instead, we seek to fully exploit the capacity of positional encoding and neural networks by adaptively slimming the global radiance field. 
We propose to leverage geometry-aware radiance fields to increase rendering speed as well as better recover high-frequency details for fidelity rendering.
Accordingly, we propose 360Roam, a virtual roaming system rendering high-quality novel-view 360$^\circ$ images at real-time speed and allows immersive indoor roaming with 6-DoF motion (Fig.~\ref{fig:teaser}). 

\begin{figure*}
	\centering
	\includegraphics[width=\linewidth]{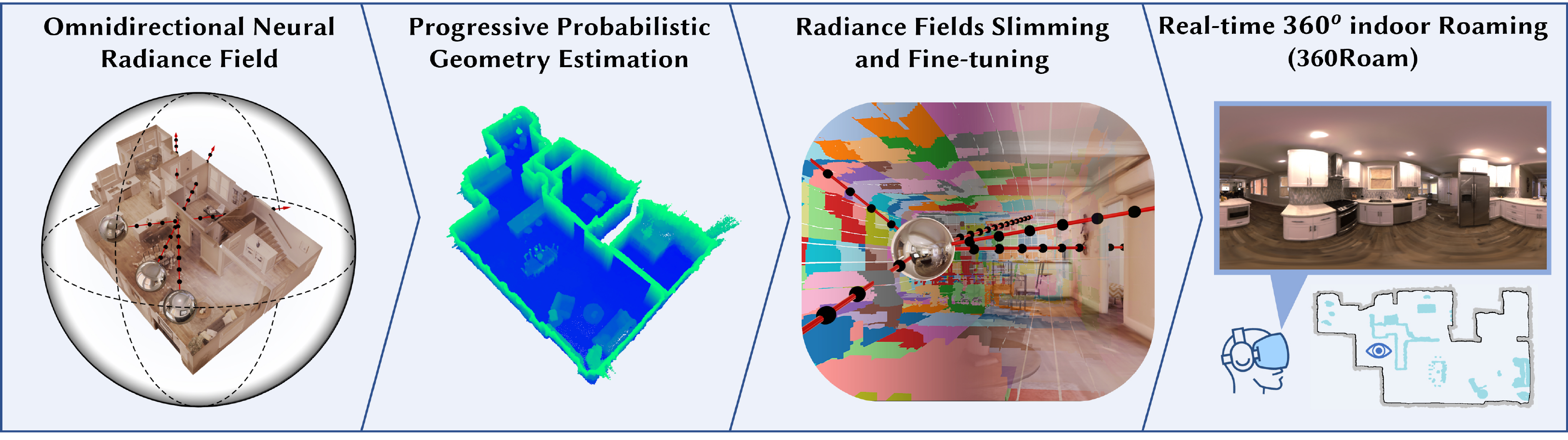}
	\caption{Overview of the proposed pipeline. First, an omnidirectional neural radiance field (360NeRF) will be learned from a set of input $360^\circ$ images (first figure). Then we sample 360NeRF at different camera positions to progressively estimate the probabilistic geometry and fuse them into an occupancy map (second figure) from which a floorplan with interior objects is extracted (floorplan inset in last figure).   
		The 360NeRF will then be slimmed and adaptively divided into multiple geometry-aware sub-fields, which improves both the rendering quality and speed significantly (last two figures).} 
	\label{fig:overview} 
\end{figure*}

360Roam consists of four stages following a coarse-to-fine process. 
First, we train a 360$^\circ$ neural radiance field (named 360NeRF) for the whole space to learn a global scene radiance field. Second, we use 360NeRF to generate panoramic depth and uncertainty of the scene simultaneously, and then progressively update the occupied probability. 
This gives us a reliable occupancy map while reducing the reliance on unreliable hyperparameters (e.g., predefined scene boundaries). The geometric information from the extracted occupancy map can effectively constrain the sampling range of the ray. Next, we slim one global neural radiance field into a number of sub-fields and fine-tune the radiance fields by skipping empty spaces and upsampling on occupied volume.
Finally, in our roaming system, we embed the recovered floorplan (with interior items) from the occupancy map.
The floorplan further enhances users' experience as a visual guide for users to circumvent item obstacles during scene roaming. Fig.~\ref{fig:overview} illustrates an overview of 360Roam. To summarize, our contributions include: 
\begin{itemize}
\item Introducing 360Roam, an effective system for high-fidelity and real-time 360$^\circ$ indoor roaming. It verifies the efficacy of the geometry-aware radiance fields for novel view synthesis in large-scale scenes.

\item Proposing a practical pipeline adapted for large-scale indoor scenes and reducing reliance on hyperparameters.

\item A new dataset and extensive benchmark of real-life indoor scenes with 360$^\circ$ image sequences covering a diverse set of indoor environments, which facilitates the future research of scene-level neural rendering. The dataset will be released to the public.
\end{itemize}

\section{Related Work}
\subsection{Neural Radiance Field and Important Variants}
NeRF~\cite{mildenhall2020nerf} is a promising and powerful neural scene representation for novel view synthesis. It uses the weight of a multi-layer perceptron to model the density and color of the scene as a function of continuous 5D coordinates and uses volume rendering to synthesize new views with no limitations on sampling rate.
Despite the outstanding visual performance, canonical NeRF is unable for real-time roaming since it suffers from fast training and rendering~\cite{piala2021terminerf, fridovich2022plenoxels, muller2022instant, chen2022tensorf, hedman2021baking, chen2023mobilenerf}, and large-scale scene reconstructions~\cite{zhang2022nerfusion, turki2022mega, xiangli2022bungeenerf}. Until now, many researchers have proposed valuable variants to address issues, significantly expanded the application range, and promoted robustness.

In our work, the most concerned issue is the efficiency of scene-level rendering to achieve real-time and immersive roaming. 
The pure implicit method equipped with volume rendering requires hundreds of network queries to render the final color of each pixel, making it hard to complete the computation in a short time. Some methods \cite{lombardi2021mixture, yu2021plenoctrees, xu2022point} exploit the sparse features of other representations such as mesh-based and primitive-based representations to improve efficiency. Other methods try to reduce the querying time and count. Neural Sparse Voxel Fields~\cite{liu2020neural} speeds up the rendering process using empty space skipping and early ray termination. DoNeRF~\cite{neff2021donerf} employs an extra network to predict sample locations directly and enforce those points closer to the intersection point of the surface and the camera ray. Instant-NGP~\cite{muller2022instant} reduces the training and rendering cost by embedding a multi-resolution hash table of learnable feature vectors into a small network that requires fewer floating-point operations. However, these methods are constrained to object-centric or relatively small scenes. In addition, KiloNeRF~\cite{reiser2021kilonerf} proves that using thousands of tiny independent MLPs to represent entire radiance fields can decrease the rendering time by three orders of magnitude. Their divide-and-conquer strategies have an underlying assumption that radiance fields are bounded by a box and its distribution is balanced. Therefore, they decompose the radiance fields into multiple volumes with the same pre-defined resolution. However, different from object-centric scenes, the structures of complex scenes are irregular with the imbalanced spatial placement of objects. Uniform decomposing will waste a lot of neural perceptrons in an empty space. Therefore, our method introduces an adaptive partition scheme to make use of the perceptrons' capacity.  Block-NeRF \cite{tancik2022block}, also decomposes the whole scene into several blocks and uses one individual NeRF to represent each block. The main difference is that their rendered image is composited from the rendering results of several block-NeRFs while we assign each block with a tinier NeRF network before the rendering process and synthesize the final image at once.
Current works~\cite{wu2022snisr, nerf2mesh, bakedsdf, mobilenerf} largely increase NeRF rendering speed by explicitly exporting reconstructed mesh, while they fail to represent the semi-transparent objects.


\subsection{VR Photography}

The omnidirectional image, which can cover 360$^\circ$ view, is one of the most typical types of VR photography and can thus provide a more comprehensive view of the environment. It allows sparser views to perform novel view synthesis on panoramas.
Some works have used RGBD panoramas~\cite{SerrMotion} including extra depth information obtained from other types of sensors for novel panorama synthesis. However, we encourage synthesis methods that are only supervised by visual images. \cite{huang20176} and \cite{SerrMotion} developed VR applications which support flexible viewing from 360$^\circ$ videos. Huang et al.~\cite{huang20176} utilized point clouds extracted from a 360$^\circ$ video to enable real-time video playback. Serrano et al.~\cite{SerrMotion} presented a layered representation for adding parallax and real-time playback of 360$^\circ$ videos. Recently, inspired by the MPI representation \cite{zhou2018stereo}, Lin et al. \cite{lin2020deep} and Attal et al. \cite{Attal:2020:ECCV} proposed multi-depth panorama (MDP) and multi-sphere image (MSI) representations, respectively, to conduct rendering from stereo 360$^\circ$ imagery. Previous works focused on processing captured 360$^\circ$ videos which cost lots of additional time to capture or post-process data. Broxton et al. \cite{broxton2020immersive} captured data using a customized low-cost hemispherical array made from 46 synchronized action sports cameras. The system can produce 6-DoF videos with a large baseline, fine resolution, wide field of view and high frame rates. The multi-sphere image (MSI) representation is used for light field video synthesis, yielding accurate results but involves a memory-heavy process. OmniPhotos~\cite{bertel2020omniphotos} reduced the capture time by attaching a 360$^\circ$ camera to a rotating selfie stick. It enables the quick and casual capture of high-quality 360$^\circ$ panoramas with motion parallax. The visual rendering quality is improved by automatically and robustly reconstructing a scene-adaptive proxy geometry that reduces vertical distortions during image-based view synthesis. However, user-centric VR methods do not scale for room- or scene-level roaming. 

\subsection{Panorama Structure Estimation}  
Methods for layout estimation are to either directly output the layout \cite{yang2019dula} or predict probability maps \cite{zou2018layoutnet,sun2019horizonnet, wang2021led} to parse the scenes. Some works \cite{zioulis2019spherical,zeng2020joint,jin2020geometric} jointly learn the layout and panoramic depth and perform panoramic view synthesis as a proxy task. 
However, these methods using a single ${360^\circ}$ image for simple layout estimation encounter limitations when dealing with multiple-room scenarios. Instead, our approach addresses this challenge by extracting a comprehensive floorplan from the estimated geometric occupancy map derived from the omnidirectional neural radiance field. This floorplan accurately aligns with the scene layout and also includes information about interior items.


\section{Preliminary}
\subsection{Neural Radiance Field (NeRF)}
NeRF~\cite{mildenhall2020nerf} represents the scene as a function of Cartesian coordinates, and outputs densities and emitted radiance colors at those querying locations. The function could be modeled by simple MLP networks. NeRF renders images via the volume rendering technique.
For any query 3D point $\boldsymbol{p}_j(x,y,z)$ in the ray, a global MLP model is applied to transform the concatenation of the point location $(x,y,z)$ and viewing direction (represented by a unit vector $\boldsymbol{d}$) into a 1-dimension density ($\sigma$) and a 3-dimension RGB color ($c$). Since the density is only dependent on point position, the model decomposes the mapping from only the location to the density, and from both the location and viewing direction the color. More specifically, the mapping is represented as a two-stage MLP. The first 8 fully-connected layers of the network, denoted as $MLP_{1:8}$, take as input the 3D position $\boldsymbol{p}_i(x,y,z)$, and output the density $\sigma$ and a 256-dimension radiance feature $f_c$. The combination of the feature $f_c$ and viewing direction $\boldsymbol{d}$ is processed by the final fully connected layer $MLP_{9:}$ and decoded into the radiance $\boldsymbol{c}_i$ in terms of RGB value. Specifically, it utilizes positional encoding \cite{vaswani2017attention} and maps the input coordinates into a high-dimensional vector, denoted by $\mathcal{F}(\cdot)$. The mapping is useful in modeling high-frequency functions. The MLP model can be represented as:
\begin{equation}
	\begin{aligned}
		&\sigma, f_c = MLP_{1:8}(\mathcal{F}(\boldsymbol{p}_i))\, \\
		&\boldsymbol{c}_i = MLP_{9:}(f_c, \mathcal{F}(\boldsymbol{d}))\, , 
		\label{MLP}
	\end{aligned} 
\end{equation}
where $\mathcal{F}(\cdot)$ is applied individually on each component of a vector.
NeRF jointly optimizes two identical networks of Eq.~\ref{MLP} with different parameters to increase rendering quality and avoid over-sampling in free spaces. The coarse network is evaluated on $N_c$ stratified sampling locations per ray. Based on the output densities of the coarse network, the coefficients represented as a function of density are normalized as weights to form a piecewise-constant PDF (probability density function). Additional $N_f$ points are sampled under this distribution. The fine network is then evaluated on all $N_c+N_f$ points along a ray, and output densities and colors for rendering the final color of the ray. 

\subsection{Omnidirectional Neural Radiance Field (360NeRF)}
The involved coordinate systems used in NeRF are all Cartesian. However, a panorama uses the panoramic pixel grid coordinate system, in which each pixel $(u,v)$  in the panoramic image corresponds to a point $(\phi, \theta)$ in a sphere surface represented by the spherical polar coordinate system. The transformation between the two coordinate systems is described by:
\begin{equation}
	\begin{aligned}
		&u = \phi * W /(2\pi) + W/2 ,\\
		&v = -\theta * H /\pi + H/2\, ,
	\end{aligned} 
\end{equation}
where $u,v$ denote the column and row of the panorama, $\phi, \theta$ denote longitude and latitude of the spherical surface, and $H,W$ denote the height and width of the panorama respectively. Additionally, the relation between the spherical and 3D Cartesian coordinate systems is:
\begin{equation}
	(x,\,y,\,z) = (\cos{\theta}\sin{\phi},\, -\sin{\theta},\, \cos{\theta}\cos{\phi}) \, .
\end{equation}
\noindent The camera ray cast from a panoramic pixel can be formulated as $\boldsymbol{r} = \boldsymbol{o} + r\boldsymbol{d}(\phi/f, \theta/f, 1)$, where $\boldsymbol{o} $ is the camera center, $f$ is the focal length, $\boldsymbol{d}$ is the ray direction and $r$ is the radial distance of a point in the ray. 
Instead of transforming the coordinate to a 3D Cartesian coordinate, we use an alternative parameterization to model the omnidirectional neural radiance field, referred to as 360NeRF.
360NeRF conducts sampling in inverse distance behavior and represents the points in the spherical coordinate system, $\boldsymbol{p}_i(\phi,\theta, r)$. Then 360NeRF queries an MLP (Eq. \ref{MLP}) for corresponding color and density. 


In the first step of the system, we query only one single MLP with $N$ samplings along the ray $\mathcal{R}_i$ instead of using a two-pass MLP query strategy.
Then, we accumulate the densities and radiance of the samplings to get the final color $\hat{C}(\mathcal{R}_i)$ and depth $\hat{D}(\mathcal{R}_i)$ estimation of the pixel, similar to the volume rendering formulated in NeRF.
Supposing the sampling position along the ray is $t_i$, the corresponding color and depth can be calculated by:
\begin{equation}
	\begin{aligned}
		&\hat{C}(\mathcal{R}_i) = \sum^{N}_{i=1}T_i\boldsymbol{c}_i \, , \\
		&\hat{D}(\mathcal{R}_i) = \sum^{N}_{i=1}T_it_i \, ,
	\end{aligned}
	\label{eq:volume_rendering}
\end{equation}
where the weight
\begin{equation}
	\begin{aligned}
        T_i=\exp{(-\sum^{i-1}_{k=1}\sigma_k(t_{k+1}-t_k))}(1-\exp{(-\sigma_i(t_{i+1}-t_i))}) \, .
	\end{aligned} 
	\label{eq:uncertain_weights}
\end{equation}
From experiments, we found our 360NeRF is well-suited for subsequent probabilistic geometry estimation in outward-facing scenes.

\begin{figure}[t]
\captionsetup[subfigure]{skip=0.5pt}
\centering
\subfloat[\small{Global density sampling}]{\label{fig:occupancy_global_frl_full}\includegraphics[width=0.9\linewidth]{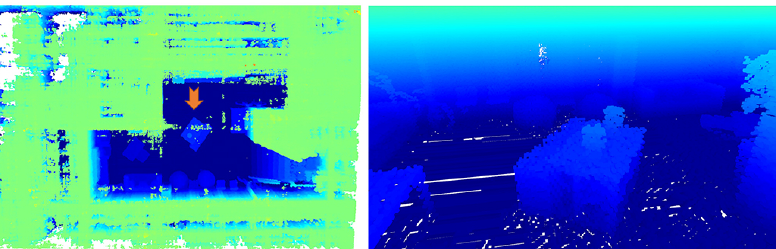}}\\
\vskip 0.1cm
\subfloat[\small{Our progressive estimation}]{\label{fig:occupancy_our_frl_full}\includegraphics[width=0.9\linewidth]{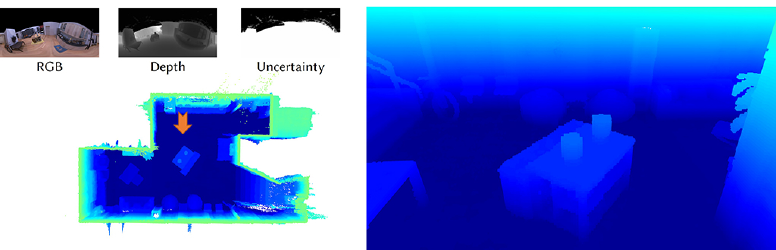}}\\
\vskip 0.1cm
\subfloat[\small{Intermediate results of our process}]{\label{fig:occupancy_our_frl_step}\includegraphics[width=\linewidth]{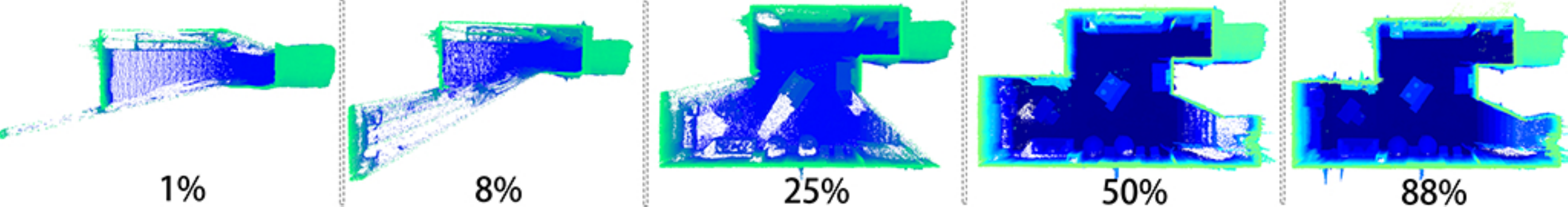}}
\caption{The comparison of occupancy maps reconstructed by different methods. In the first two rows, the left is the top-down view of the reconstructed model while the right is the perspective view at the position identified by the orange arrow.
Exiting method (a) relies on a global sampling of radiance fields within a prior bounding box which results in overfitting. We propose a progressive method to recover the occupancy map and achieve better performance (b), and the intermediate results are demonstrated in (c). }\label{fig:occupancy}
\end{figure}

\section{Geometry-aware Radiance Fields}
In structured scenes, characterized by numerous open spaces and spatially sparse object distributions, the density of radiance fields becomes highly imbalanced. Consequently, uniform sampling along the rays proves highly inefficient and often leads to being trapped in local minima. In contrast, geometry-aware sampling offers spatial efficiency and facilitates accelerated volume rendering.
We propose a progressive probabilistic method to recover geometric information in the form of an occupancy map from 360NeRF. Subsequently, leveraging the acquired occupancy map, our system adaptively adjusts and refines the radiance fields in a geometry-aware manner. This involves slimming down the model and fine-tuning it into multiple tiny local radiance fields.

\subsection{Progressive Probabilistic Geometry Estimation} \label{sec:geo_estimation_occ_map}

The existing approach for geometry extraction from NeRF, which relies on global density sampling \cite{reiser2021kilonerf}, has two limitations. Firstly, it assumes that the radiance field is enclosed within a cuboid and estimates the geometry based on the known boundaries. This assumption is generally impractical, particularly for outward-facing scenes. Although a structure-from-motion algorithm can provide a rough estimation of the bounding box for the radiance field, global density sampling remains sensitive to outliers and struggles to handle polyhedral scenes, resulting in the generation of redundant structures, as shown in Fig.~\ref{fig:occupancy_global_frl_full}. 
Consequently, this approach will lead to inefficient utilization of neural perceptrons, which poses a disadvantage during the subsequent decomposition and assignment of radiance fields (Sec.~\ref{sec:slimming}).
The second limitation lies in its heavy reliance on a prior assumption of density values (Eq.~\ref{MLP}), making it vulnerable when dealing with transparent objects and mirror reflections.

To extract accurate and robust geometric information, we introduce to recover \textbf{Probabilistic Occupancy Map} from depth map derived from depth values (Eq.~\ref{eq:volume_rendering}) and the according uncertainty map from normalizing weights (Eq.~\ref{eq:uncertain_weights}). We formulate occupancy recovery as a progressive probability estimation problem. Intuitively, the present state of each voxel depends on the current measurement and the previous estimation. The probabilistic model~\cite{yguel2008update} could be formulated in logit notation:
\begin{equation}
\begin{aligned}
    logit(p|z_{1:t}) = logit(p|z_{1:t-1}) + logit(p|z_{t}) \, ,
\end{aligned}
\label{eq:probabilistic}
\end{equation}
where $logit(p)=log\frac{p}{1-p}$ and $p$ is the probability of occupancy; $z_{1:t}$ represents the accumulated observation from the start to time $t$ and $z_t$ is the current observations at time $t$; 
With the incoming observations, it can progressively update the occupied probability of each voxel. The range of $logit(p)$ is from -2 and 3.5 which indicates the occupied probability of 0.12 and 0.97. When the probability is larger than the threshold 0.97, the voxel will be considered occupied. 

To facilitate these processes, we adopt a typical reconstruction framework Octomap \cite{hornung2013octomap} which utilizes an octree to store and maintain the occupied probability of the voxel. Based on predefined measurement uncertainty, Octomap only takes depth maps as input to generate occupancy voxels.
However, we can take advantage of 360NeRF to obtain a panoramic depth map (Eq.~\ref{eq:volume_rendering}) at each sampling location as well as the uncertainty estimation by normalizing the weights of Eq.~\ref{eq:uncertain_weights}. Therefore, we make use of the information to enhance occupancy map reconstruction (Fig.~\ref{fig:occupancy_our_frl_full}). The measurement probability will be changed according to the current 360NeRF estimation.
To reduce the use of hyperparameters, the resolution of the occupancy map depends on the ray sample rate of 360NeRF. 
Finally, we can rely on the pose of the training images to perform sampling. Since the training images are well-distributed throughout our scenes, sampling at each location of training images is typically sufficient to reconstruct an appropriate occupancy map of the scene. Please refer to Fig.~\ref{fig:occupancy_our_frl_step} as an illustrative example.
This progressive probabilistic geometry estimation process is outlined in the first stage of Algorithm~\ref{alg:one}. As depicted in Fig.~\ref{fig:occupancy}, this approach exhibits higher precision compared to global density sampling.

\begin{figure}[t]
    \captionsetup[subfigure]{skip=1pt}
    \def\imgw{0.45}
	\centering
	\subfloat[\small{NeRF}]{
        \label{fig:mlp_nerf}\includegraphics[width=\imgw\linewidth]{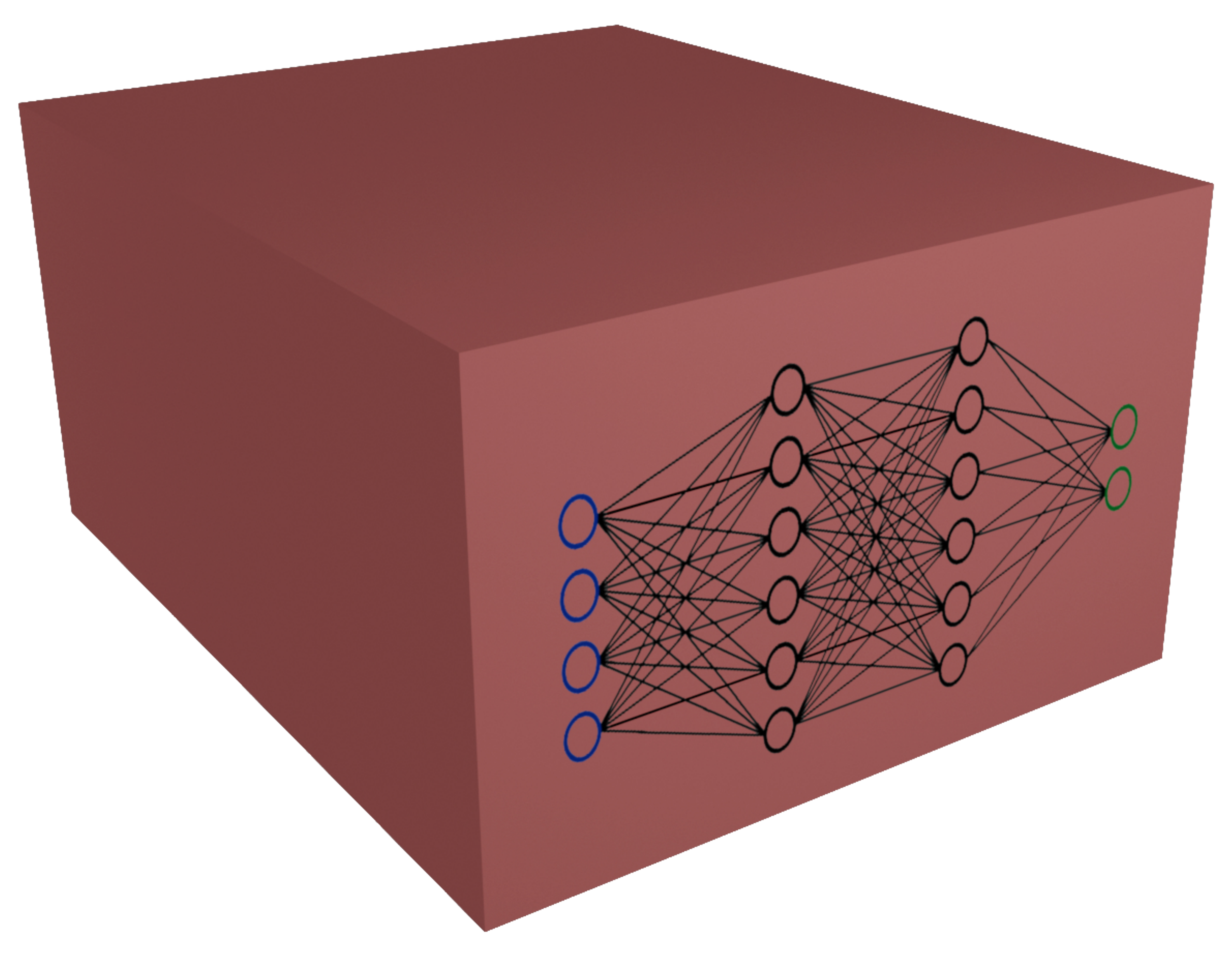}
    }\quad
    \subfloat[\small{KiloNeRF}]{
        \label{fig:mlp_kilonerf}\includegraphics[width=\imgw\linewidth]{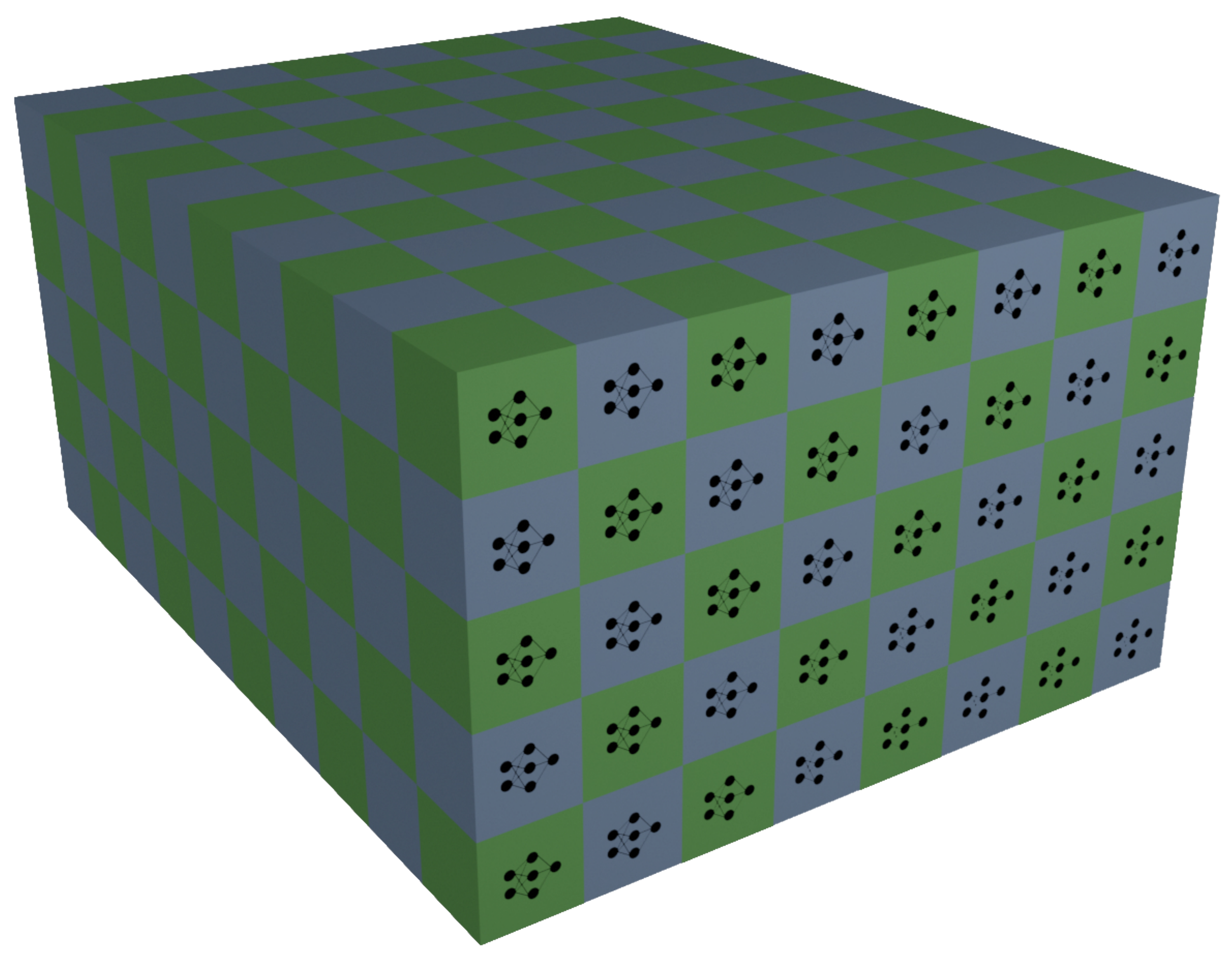}
    }\\
    \vskip 0.2cm
    \subfloat[\small{Ours}]{
        \label{fig:mlp_ours}\includegraphics[width=\imgw\linewidth]{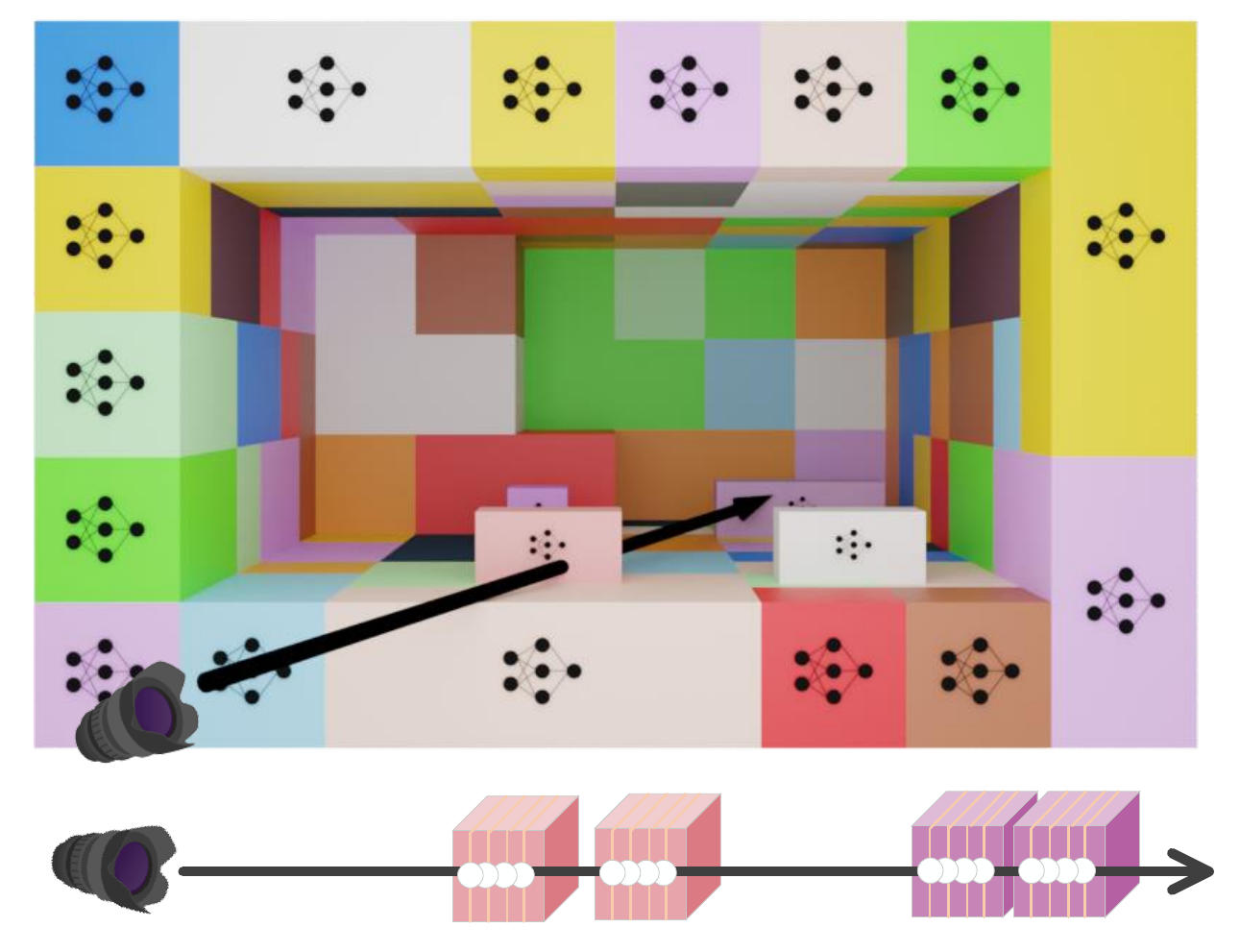}
    }\quad
    \subfloat[\small{The represented scene}]{
        \label{fig:mlp_scene}\includegraphics[width=\imgw\linewidth]{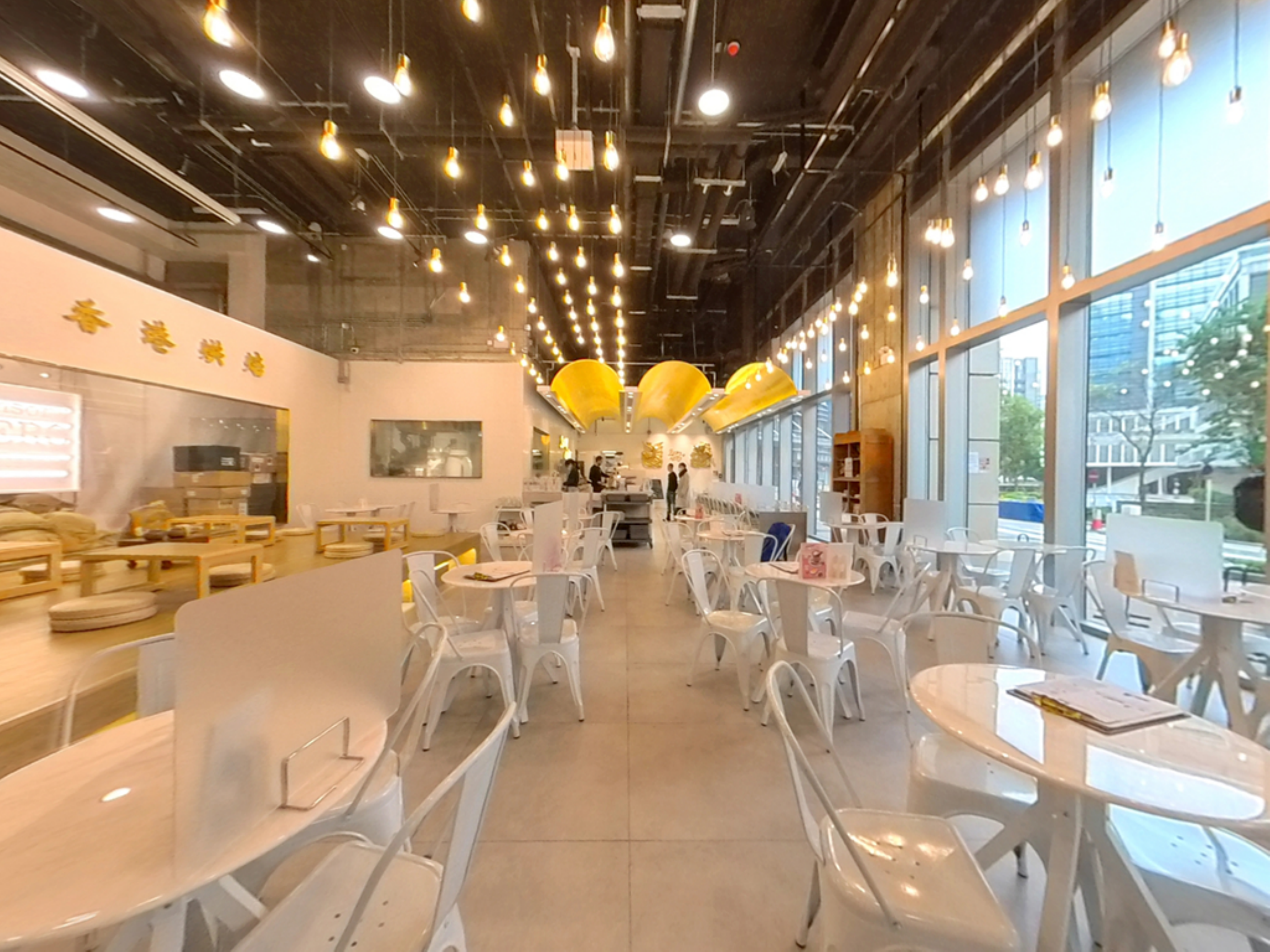}
    }
	\caption{Different representations of radiance fields. (a) NeRF uses a single deep MLP to represent the entire field. (b) KiloNeRF uniformly decomposes the field into thousands of tiny cubes (MLP) with fixed resolution and parallelly handles ray samplings for fast rendering. (c) By contrast, we consider the distribution of the scene and adaptively decompose the global radiance field. Each local sub-field has a similar amount of occupied voxels but a different size of volume. For illustration purposes here each sub-field corresponds to two occupied voxels.  (d) is the represented scene of (c).
    }
	\label{fig:mlp}
\end{figure}

\begin{algorithm}[t]
	\small
	\caption{\\ Probabilistic occupancy estimation \& networks assignment }\label{alg:one}
	{\bf Input:} 360NeRF and the amount of networks $n$\\
	/* Stage \uppercase\expandafter{\romannumeral1} (Sec.~\ref{sec:geo_estimation_occ_map})*/ \\
	Sample positions: $Q \gets$ training set\;
	\For{$q \in Q$}
	{
		distance map, uncertainty map $\gets$ 360NeRF ($q$)\;
		Updating occupied probability map using Eq.\ref{eq:probabilistic} \;
	}
	Obtained occupancy map $O$ and the amount of occupied voxel $S$\;
	/* Stage \uppercase\expandafter{\romannumeral2} (Sec.~\ref{sec:slimming})*/ \\
	Size of each sub-field $s \gets S/ n$ \;
	\For{$k\gets1$ \KwTo $n$}
	{
		Pop an unassigned voxel with the minimum coordinate from $O$ as the anchor\; Search $s-1$ nearest unassigned neighbors\;
		Assign these $s$ new voxels into the $k^{th}$ network\;
	}
\end{algorithm}

\subsection{Adaptive Radiance Fields Slimming and Fine-tuning}\label{sec:slimming}
Rendering an image from the NeRF for one pixel costs hundreds of network queries. In order to accelerate inference speed and reach real-time rendering, we can divide and decouple the entire neural radiance field and parallelly process each bundle of rays. However, in a large-scale scene, the distribution of objects and their corresponding radiance fields are highly imbalanced and the empty space does not contribute to the volume rendering. 
Therefore, contrary to the uniform divide-and-conquer strategy, our decomposition strategy is geometry-aware and takes scene geometry information into account.
Based on the extracted occupancy map (represented as occupied voxels), we decompose the entire field into $n$ sub-fields assigned with a similar amount of occupied voxels, as described in the second stage of Algorithm~\ref{alg:one}. Since the occupied voxels in 360NeRF are spatially imbalanced, each sub-field has a different size of volume (see networks covering regions in different colors in  Fig.~\ref{fig:mlp_ours}).
This step only requires one hyperparameter $n$, the number of tiny MLPs, which can be adjusted according to computational resources and demand in terms of rendering performance.  Fig.~\ref{fig:mlp} illustrates the different representations of radiance fields among vanilla NeRF, KiloNeRF and our geometry-aware radiance fields. 

To decouple each sub-field while pursuing global consistency, a fine-tuning process is conducted. The initial weights of tiny MLPs are distilled from the 360NeRF to increase the fine-tuning speed. Similar to distillation procedures~\cite{reiser2021kilonerf}, each MLP $\mathcal{M}_k$ will sample $I$ corresponding points in the 360NeRF to obtain the global domain radiance values $\boldsymbol{c}_i$ and densities $\sigma_i$. $\mathcal{M}_k$'s parameters are optimized via minimizing the mean squared errors between global and local values ($\boldsymbol{c}_i^k, \sigma_i^k$), which is formulated as:
\begin{equation}
\begin{aligned}
\mathcal{L}_k =\frac{1}{I} \sum_{i=1}^{I} \|\sigma_i^k-\sigma_i\|^2_2 + \|\boldsymbol{c}_i^k-\boldsymbol{c}_i\|^2_2
\end{aligned}\, .
\end{equation}
After initialization, the parameters of all tiny neural networks will be fine-tuned on the original training images with $L_2$ loss between rendered pixels and their ground truth. During the fine-tuning phase, it will densely sample the occupied voxels instead of uniform sampling along the rays. As radiance fields are slimmed and a vast amount of empty sampling is being avoided, the fine-tuning process can be converged quickly.

\begin{figure}[t]
    \def\imgw{1}
    \centering
    \includegraphics[width=\imgw\linewidth]{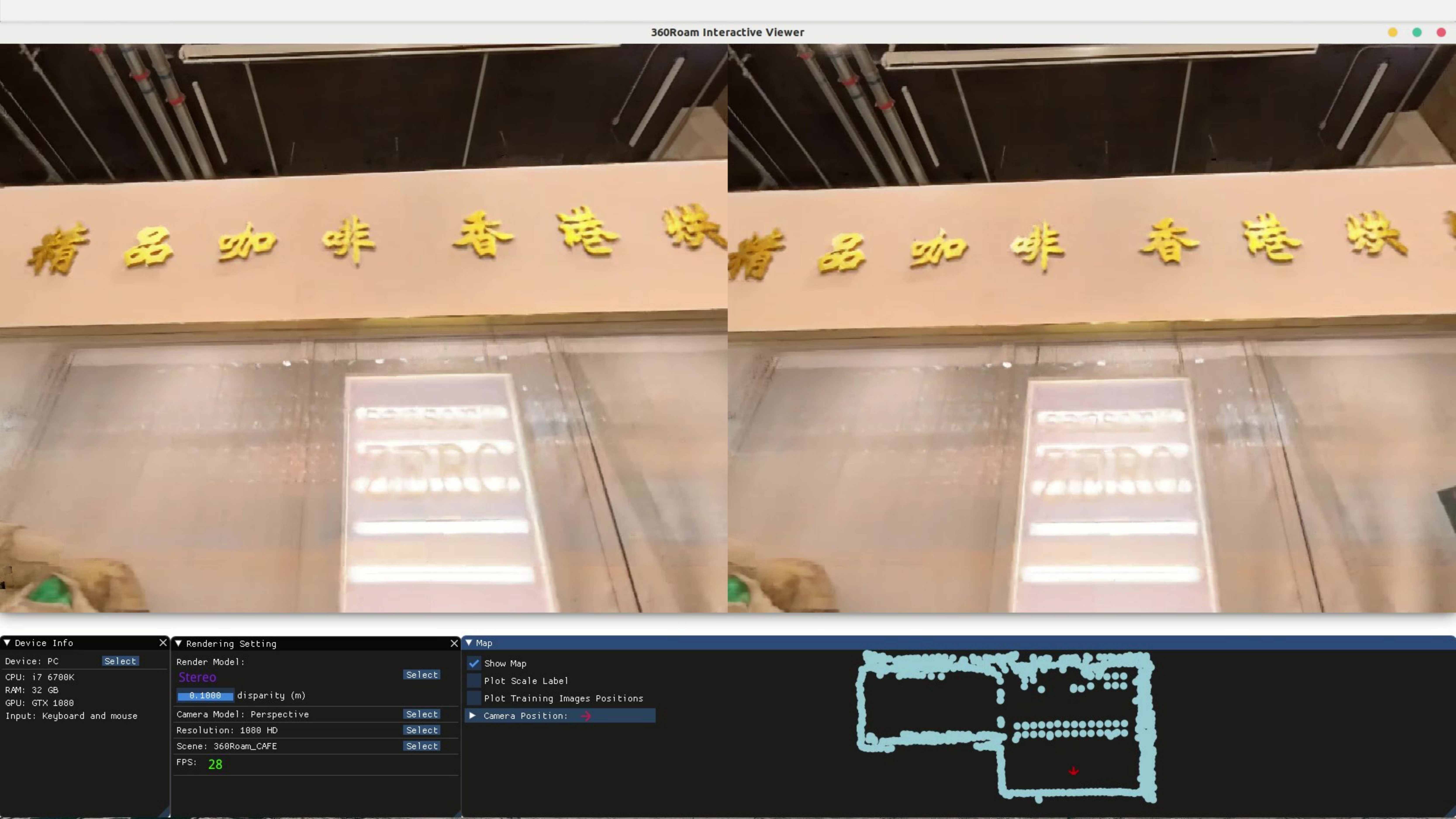}
    \caption{360Roam system in stereoscopic mode.}
    \label{fig:360roam_ui}
\end{figure}

\section{Real-time $360^\circ$ Indoor Roaming (360Roam)}
\subsubsection{Real-time Rendering Speedup}
To achieve real-time rendering performance for immersive $360^\circ$ roaming for common machines, we have taken advantage of ray casting techniques to quickly localize occupied voxels. After partitioning the radiance field as described in Sec.~\ref{sec:slimming}, each occupied voxel in the occupancy map has been labeled to the corresponding tiny MLP $\mathcal{M}_k$'s. 
In addition, we converted the occupancy map into a voxel hashing data structure and stored it in a GPU such that each ray can be concurrently processed. Due to the support of data structure, efficient empty space skipping, and parallel MLPs queries, our proposed system can run at real-time speed on a GTX1080 GPU in both monoscopic (Fig.~\ref{fig:overview} and \ref{fig:results}) and stereoscopic modes (Fig.~\ref{fig:360roam_ui}). 

\subsubsection{Floorplan Estimation and Visualization}
The estimated occupancy map (Sec.~\ref{sec:geo_estimation_occ_map}) provides useful geometric information to refine the radiance fields and we further exploit it to estimate the floorplan of the scene. The floorplan can be used to guide virtual roaming and restrict the walking path at inference time.

\begin{figure}[t]
	\centering

    \captionsetup[subfigure]{skip=1pt}
    \begin{subfigure}[b] {\linewidth}
        \centering
	    \includegraphics[width=1\linewidth]{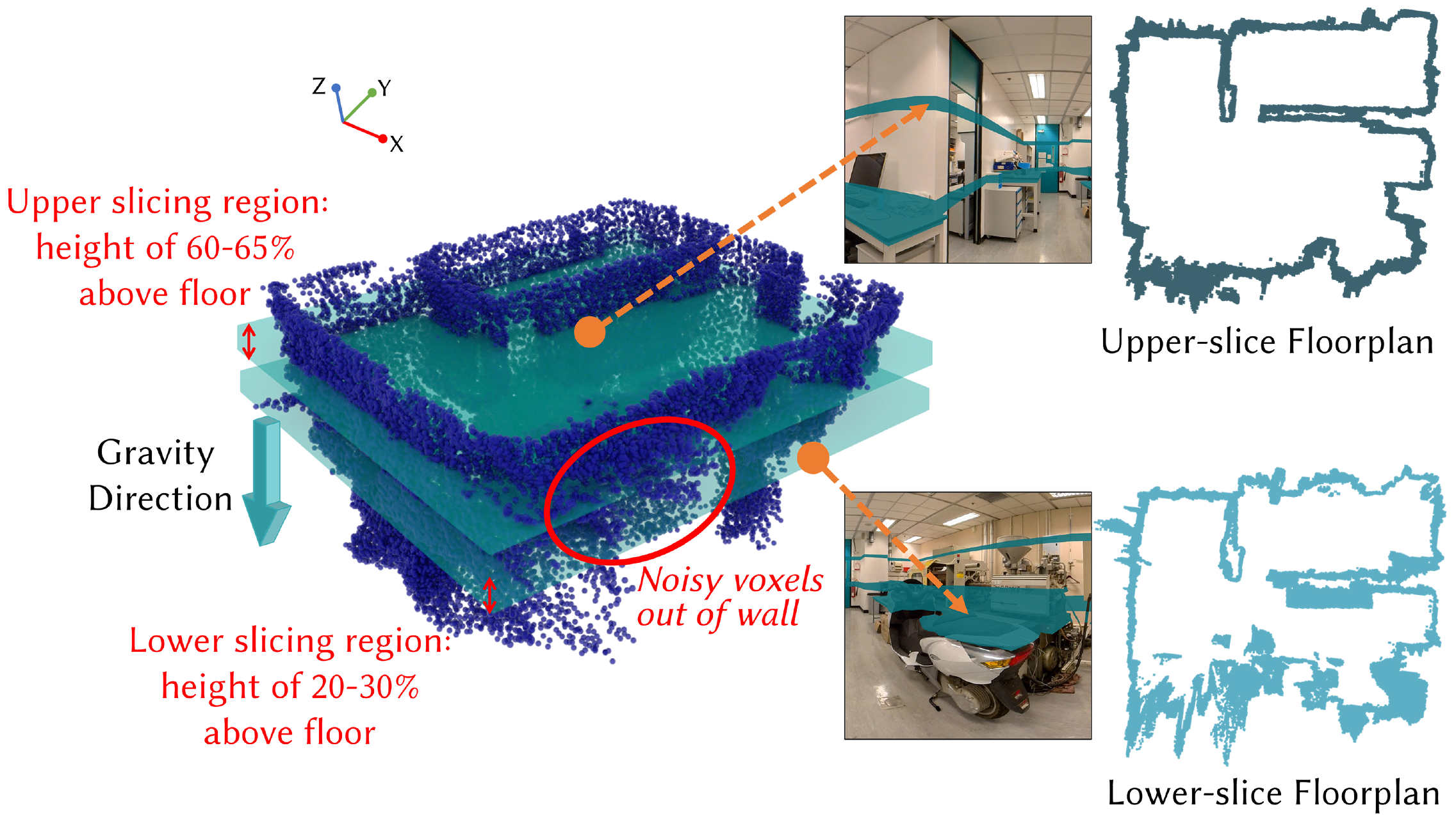}
	    \caption{\small Upper \& lower slicing}\label{fig:label_pipeline_slicing}
	\end{subfigure}
	\vskip 0.1em
	\begin{subfigure}[b] {0.45\linewidth}
        \centering
	    \includegraphics[width=0.7\linewidth]{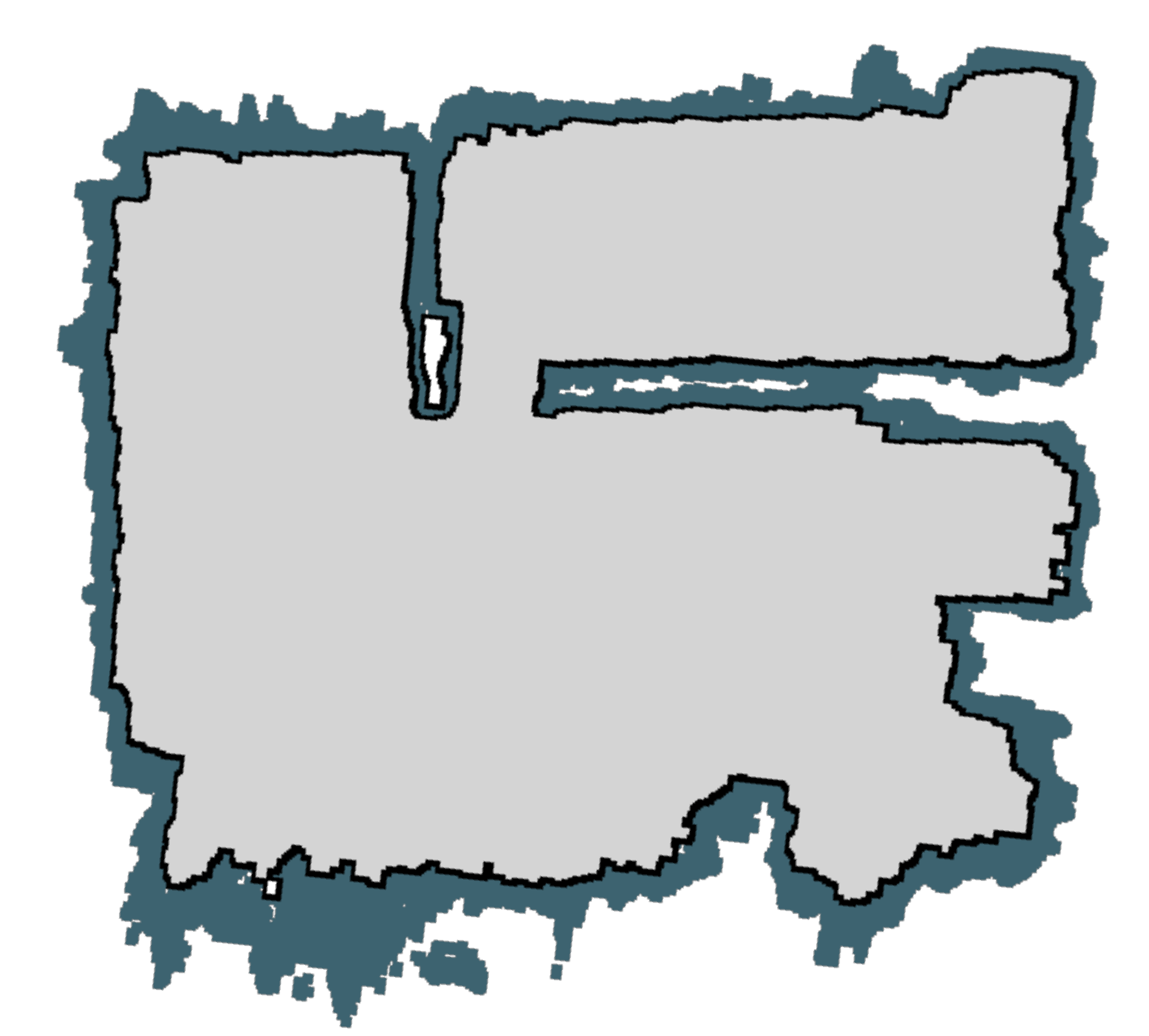}
	    \caption{\small Segmentation \& boundary} \label{fig:label_pipeline_segmentation}
	\end{subfigure} 
	\begin{subfigure}[b] {0.45\linewidth}
        \centering
	    \includegraphics[width=0.7\linewidth]{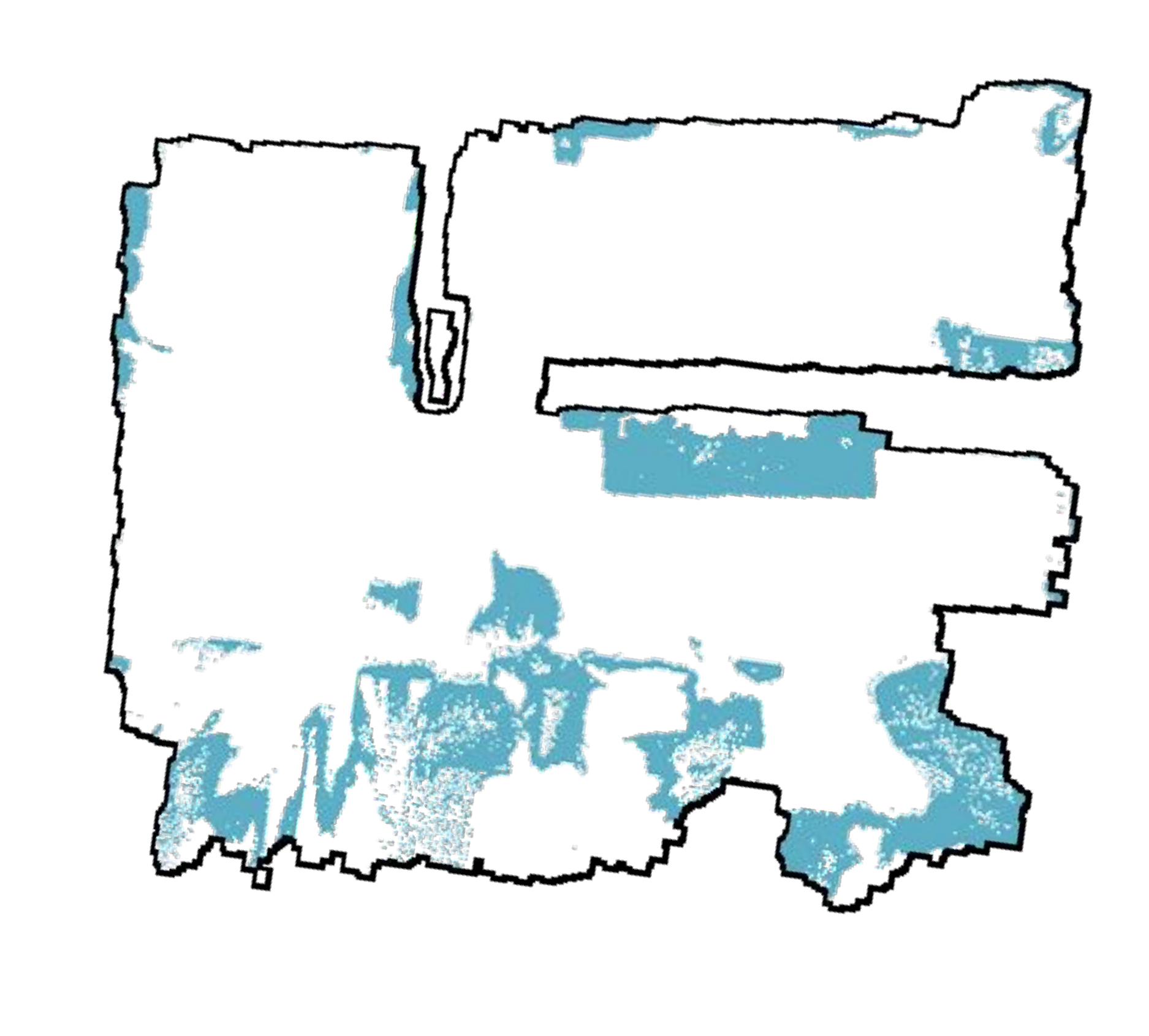}
	    \caption{\small Floorplan intersection}\label{fig:label_pipeline_intersection}
	\end{subfigure} 

	\caption{Floorplan estimation on scene \textsc{Lab}. (a) We collect occupied voxels at the height of upper and lower ranges and project them to floor-parallel plane as the initial floorplan. (b) We apply simple morphological operations on the upper-slice floorplan to extract the interior segment (gray area) and hence the floorplan's inner boundary (black line). (c) Intersecting the inner boundary from (b) with the lower-slice floorplan gives the final floorplan.}
	\label{fig:label_pipeline}
\end{figure}

The floorplan we estimate is essentially the inner surface of the wall and any objects in the scene that obstruct the navigation path. We project a range of occupied voxels in the upper slicing regions (60\% to 65\% of the height to avoid head jamb for multi-room contours) to a floor-parallel plane to form an initial boundary layout (see Fig.~\ref{fig:label_pipeline_slicing} upper-slice floorplan). 
To determine the floor-parallel slicing direction, we need to estimate the direction of the gravity of the reconstructed scene. Benefiting from 360$^\circ$ images, we get the gravity direction of the structured scene by calculating three vanishing points. After aligning the 360$^\circ$ images with gravity, we are able to estimate the height and extract the floor-parallel plane.
With the upper-slice floorplan layout, we apply simple morphological operations and segment out the inner boundary as shown in Fig.~\ref{fig:label_pipeline_segmentation}. 
Finally, to complete the floorplan with interior items such as chairs, tables which obstruct the indoor roaming path, we project occupied voxels in the lower slicing region (20\% to 30\% of the height that covers most of the obstructing objects) to the floor-parallel plane (see Fig.~\ref{fig:label_pipeline_slicing} lower-slice floorplan) and intersect the projected voxels with the inner boundary (see Fig.~\ref{fig:label_pipeline_intersection}).
In 360Roam, the intersected clean floorplan can enhance the user experience with a visualized scene boundary and item obstacles. For example, we visualized some floorplan samples in Fig.~\ref{fig:results}. Our system can support diverse shapes of indoor scenes which are not limited to a single room or cuboid combinations. 

\begin{figure*}[t]
\captionsetup[subfigure]{skip=0.15pt}
\def\imgw{0.194}
\centering
\subfloat[\small \textsc{Bar} \footnotesize (203 views)]{\label{fig:dataset_bar}\includegraphics[width=\imgw\linewidth]{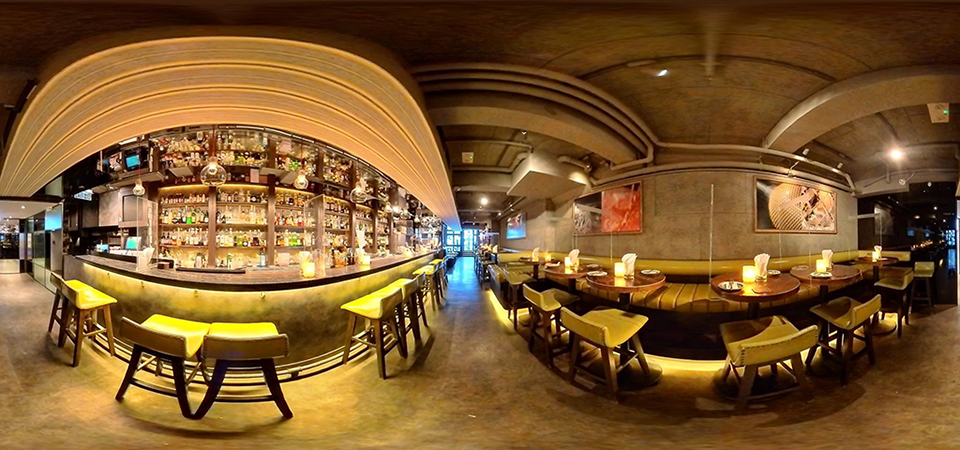}}\hfill 
\subfloat[\small \textsc{Base} \footnotesize (198 views)]{\label{fig:dataset_base}\includegraphics[width=\imgw\linewidth]{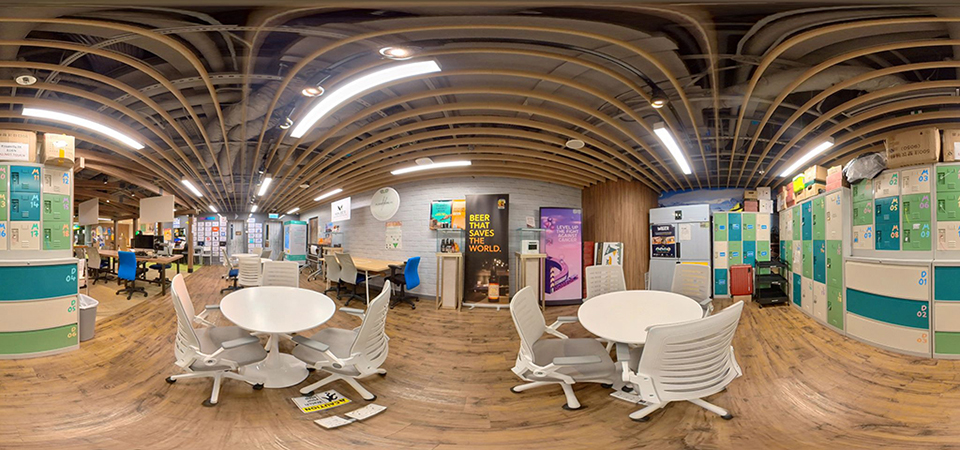}}\hfill
\subfloat[\small \textsc{Cafe} \footnotesize (100 views)]{\label{fig:dataset_cafe}\includegraphics[width=\imgw\linewidth]{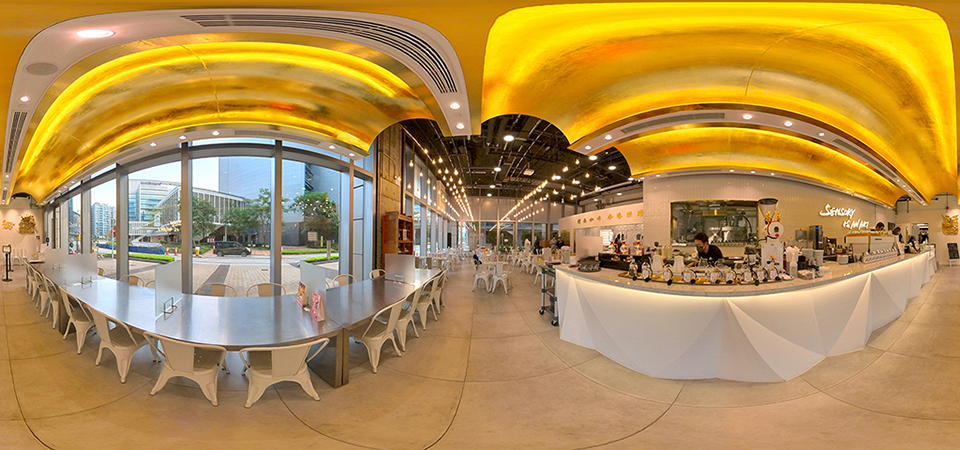}}\hfill
\subfloat[\small \textsc{Canteen} \footnotesize (85 views)]{\label{fig:dataset_canteen}\includegraphics[width=\imgw\linewidth]{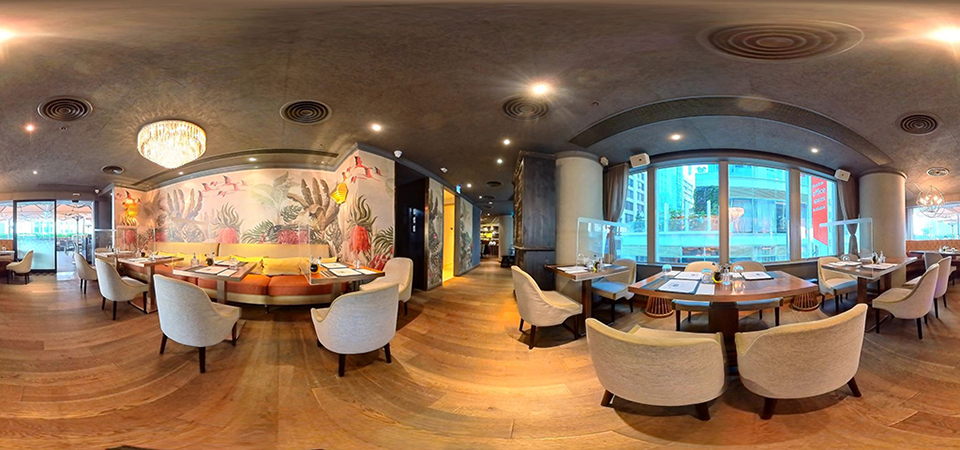}}\hfill
\subfloat[\small \textsc{Center} \footnotesize (127 views)]{\label{fig:dataset_center}\includegraphics[width=\imgw\linewidth]{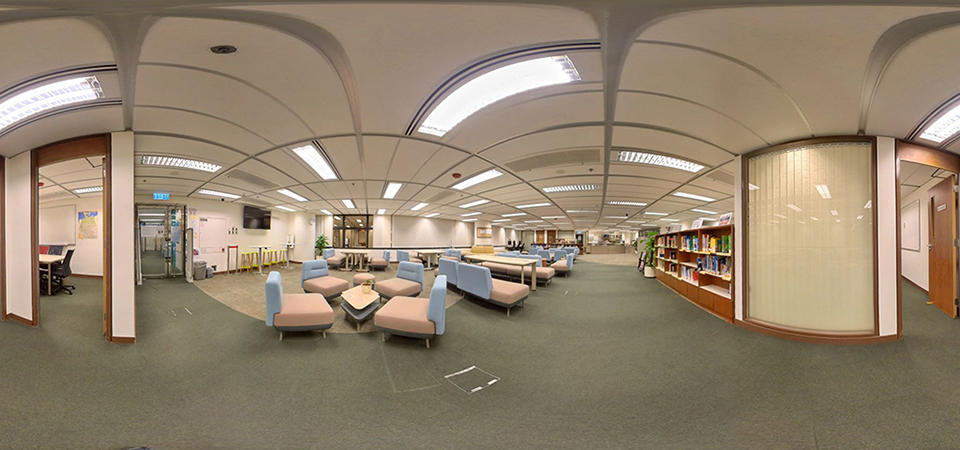}}\\

\subfloat[\small \textsc{Corridor} \footnotesize (71 views)]{\label{fig:dataset_corridor}\includegraphics[width=\imgw\linewidth]{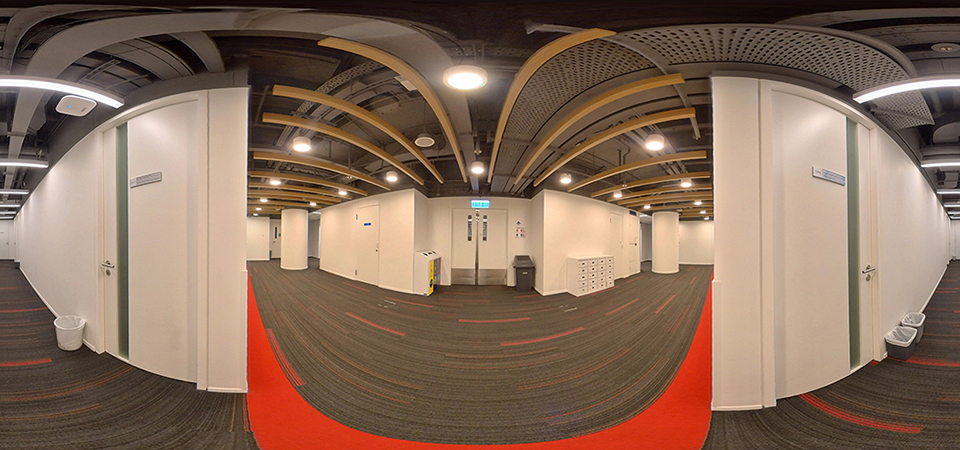}}\hfill
\subfloat[\small \textsc{Innovation}\footnotesize(215 views)]{\label{fig:dataset_inno}\includegraphics[width=\imgw\linewidth]{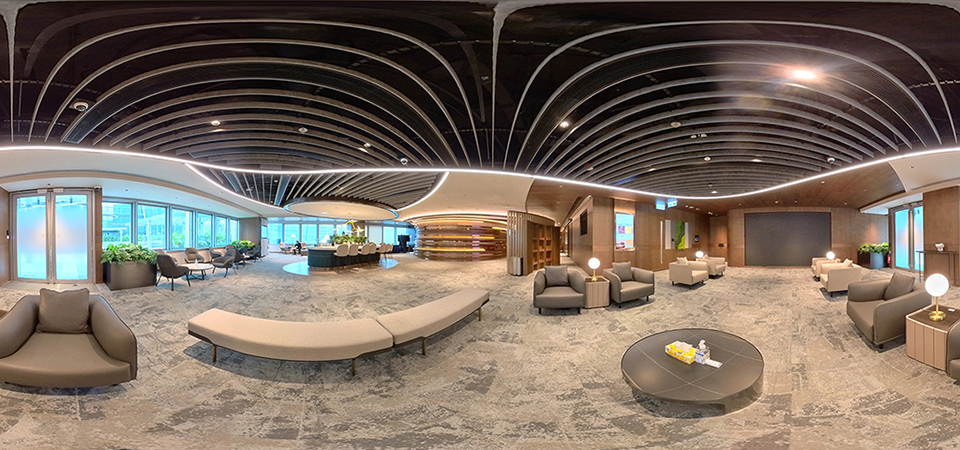}}\hfill
\subfloat[\small \textsc{Lab} \footnotesize (112 views)]{\label{fig:dataset_lab}\includegraphics[width=\imgw\linewidth]{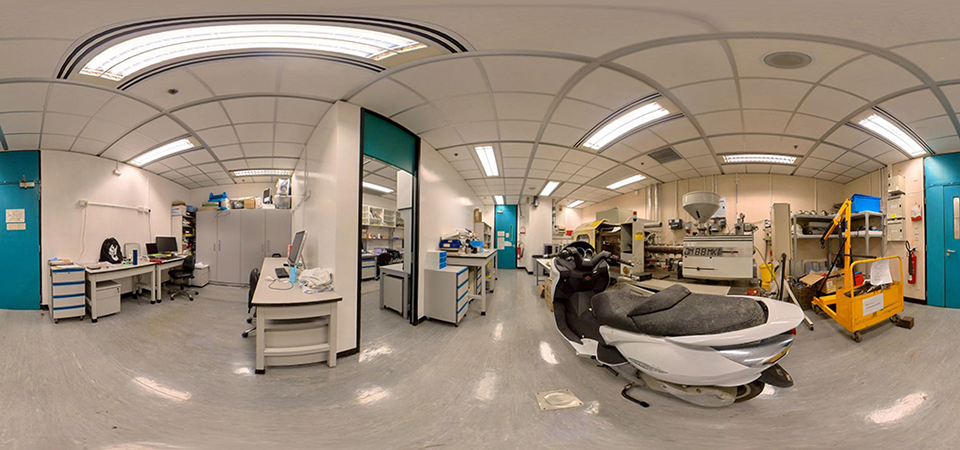}}\hfill
\subfloat[\small \textsc{Library} \footnotesize (89 views)]{\label{fig:dataset_library}\includegraphics[width=\imgw\linewidth]{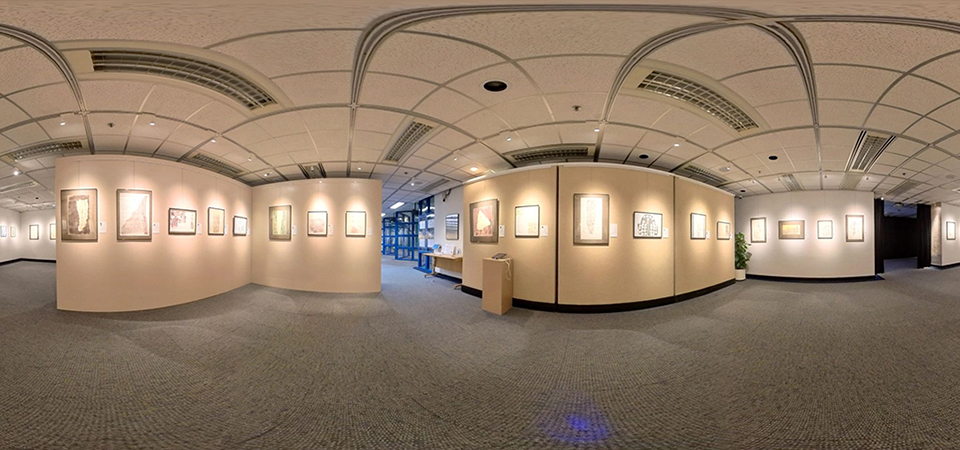}}\hfill
\subfloat[\small \textsc{Office} \footnotesize (141 views)]{\label{fig:dataset_office}\includegraphics[width=\imgw\linewidth]{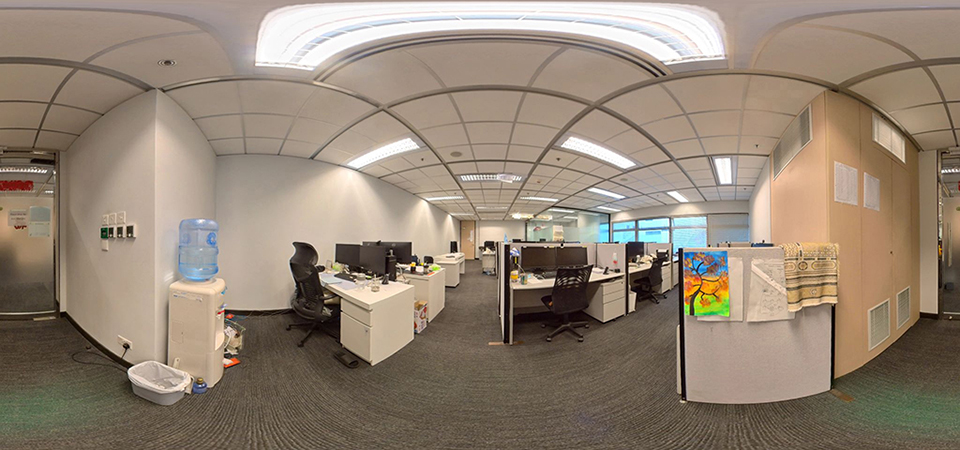}}\\

\caption{The 360Roam dataset contains 10 real-world indoor scenes with panoramic sequences captured using a $360^\circ$ camera.}
\label{fig:dataset}
\end{figure*}

\section{Results and Evaluation} 
In this section, we first detail the collection of panoramic real-world indoor scene data. 
Then, we conducted ablation studies to prove the validity of our proposed method.
Moreover, we evaluated the novel view synthesis performance for real large-scale scenes quantitatively and qualitatively. Finally, we validated the generalization of our method. The thorough evaluations show our system outperforms baseline methods and achieves the SOTA performance in terms of rendering efficiency and quality.  

\subsection{360Roam Dataset}\label{sec:dataset}
To evaluate our proposed method, we employed real-world data which are a collection of $360^\circ$ image sequences captured at various large-scale indoor scenes in the real world as Fig.~\ref{fig:dataset} shows. 
We used an Insta360 camera to capture real-world panoramas. We carefully selected 10 scenes named \textsc{Bar, Base, Cafe, Canteen, Center, Corridor, Innovation, Lab, Library} and \textsc{Office} respectively and picked 140 panoramas on average for each scene. They cover diverse indoor environments which have different styles, furniture arrangements, and layouts with multiple separate spaces. To facilitate data collection, we fixed the camera rig on a tracked mobile robot and controlled the robot remotely. In addition, we used the HDR capture model to preserve image quality. The resolution of the image is $6080\times 3040$. Finally, we used ~\cite{hhuang2022VO} to calculate the poses of $360^\circ$ images.
The keyframes extracted by the SFM pipeline are separated into two sets, one set serves as training data and the other serves as testing data at a ratio of 4:1. 
\begin{figure*}[t]
	\centering    
    
    
    \includegraphics[width=\linewidth]{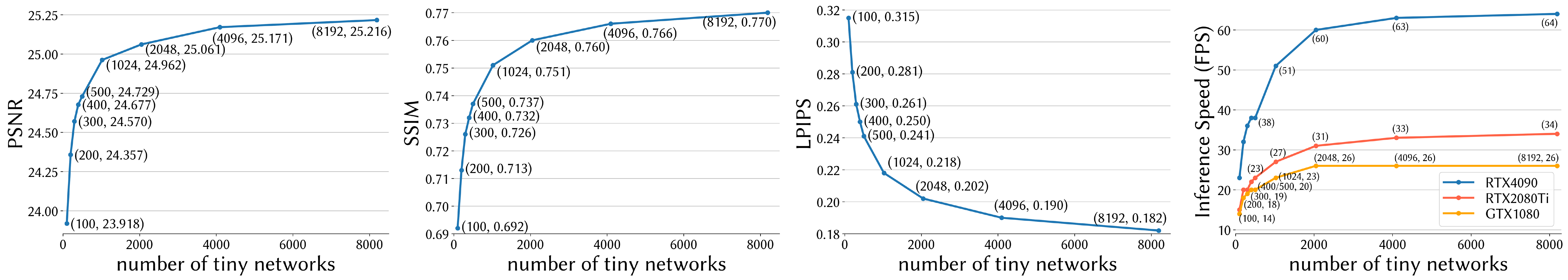}

	\caption{The influence of the number of the tiny networks.}
	\label{fig:network}
\end{figure*}

\begin{figure}[t]
    \def\imgw{0.44}
    \captionsetup[subfigure]{skip=0.8pt, justification=centering}
    
    \def\gaph{0.3em}
    \centering
    
    \subfloat[\small Rely on ground-truth geometry \\\footnotesize{$\left\langle PSNR=32.734, SSIM=0.925, LPIPS=0.032\right\rangle$}]{
    \label{fig:gtdepth}
    \includegraphics[width=\imgw\linewidth]{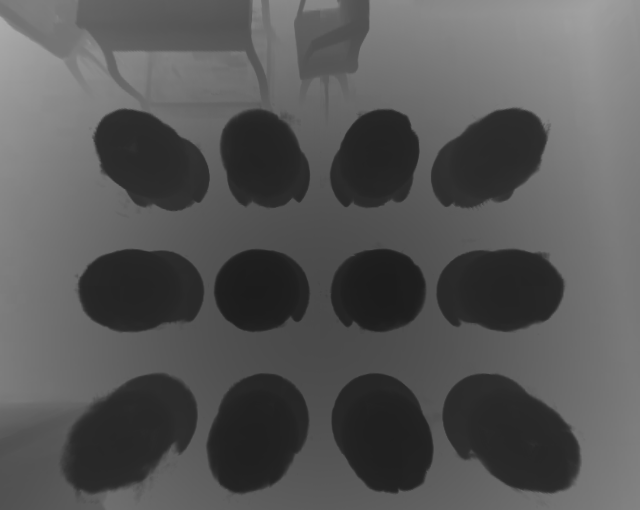}
    \includegraphics[width=\imgw\linewidth]{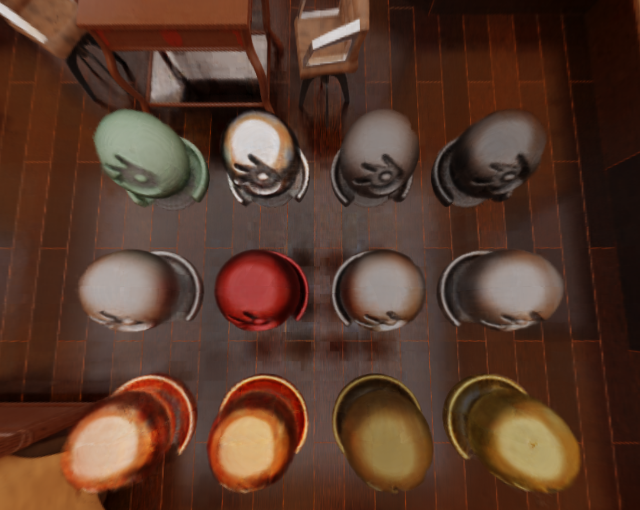}}\\
    \vspace{\gaph}
    \subfloat[\small Rely on geometry recovered by global sampling\\\footnotesize{$\left\langle PSNR=23.816, SSIM=0.880, LPIPS=0.132\right\rangle$}]{\label{fig:globalmap}
    \includegraphics[width=\imgw\linewidth]{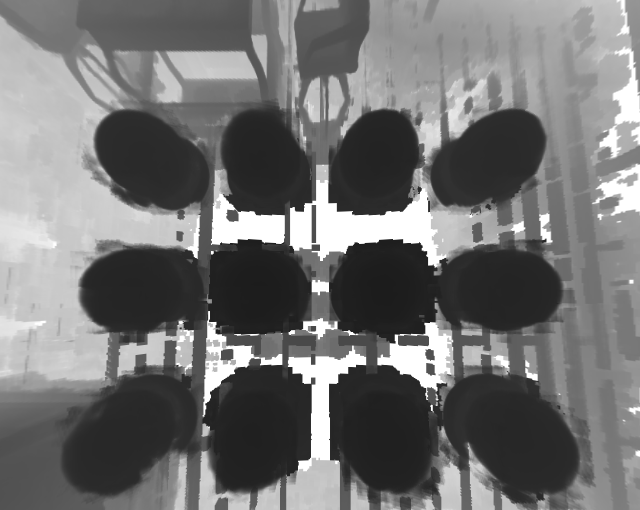}
    \includegraphics[width=\imgw\linewidth]{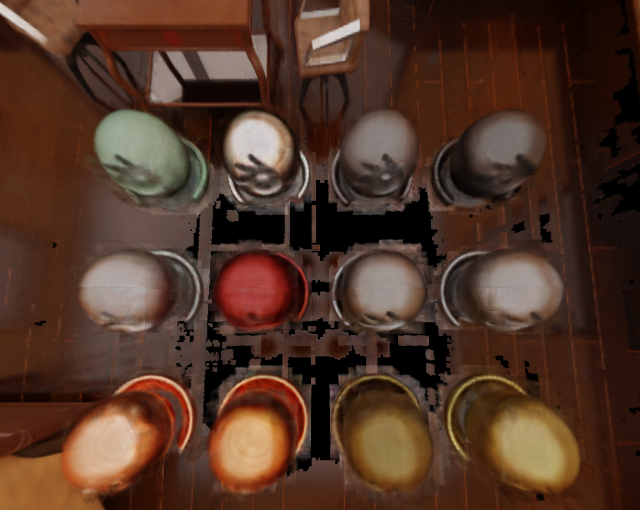}}\\
    \vspace{\gaph}
    \subfloat[\small Rely on probabilistic occupancy map\\\footnotesize{$\left\langle PSNR=32.660, SSIM=0.921, LPIPS=0.051\right\rangle$}]{\label{fig:progressivemap}
    \includegraphics[width=\imgw\linewidth]{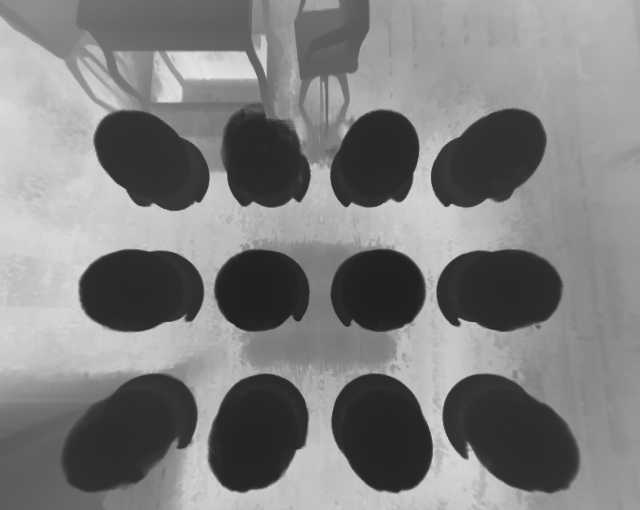}
    \includegraphics[width=\imgw\linewidth]{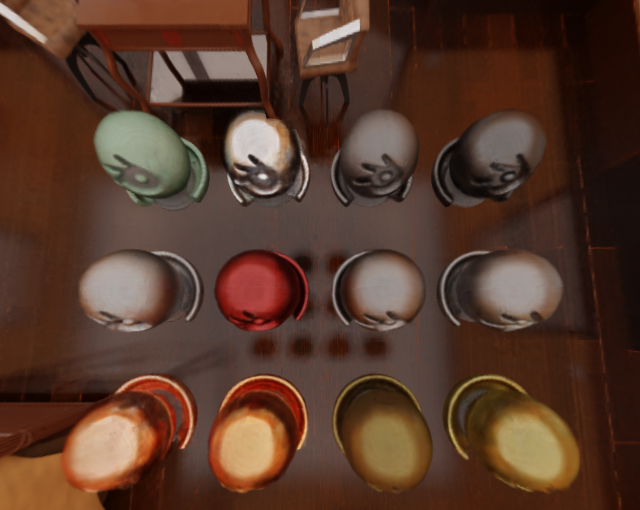}}
    
    \caption{Comparison of 360Roam relying on the geometry reconstructed by various methods. Left: estimated depth maps. Right: synthesized images. (a) The novel view is synthesized using ground-truth geometry. (b) The global sampling method fails to model the correct geometry and significantly impairs the rendering quality. (c) Our pipeline achieves comparable results with (a), showing the effectiveness of our progressive probabilistic geometry estimation.}
    \label{fig:geo_effect}
\end{figure}

\subsection{Ablation Studies}

\subsubsection{Influence of the Number of Tiny Networks} 
360Roam is composed of \textit{n} tiny MLPs, each with 4 fully-connected 32-channel layers. To analyze the effect and identify the necessary amount to achieve a balance between the rendering quality, speed and computational cost, we measured the rendering performance of 10 real scenes using different numbers of tiny networks $n$, including 100, 200, 300, 400, 500, 1024, 2048, 4096 and 8192. The evaluated results are illustrated as charts in Fig.~\ref{fig:network}, in terms of average PSNR, SSIM, LIPIPS and inference speed among all scenes. Particularly, inference speed was tested for images in 720p resolution on a GPU of GTX1080, RTX2080 Ti, and RTX4090, with 8GB, 11GB and 24GB memory respectively. By thousands of tiny networks, the more networks used, the better performance is. When the number exceeds two thousand, there is no obvious improvement.
In general, rendering speed relies on parallel computation, thus a larger number of networks can have a faster rendering speed until they reach the GPU capability of parallel computing. The memory and the number of tiny networks are linearly dependent, while 100 and 4096 tiny network models require $7.13MB$ and $582.42MB$ memory space, respectively. It is worth noting that the memory cost of 360Roam-100 using 100 tiny networks is similar to NeRF while the capacities of neural perceptrons are comparable in 360Roam-200 and NeRF, having about 1.2M parameters.
Overall, we used 2048 as a typical parameter to conduct the overall evaluations.

\subsubsection{Influence of Occupancy Map}
Different from vanilla NeRF or 360NeRF which models geometry implicitly, the complete 360Roam is a hybrid system where the ray sampling is based on the explicit geometry (illustrated in Fig.~\ref{fig:mlp_ours}) for increasing synthesis quality and speed. Our progressive probabilistic geometry estimation allows us to obtain a proper occupancy map fitting the scene rather than to get an exquisite 3D reconstruction. It is capable of modeling (semi-)transparent objects, as shown in the last row of Fig.~\ref{fig:results}. 
We evaluated the 360Roam performance when relying on the different occupancy maps reconstructed by different methods. Particularly, we rendered another set of 360$^\circ$ color and depth images of a synthetic scene with complex materials, and reconstructed a precise occupancy map as ground truth using ground-truth depth maps. 
In Fig.~\ref{fig:geo_effect} we can clearly see that the global sampling method cannot properly model the scene and the rendering quality decreases dramatically while using the progressively recovered geometry is competitive with the system using ground-truth geometry. Although the depth estimation of the reflective area is of low accuracy, 360Roam is able to render satisfying images, as demonstrated in Fig.~\ref{fig:progressivemap}.


\begin{figure}[t]
    \centering
    \includegraphics[width=\linewidth]{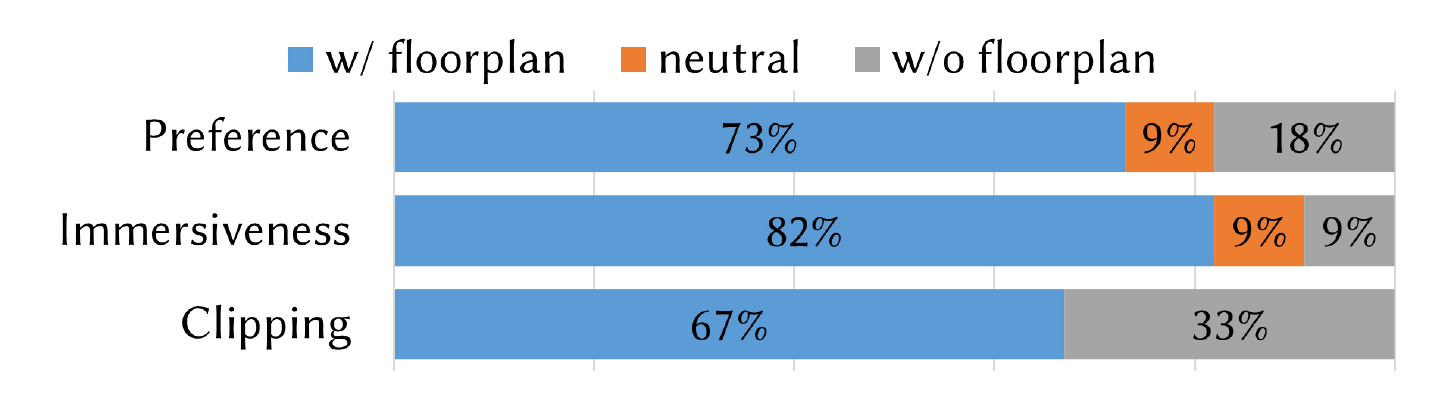}
    \caption{User study results on roaming system.}
    \label{fig:floorplan_user_study}
\end{figure}
\subsubsection{Influence of Floorplan}
The estimated floorplan with interior items can provide users with visual guidance for indoor roaming and avoiding occlusions. 
We conducted a user study to evaluate the effectiveness of the visual floorplan guide when roaming with perspective view display. Participants finished the scene roaming either with or without floorplan guidance, and then took a user experience survey. We collected user preference for using floorplan, immersiveness of using floorplan, and recorded the occurrence of camera clipping through wall/object. The average numbers per scene are reported in Fig. \ref{fig:floorplan_user_study}. 73\% of participants prefer using floorplan, some subjects dislike to use floorplan and can concentrate on roaming, and the remains keep neutral to floorplan usage. The majority agree with the immersiveness enhancement from floorplan. From observation, clipping occurrences decreased somewhat with floorplan guidance and roadblocking, even though 33\% of participants still avoided clipping with scene perception.

\subsection{Comparison on 360Roam Dataset}
This section validates the efficiency and rendering quality of 360Roam for large-scale real indoor scenes given panoramic multi-views, by comparing novel view synthesis performance of recent NeRF-based approaches.


\subsubsection{Baselines}
To evaluate our model on 360Roam Dataset, we compared it to the NSVF~\cite{liu2020neural}, canonical NeRF~\cite{mildenhall2020nerf}, KiloNeRF~\cite{reiser2021kilonerf}, Mip-NeRF360~\cite{barron2022mip360}, and recent fast training instant-NGP~\cite{muller2022instant} and TensoRF~\cite{chen2022tensorf}. 
NSVF aforehand defines a set of voxel-bounded implicit fields organized in a voxel octree to model local properties in each cell and progressively learns the underlying voxel structures during training which achieves good performance in object-centric scenarios. 
KiloNeRF dramatically improves inference speed by introducing a uniform divide-and-conquer strategy while maintaining similar rendering quality to vanilla NeRF.
Mip-NeRF360 \cite{barron2022mip360} is the state-of-the-art method in terms of rendering quality.
Instant-NGP \cite{muller2022instant} applies multiresolution hash encoding for grid-based representation with a large memory cost, while 
TensoRF \cite{chen2022tensorf} factorizes radiance fields into compact low-rank tensor components reducing memory.

\subsubsection{Training Details and Parameters} 
As the training of NeRF and most NeRF-based methods are commonly time-consuming, we resized images into $1520 \times 760$ for training and evaluation. 
All the experiments were trained on RTX2080 Ti GPUs. We used default training configurations for all baselines. 
For NeRF, it sampled 64 points per ray for the coarse query. It then sampled another 128 points and used all 192 points for the fine query. Therefore, it took 256 network queries in total to render a pixel. Mip-NeRF360 was evaluated using two configurations using either common (same as canonical NeRF) 256-channel MLP or 1024-channel MLP network. KiloNeRF requires bounding boxes of scenes and fixed voxel resolutions for training. These prerequisite parameters were estimated from the occupancy maps reconstructed by our methods. 
NSVF is extremely GPU-memory hungry, and the demand on memory would continuously increase as it iteratively subdivides and optimizes voxels during training. Following the default parameters configuration, all the models of NSVF were trained in a multi-process distributed manner using 4 NVIDIA RTX 3090 GPUs with 24 GB memory. NSVF would stop training once out-of-memory error occurs.

Our 360Roam applied the Adam optimizer with a learning rate that begins at 1e-3 and decays exponentially to 5e-5 during optimization. The hyperparameters of the optimizer are set to the default values with $\beta_1=0.9,\beta_2=0.99$. For every backward stage, we used a batch of 4096 rays per process. The coarse training of a 360NeRF takes about 15 hours on 4 RTX2080 Ti GPUs while the fine-tuning of thousands of tiny networks generally takes 4 hours after slimming.

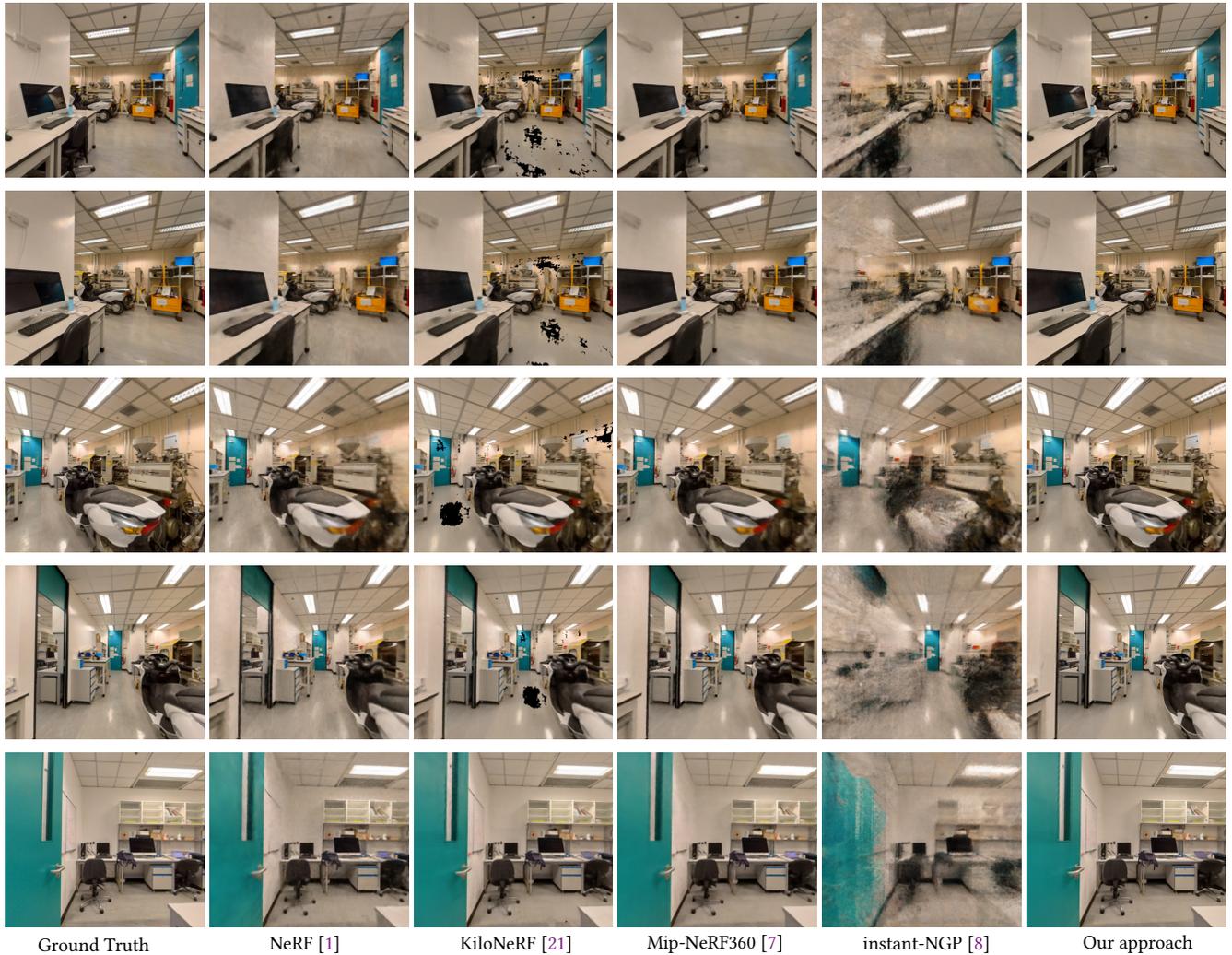
\begin{figure*}
    \def\imgw{0.16}
    \def\gaph{0.5em}
	\centering
	\includegraphics[width=\imgw\linewidth]{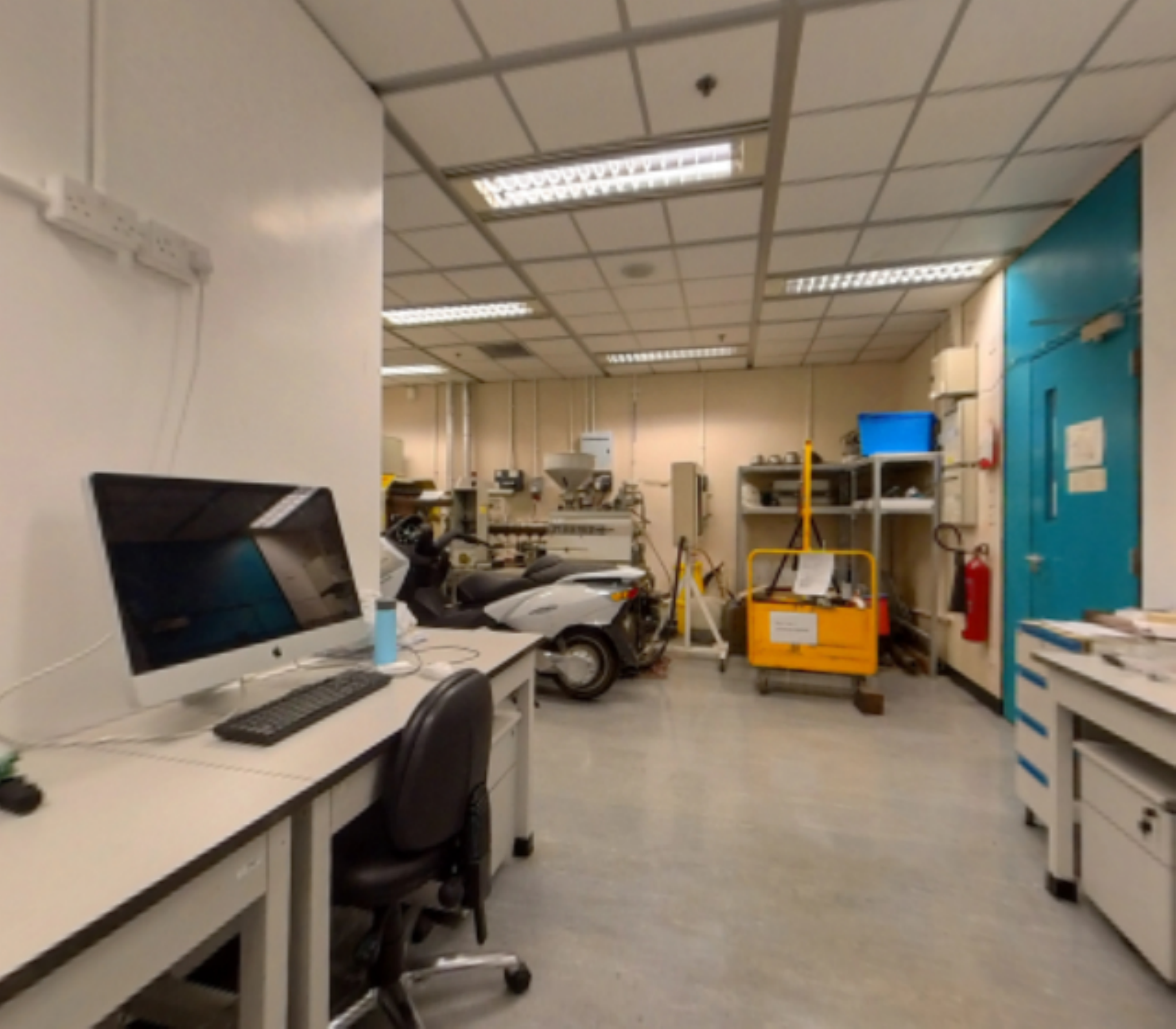}
	\includegraphics[width=\imgw\linewidth]{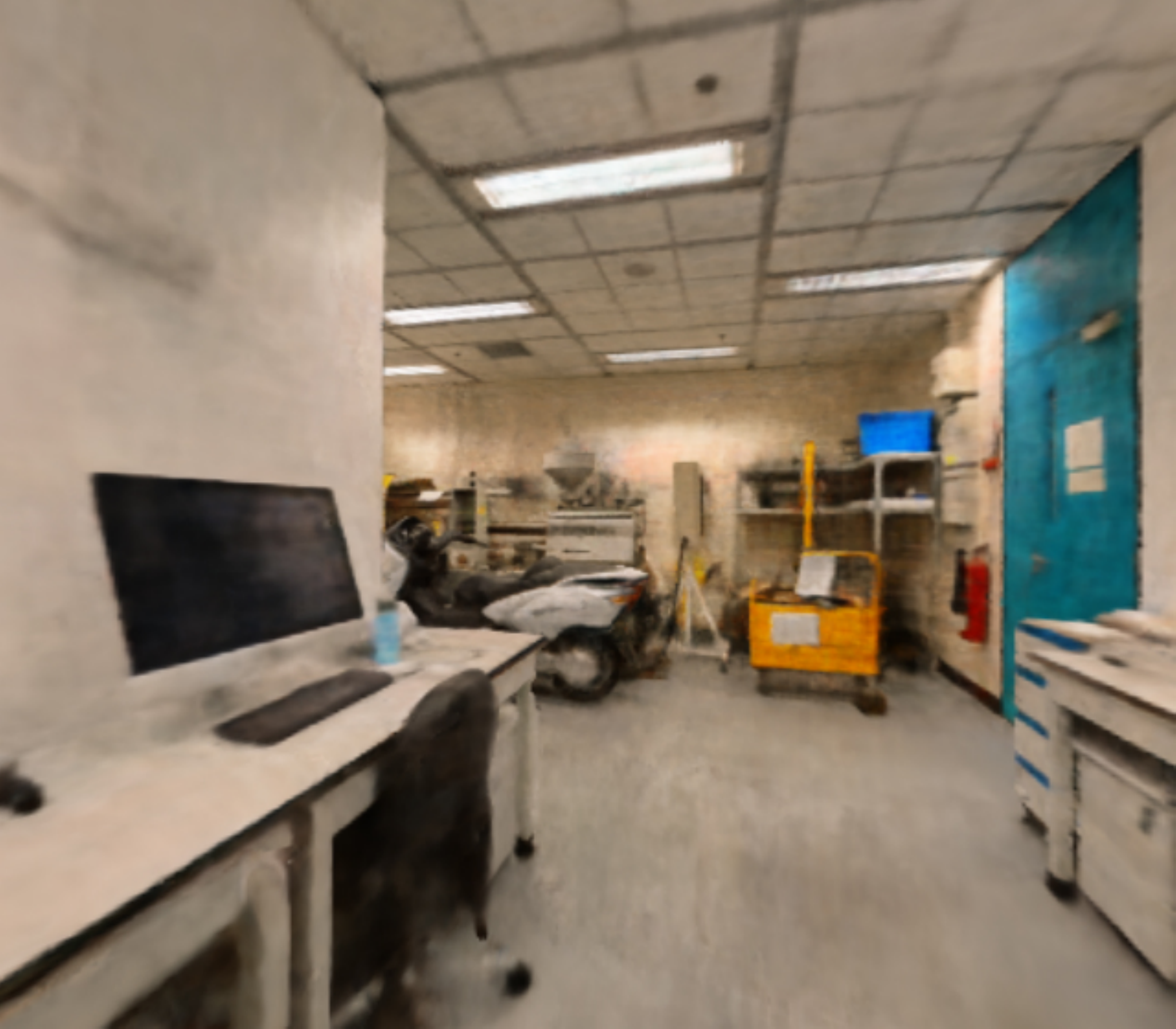}
	\includegraphics[width=\imgw\linewidth]{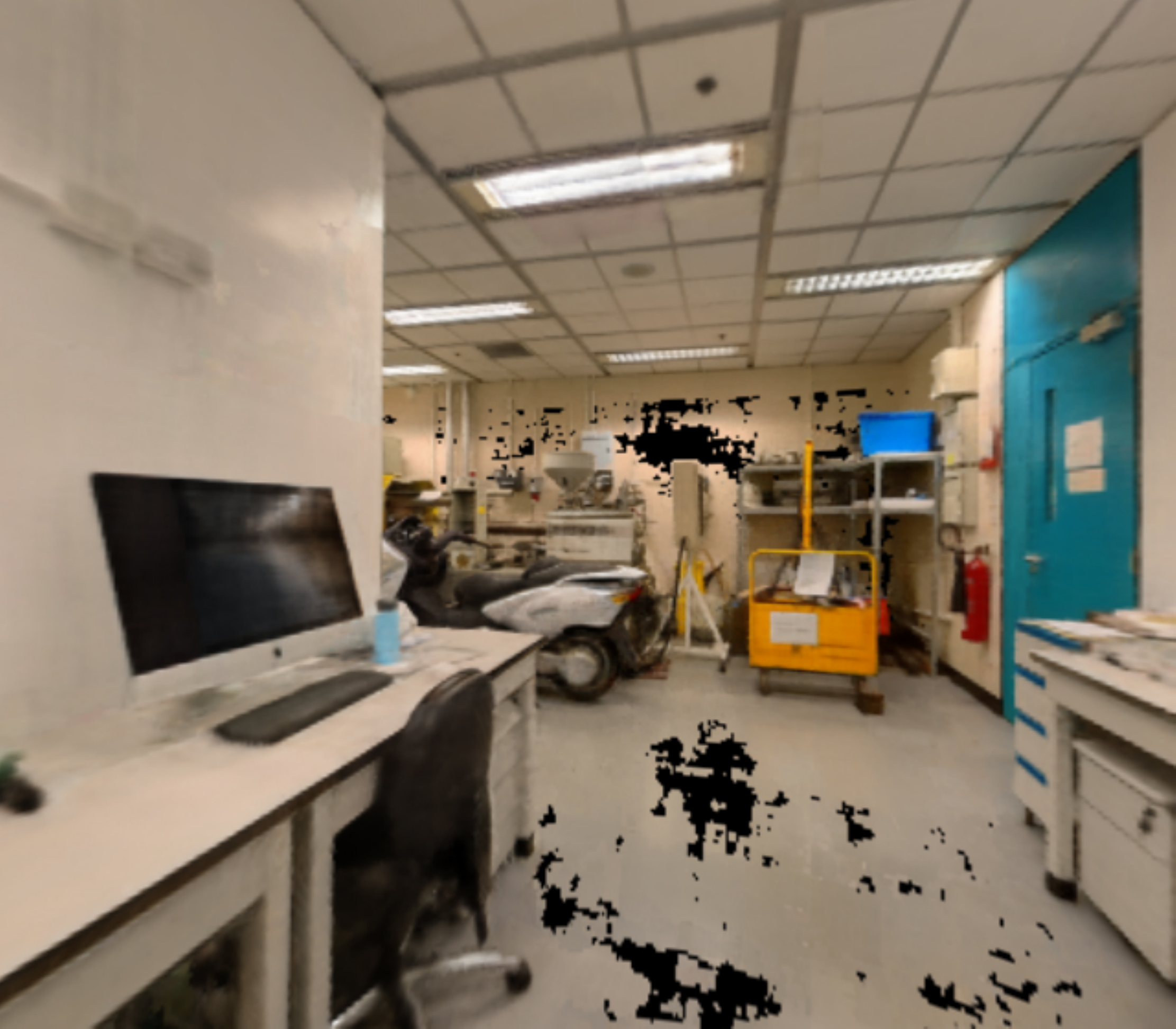}
	\includegraphics[width=\imgw\linewidth]{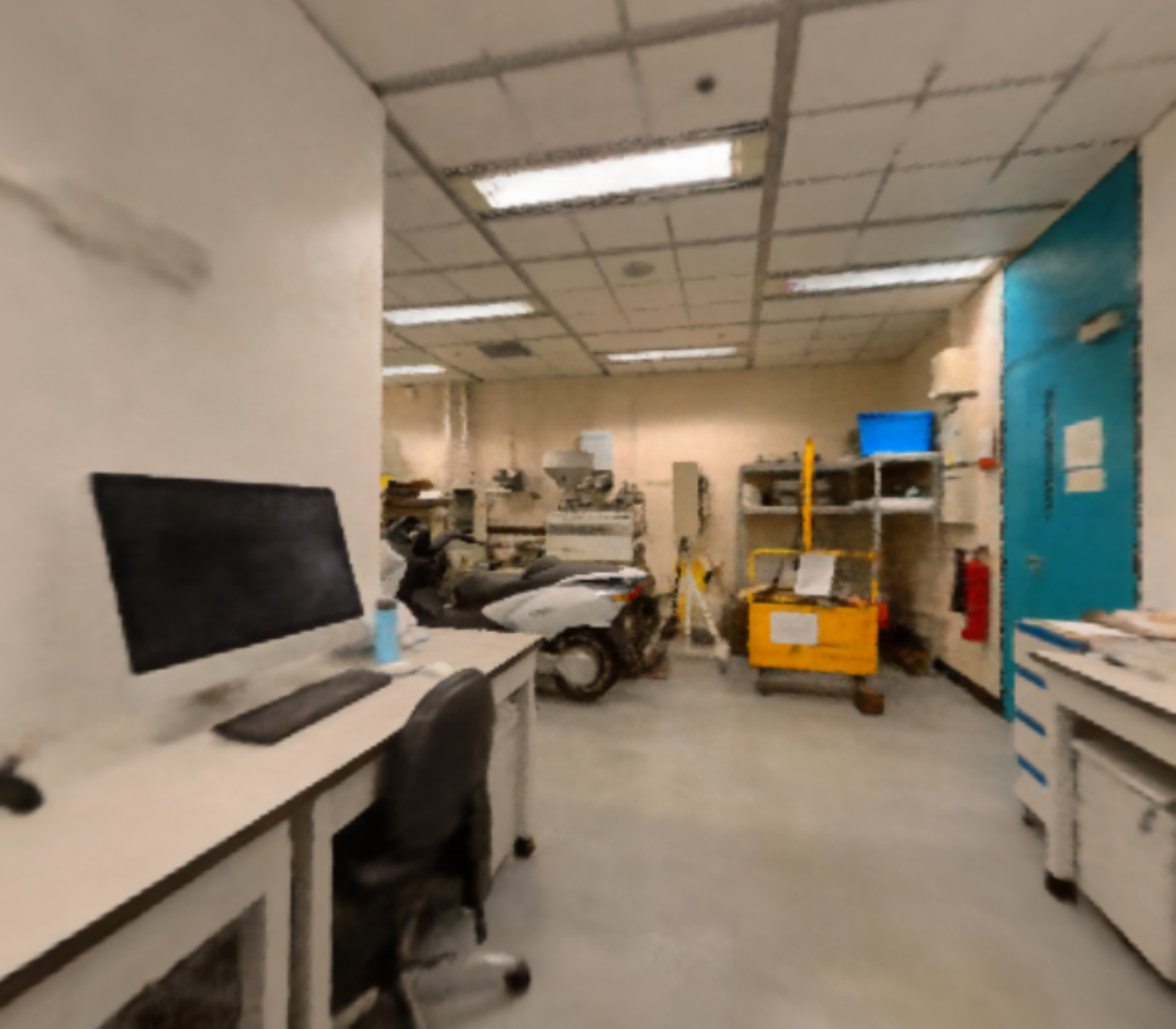}
	\includegraphics[width=\imgw\linewidth]{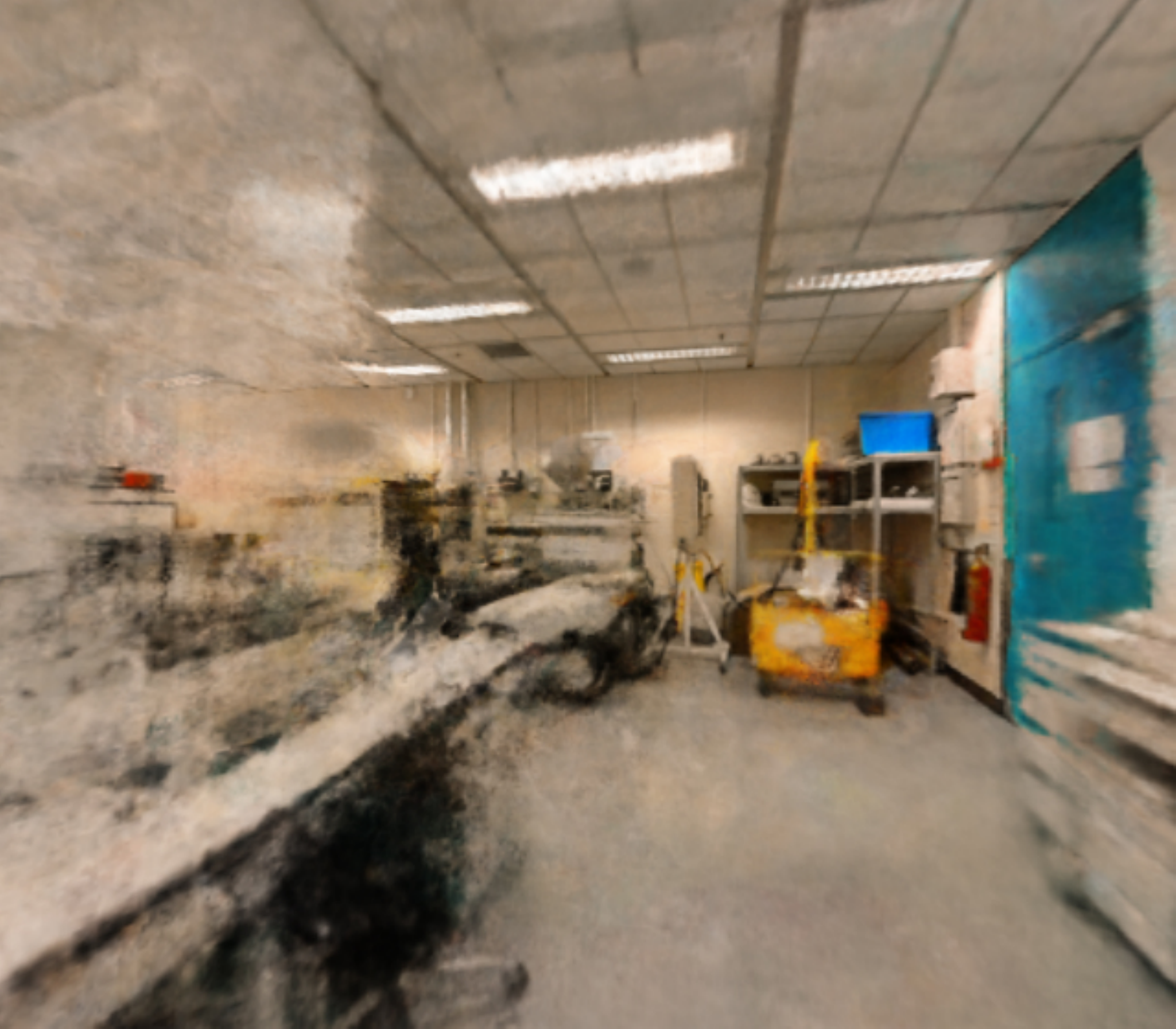}
	\includegraphics[width=\imgw\linewidth]{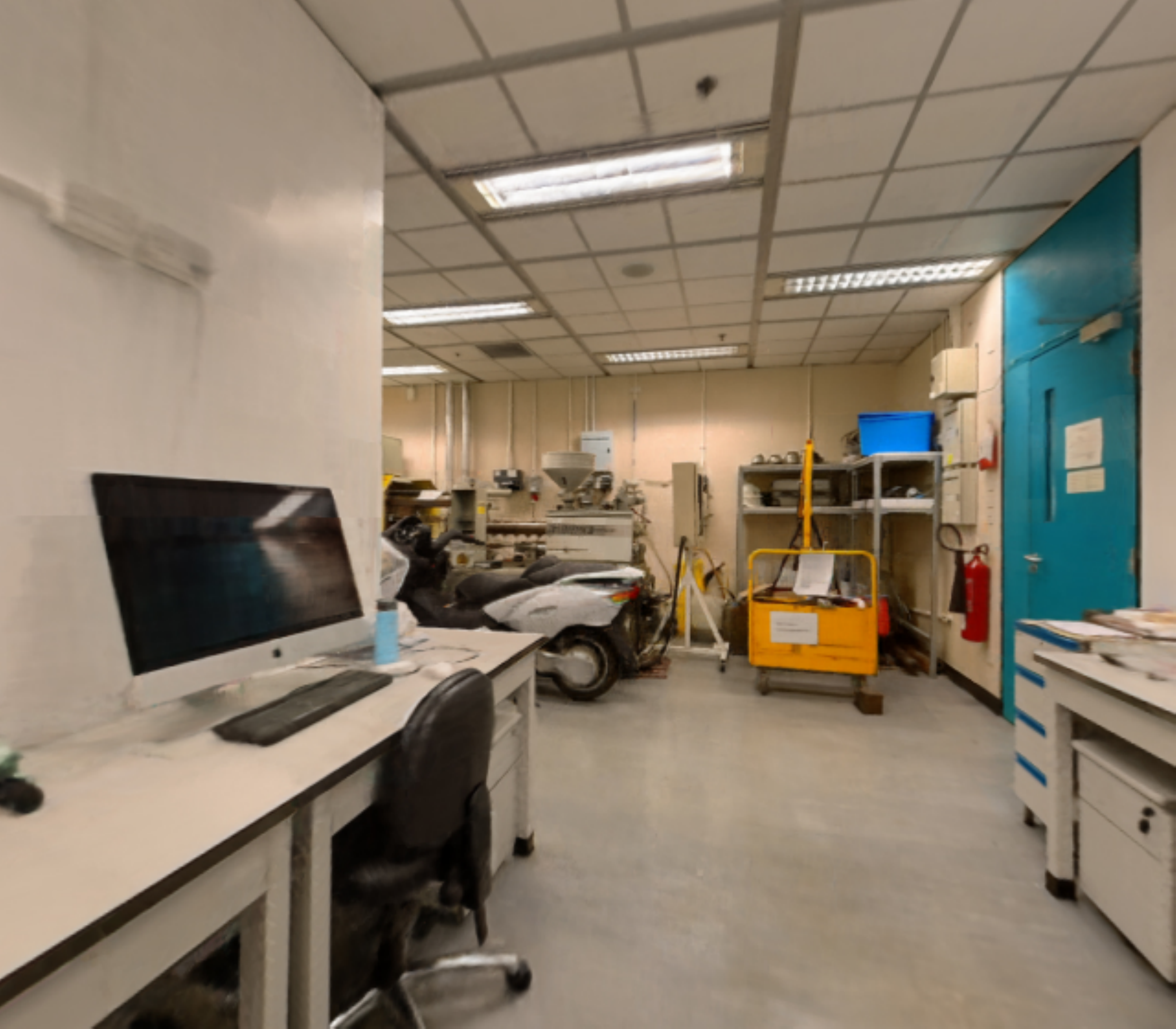}\\
	\vspace{\gaph}
    \includegraphics[width=\imgw\linewidth]{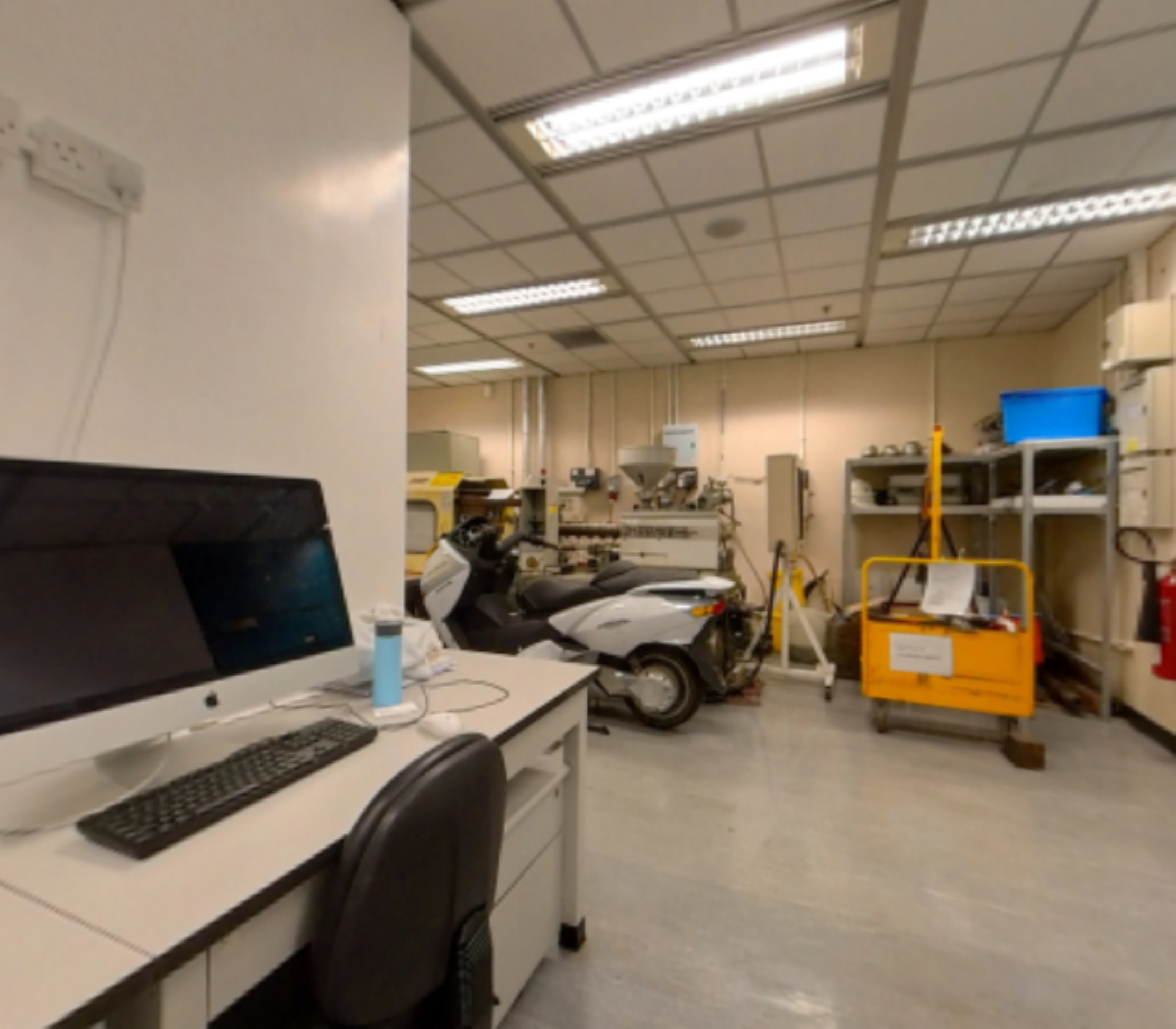}
	\includegraphics[width=\imgw\linewidth]{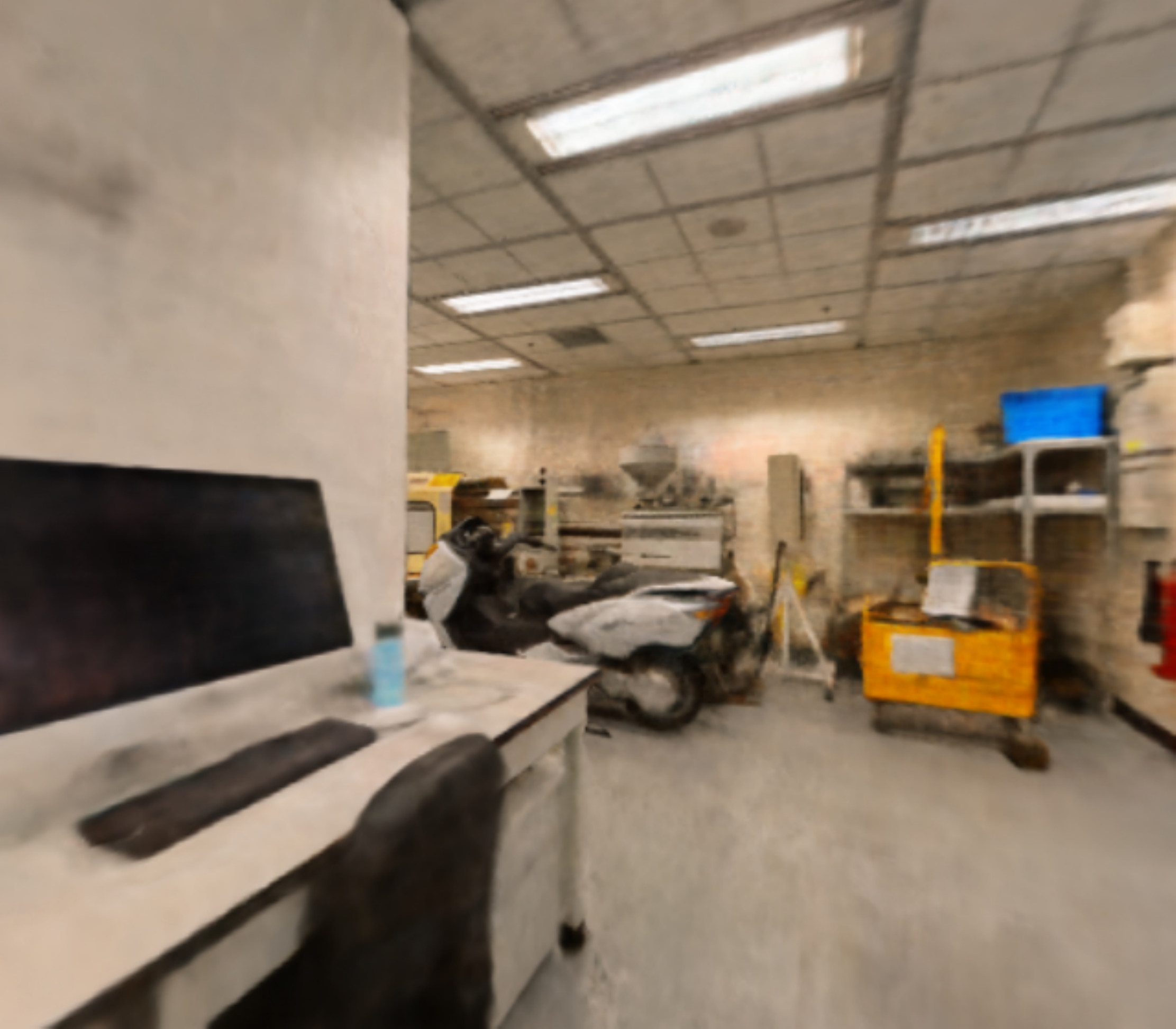}
	\includegraphics[width=\imgw\linewidth]{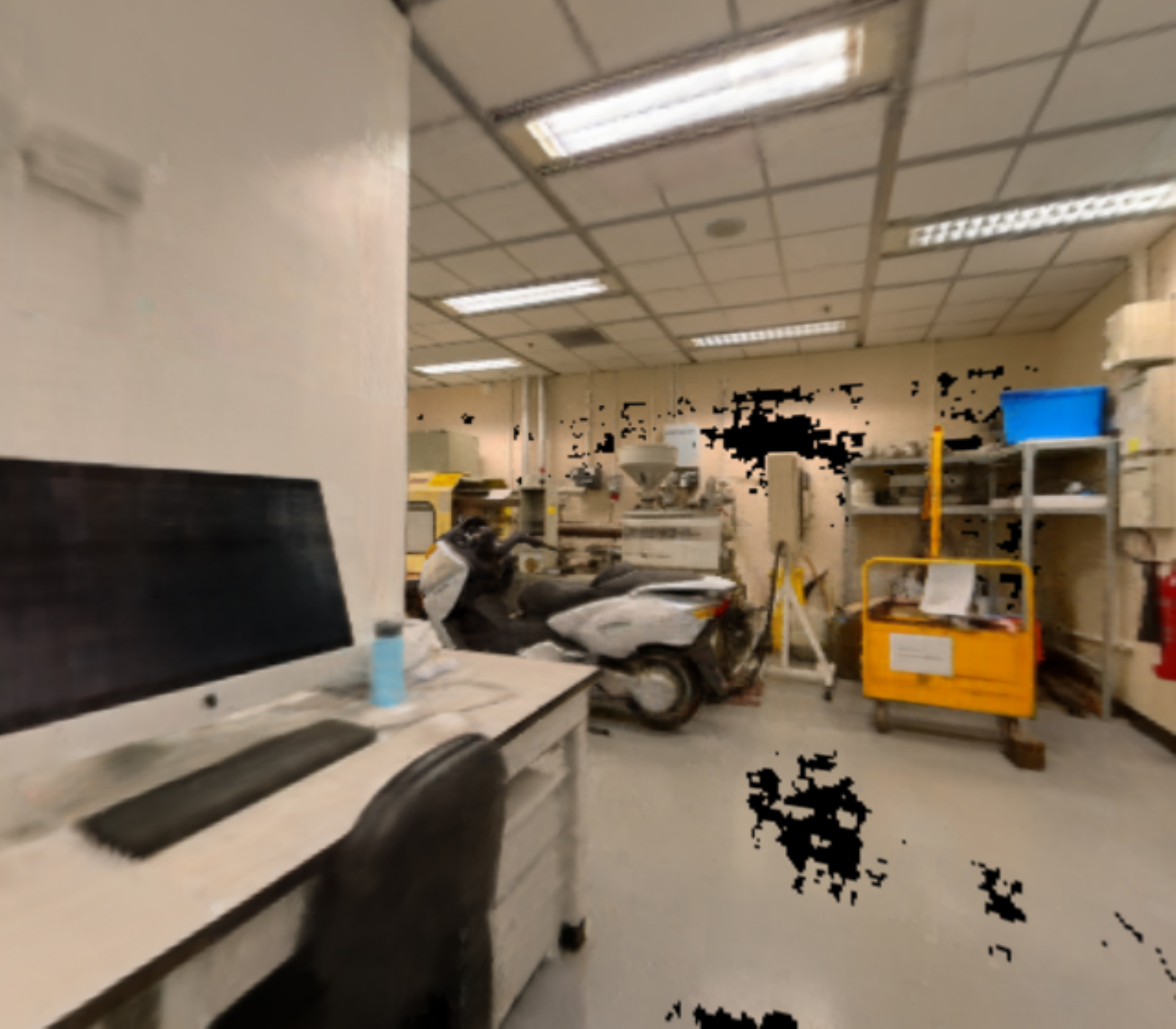}
	\includegraphics[width=\imgw\linewidth]{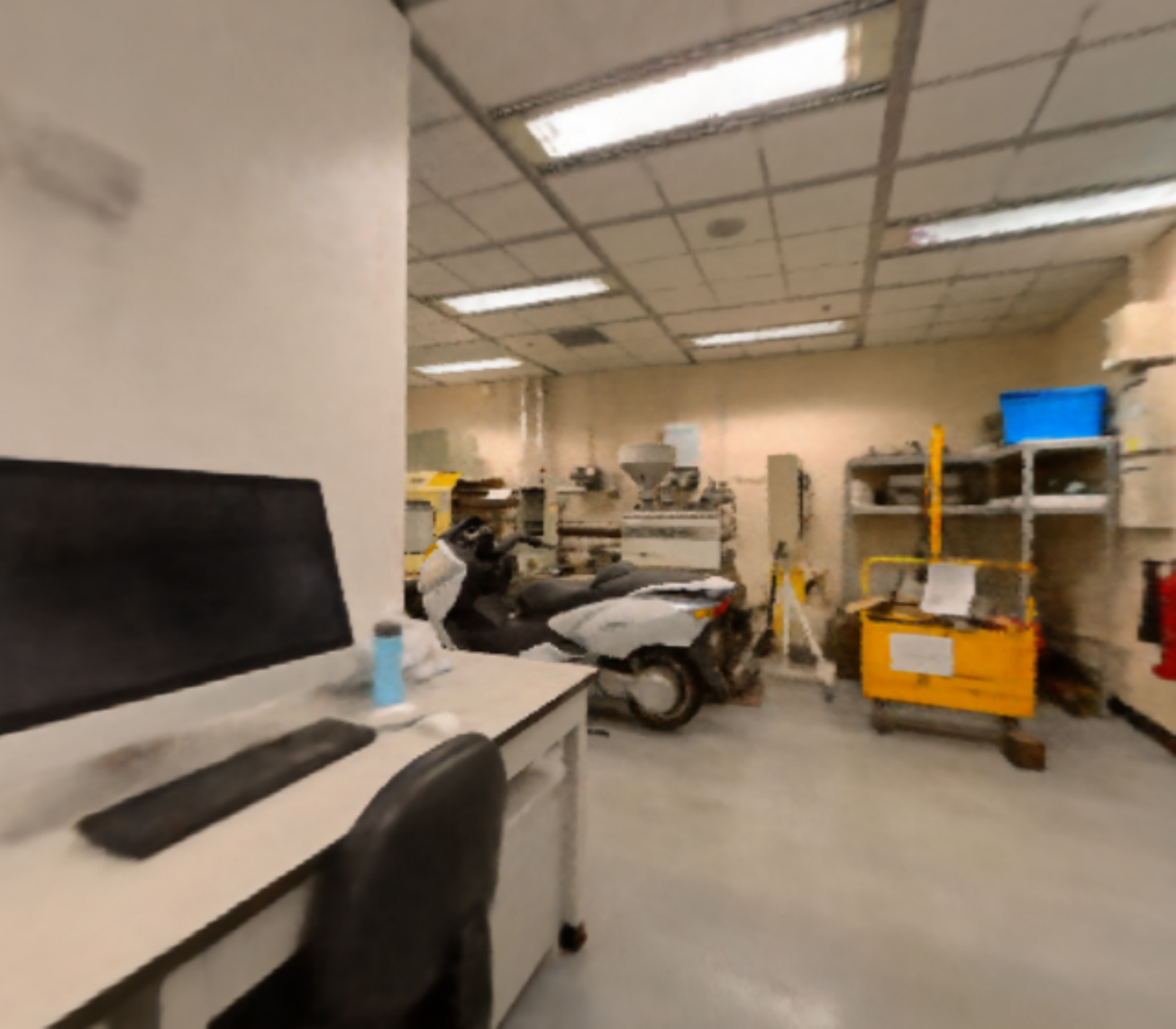}
	\includegraphics[width=\imgw\linewidth]{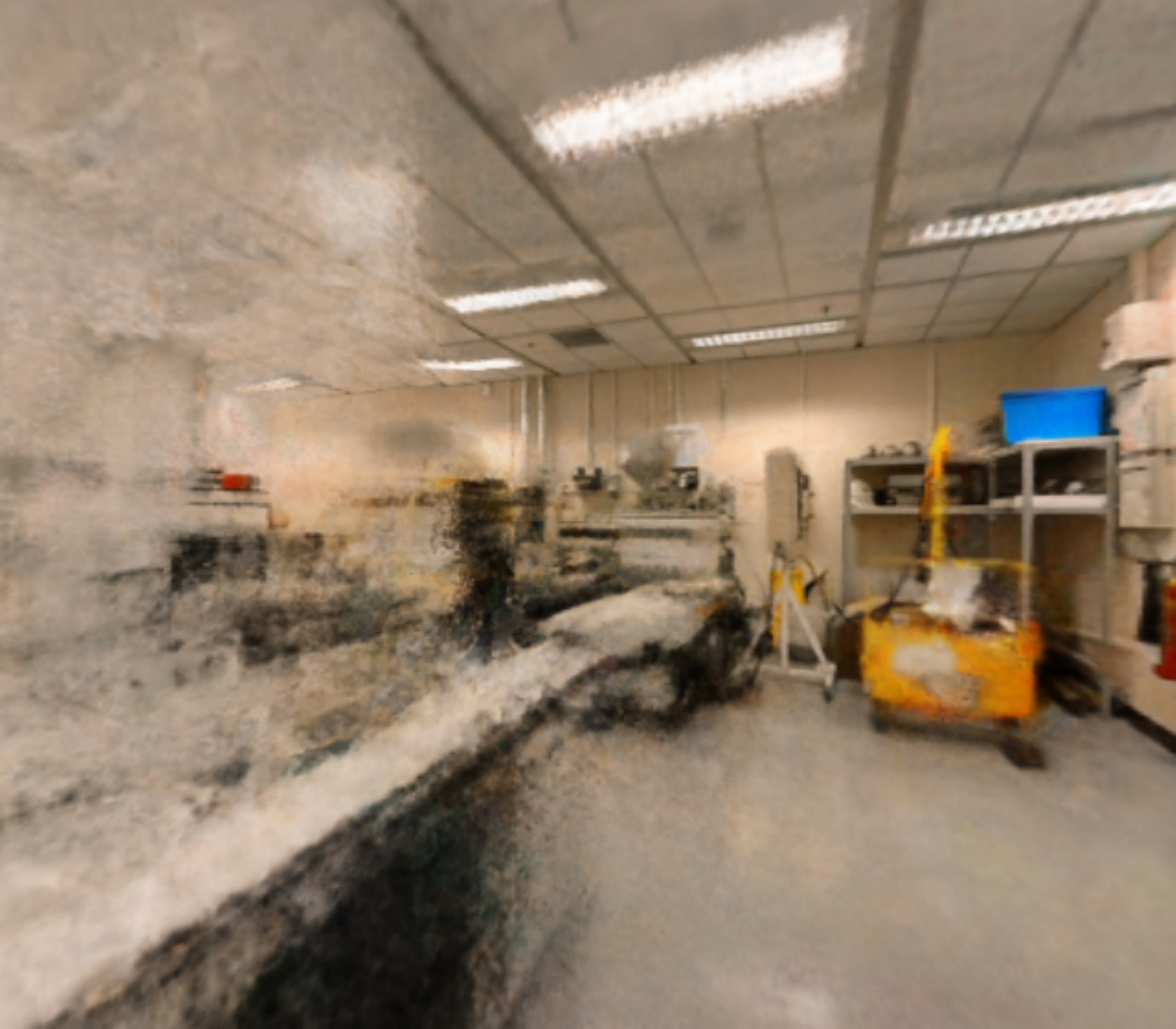}
	\includegraphics[width=\imgw\linewidth]{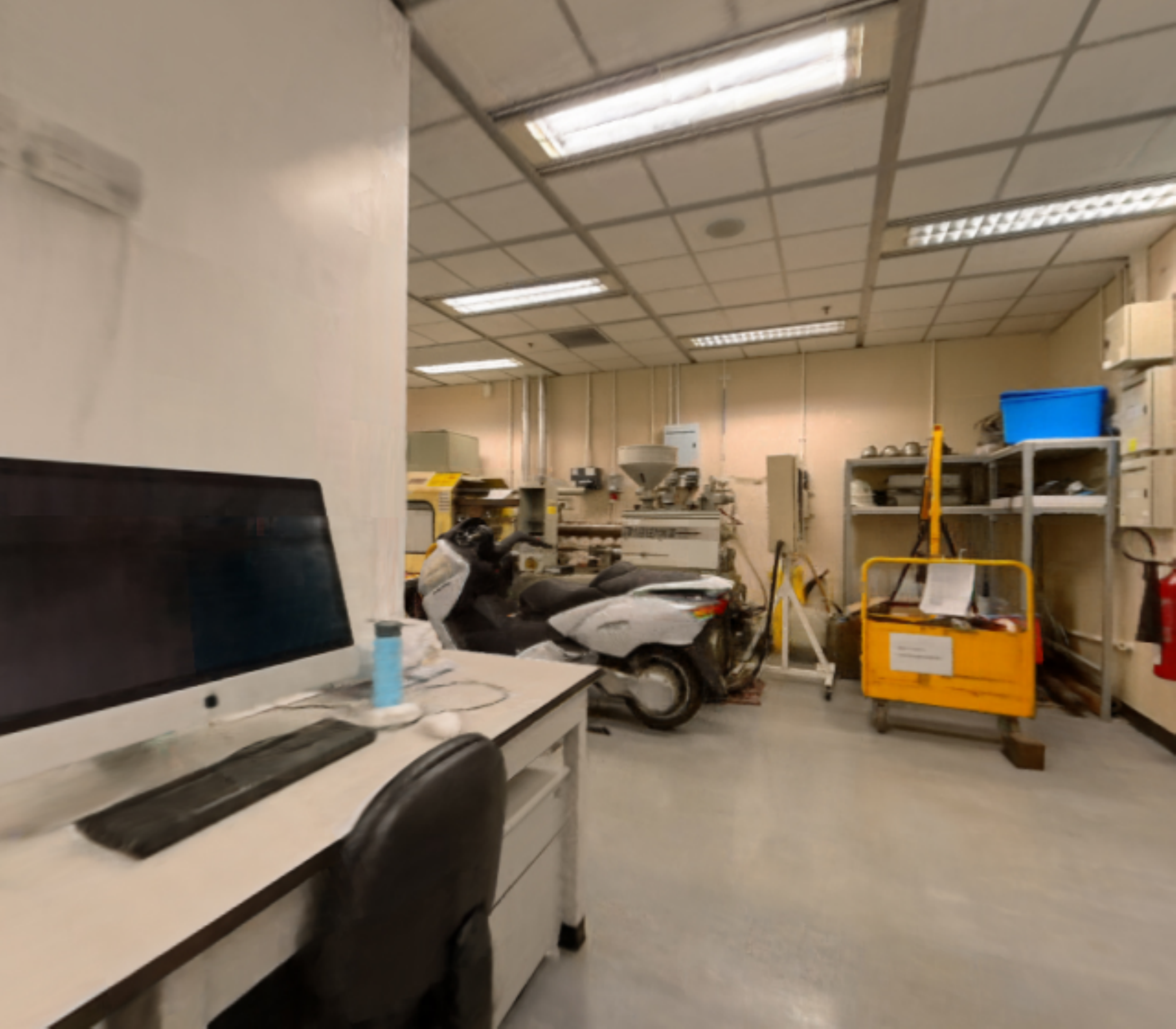}\\
	\vspace{\gaph}
	\includegraphics[width=\imgw\linewidth]{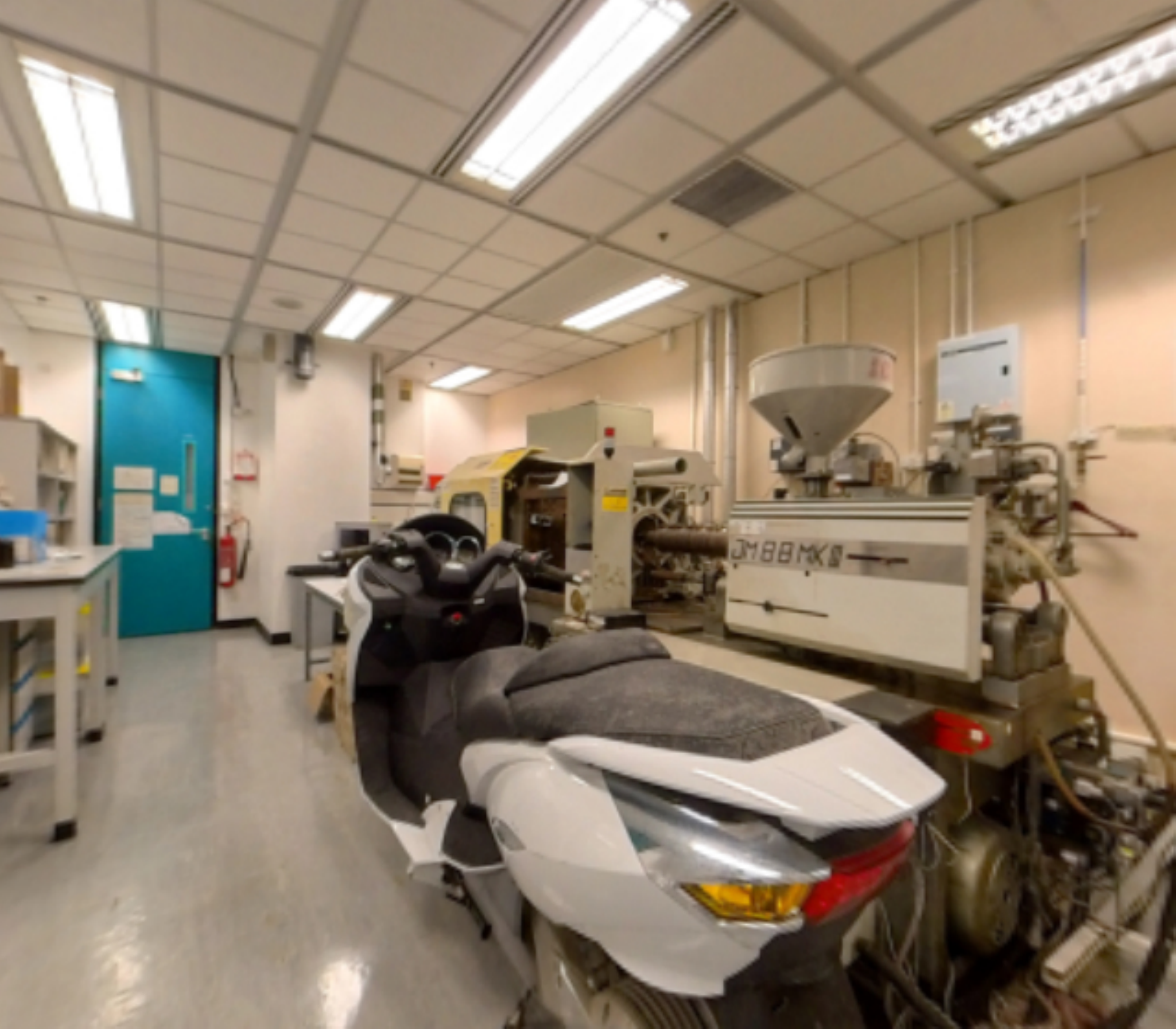}
	\includegraphics[width=\imgw\linewidth]{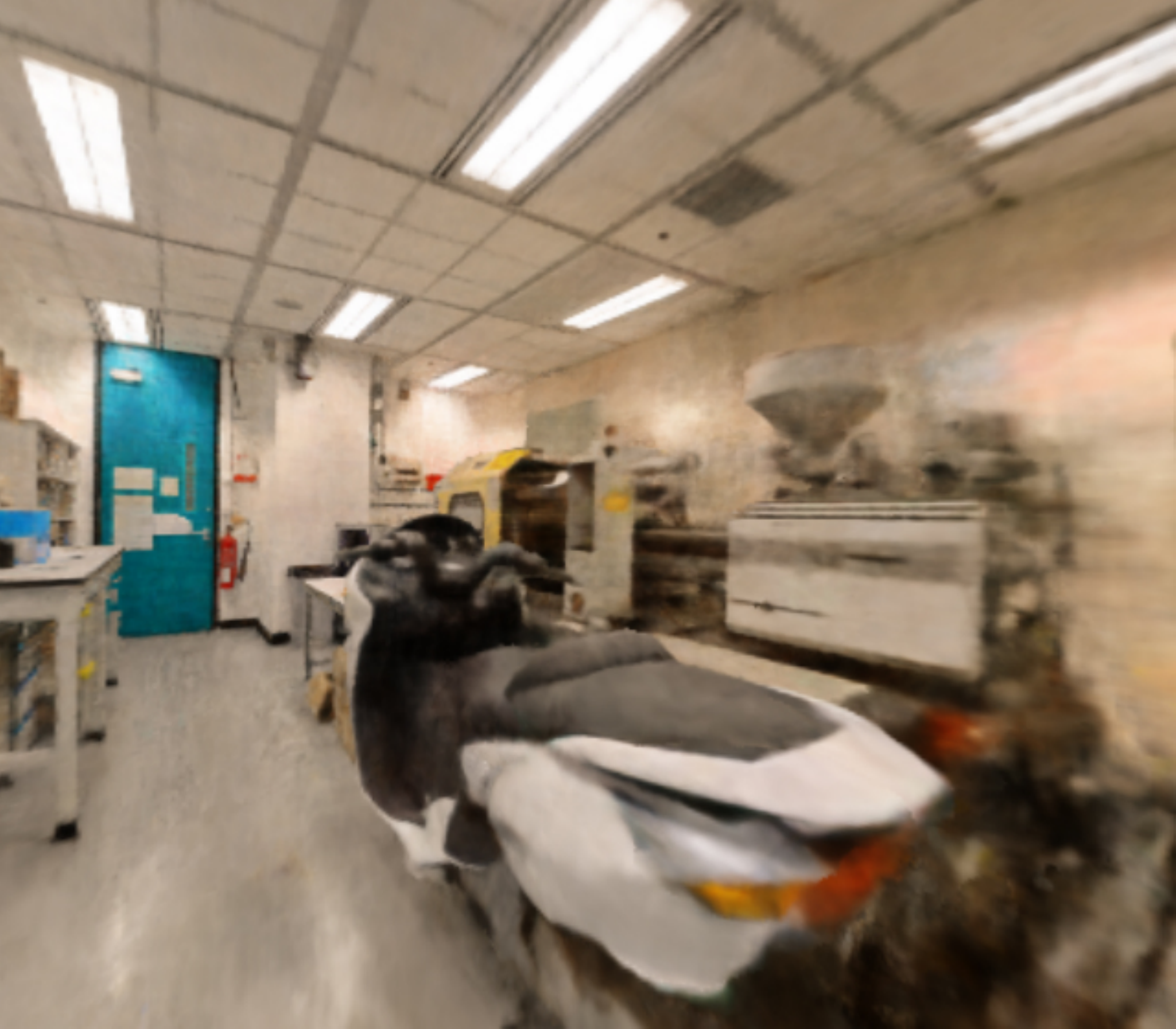}
	\includegraphics[width=\imgw\linewidth]{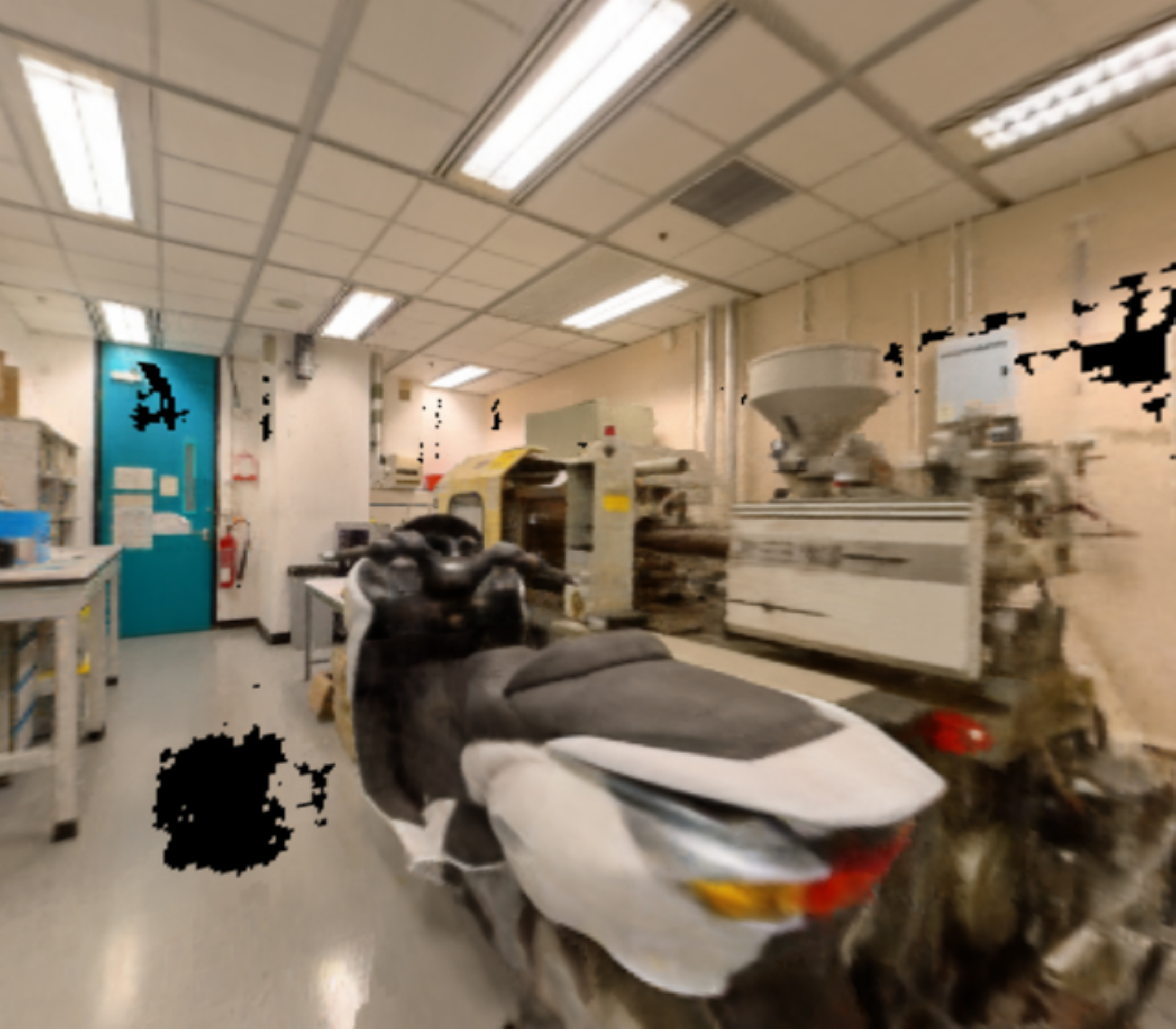}
	\includegraphics[width=\imgw\linewidth]{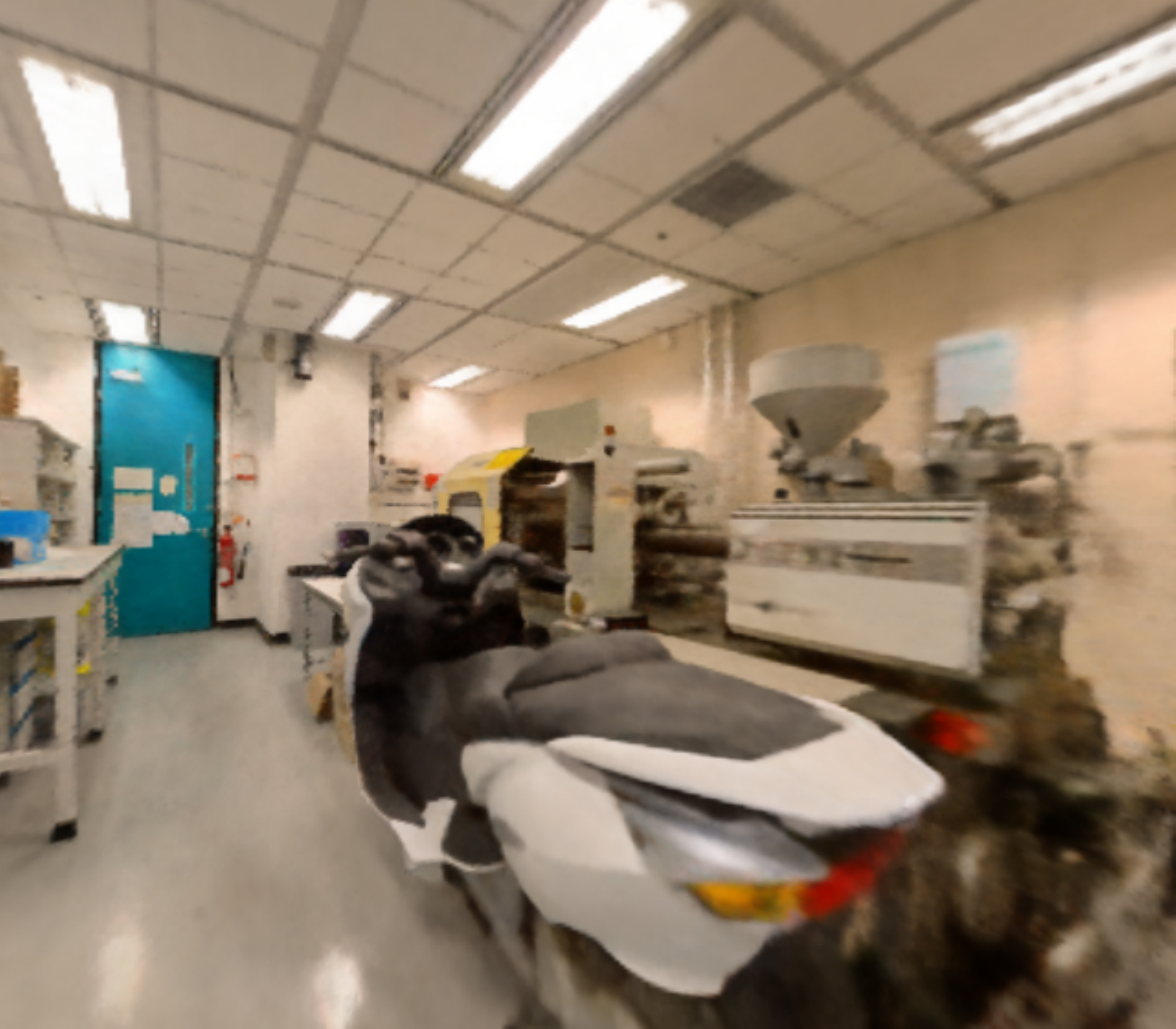}
	\includegraphics[width=\imgw\linewidth]{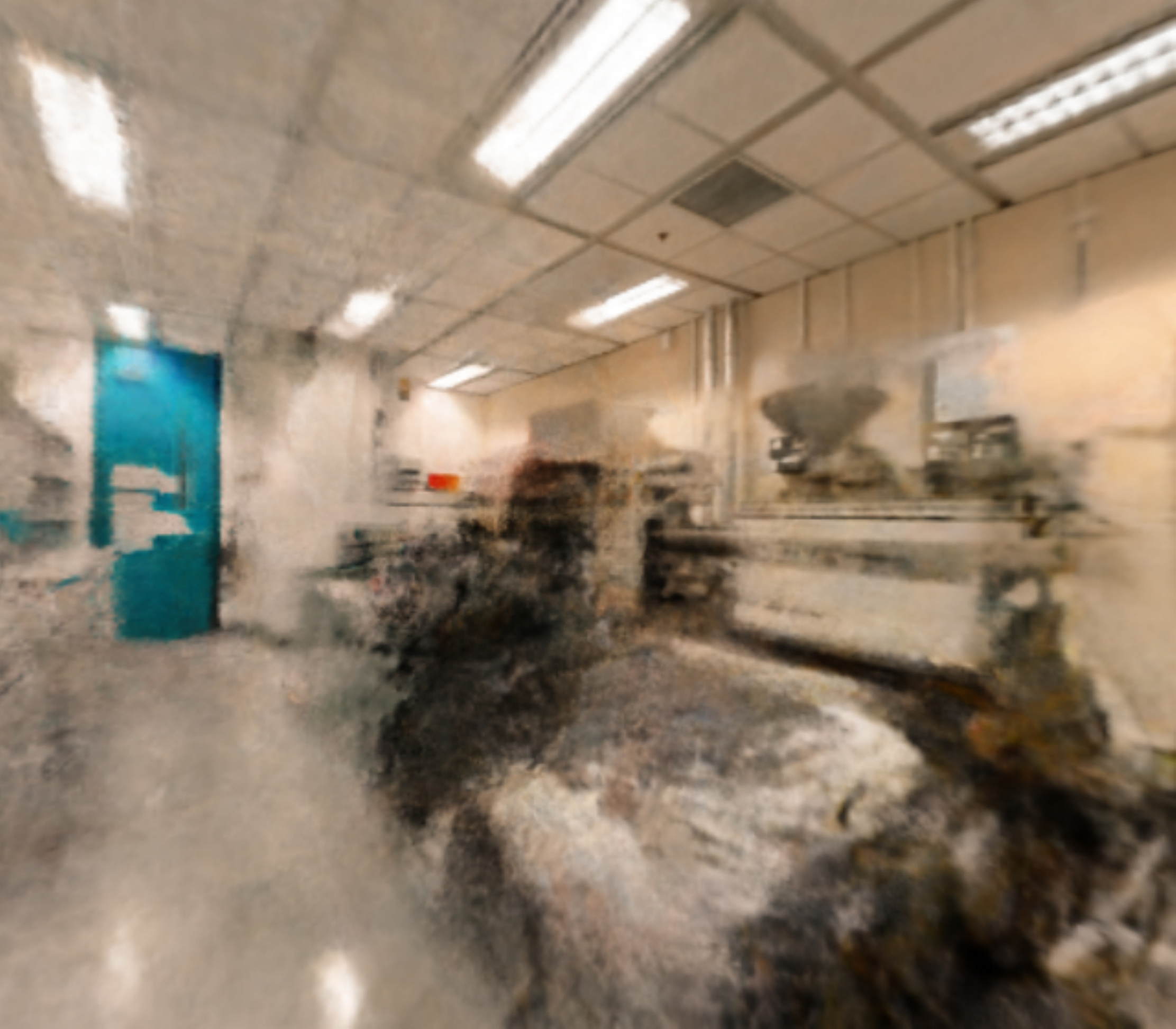}
	\includegraphics[width=\imgw\linewidth]{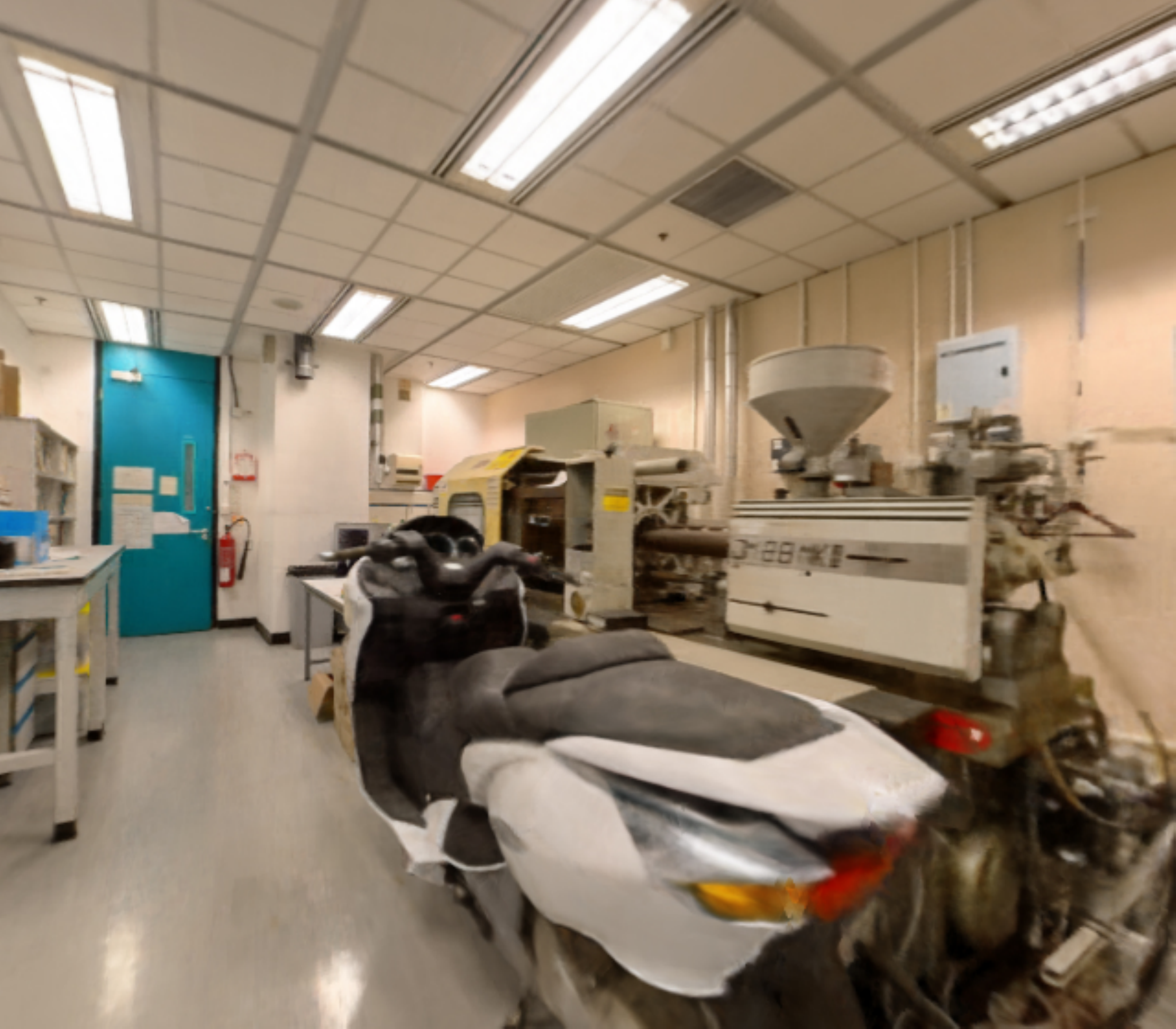}\\
	\vspace{\gaph}
	\includegraphics[width=\imgw\linewidth]{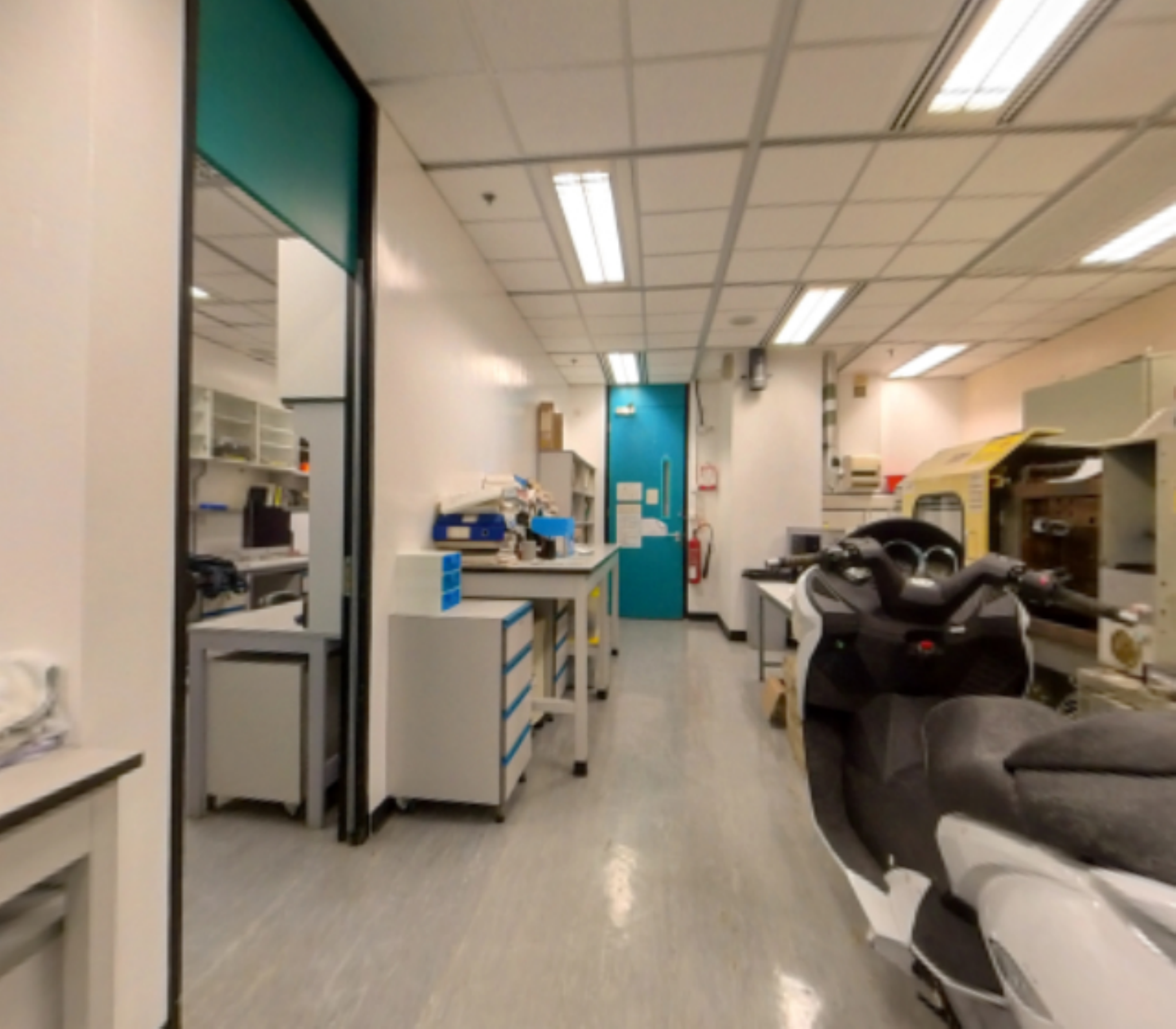}
	\includegraphics[width=\imgw\linewidth]{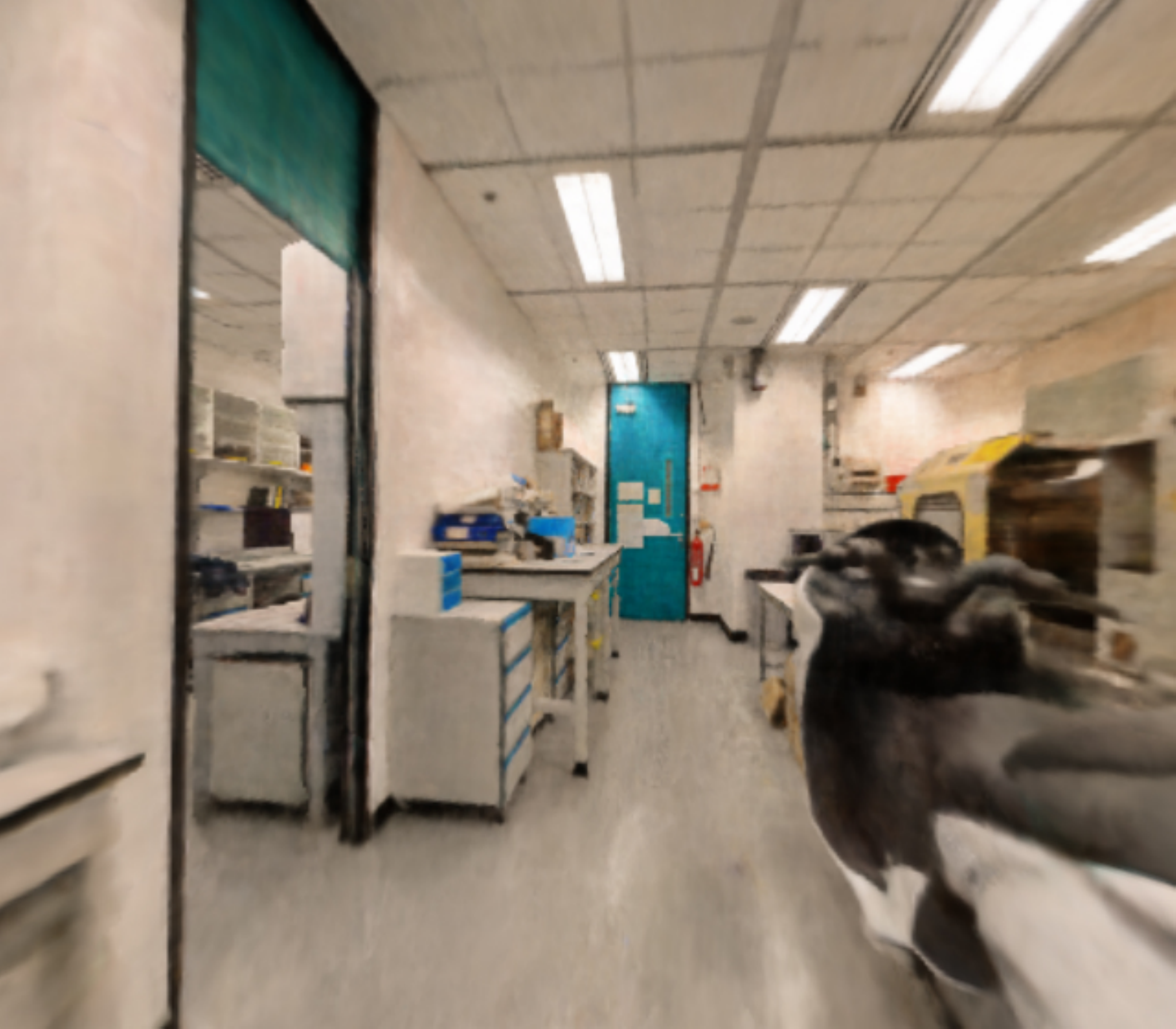}
	\includegraphics[width=\imgw\linewidth]{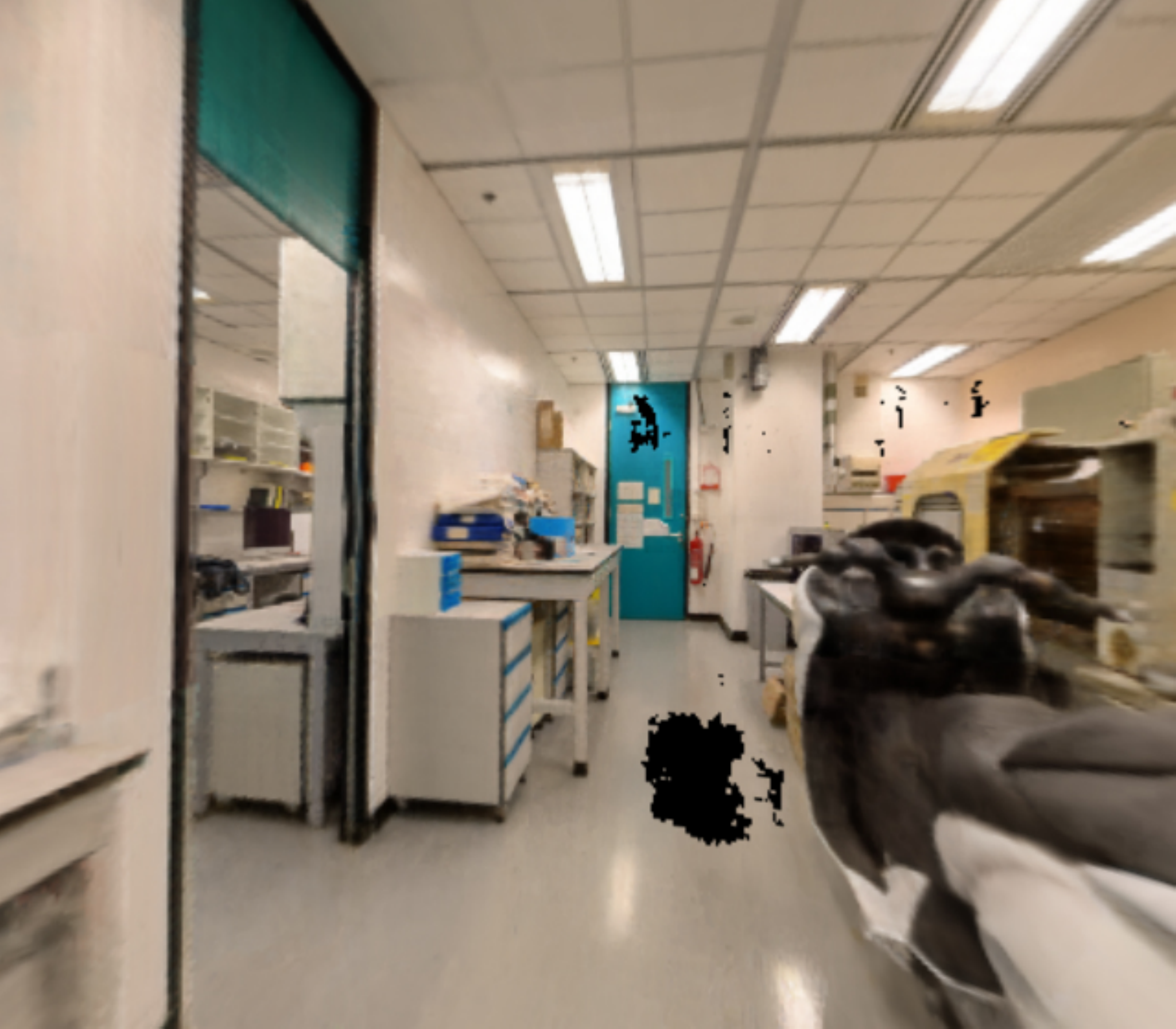}
	\includegraphics[width=\imgw\linewidth]{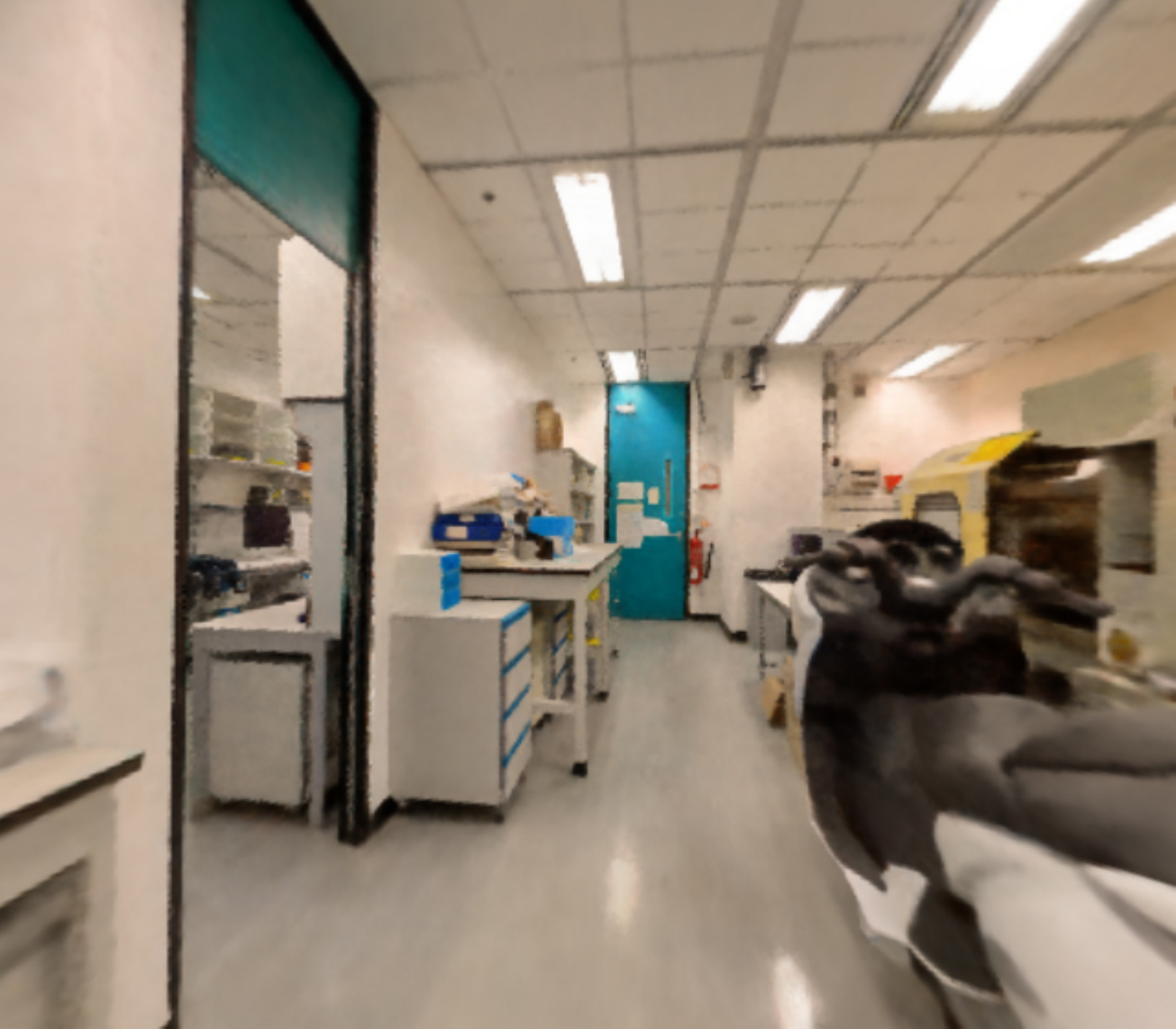}
	\includegraphics[width=\imgw\linewidth]{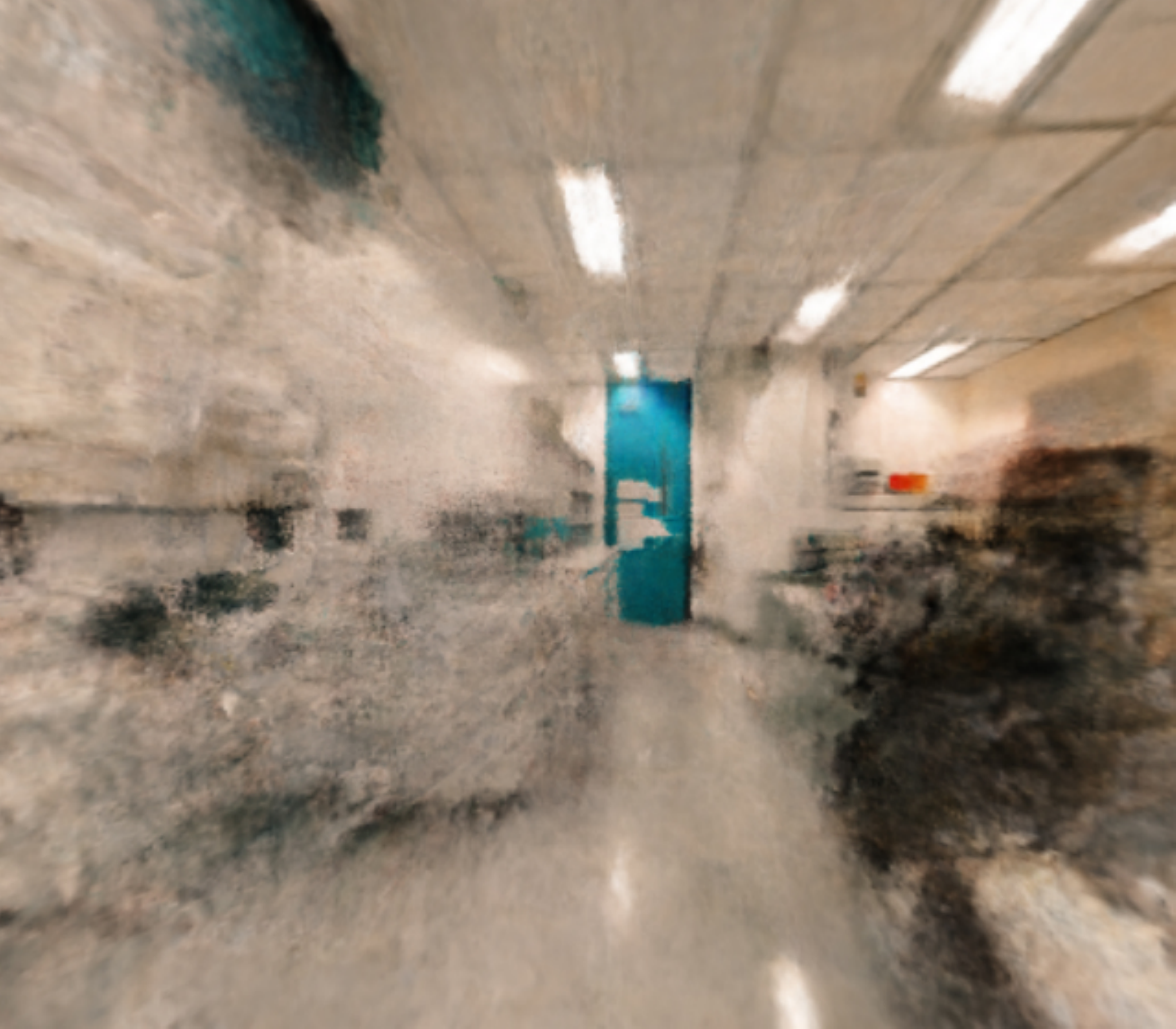}
	\includegraphics[width=\imgw\linewidth]{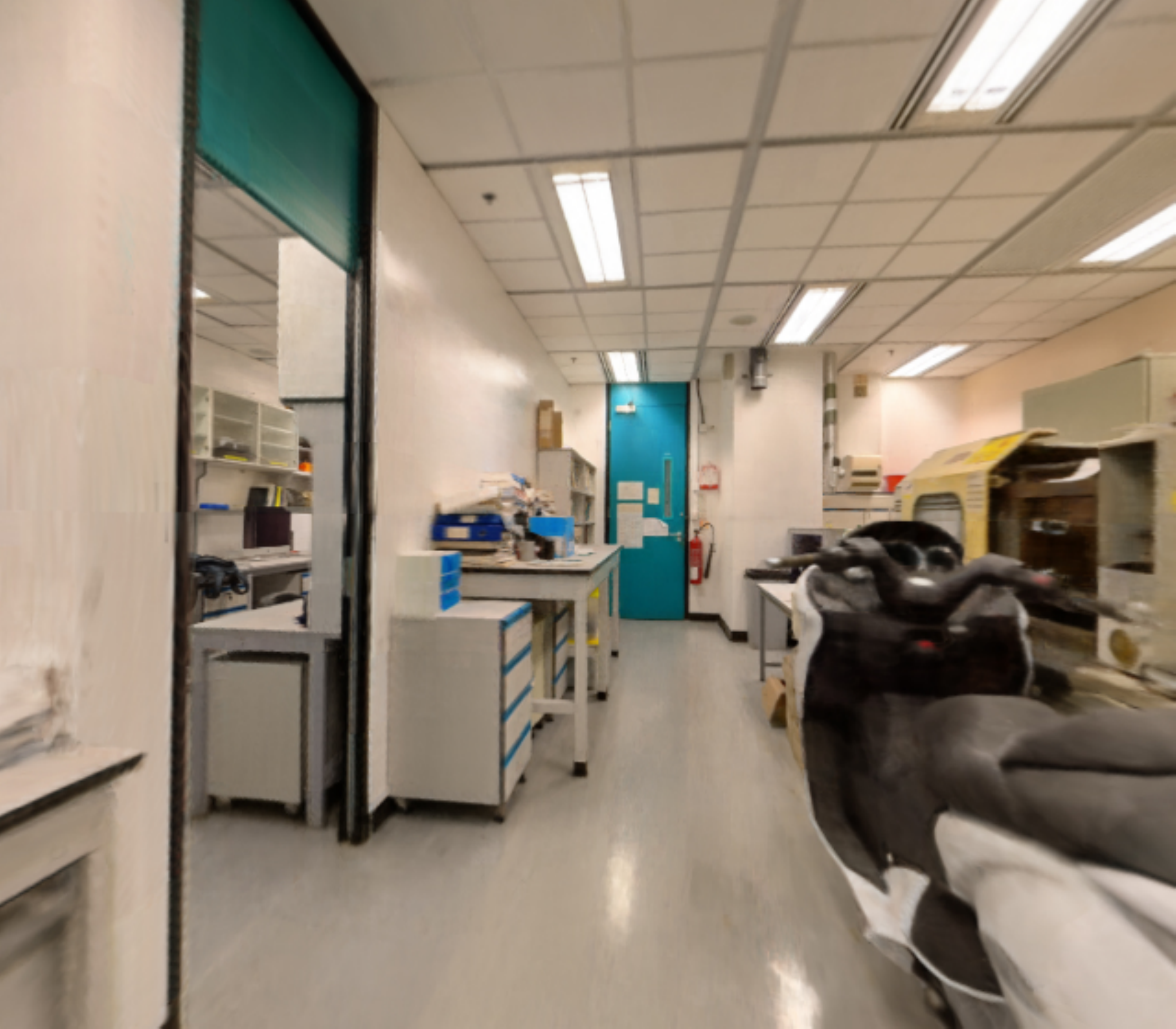}\\
	\vspace{\gaph}
	\includegraphics[width=\imgw\linewidth]{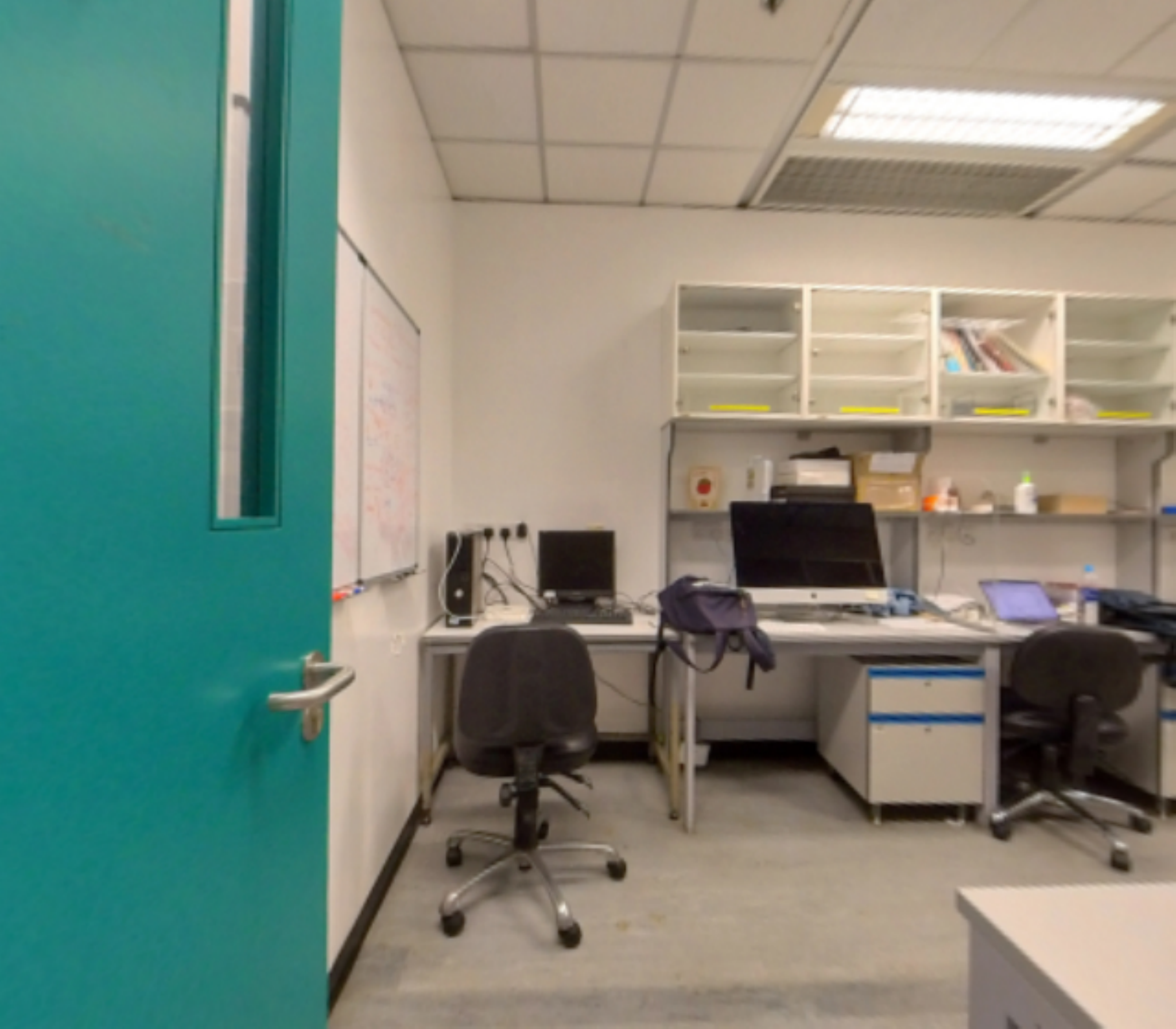}
	\includegraphics[width=\imgw\linewidth]{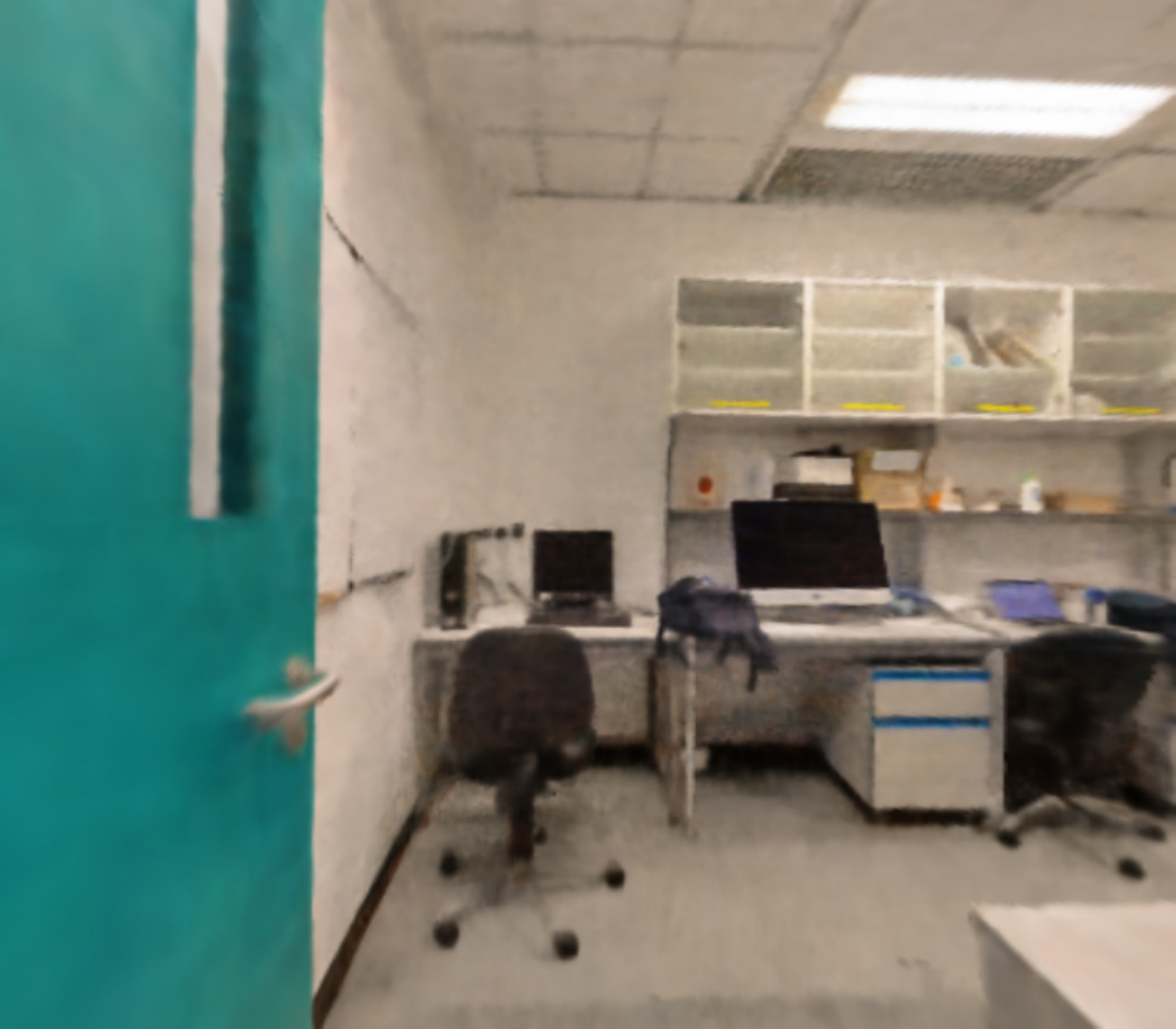}
	\includegraphics[width=\imgw\linewidth]{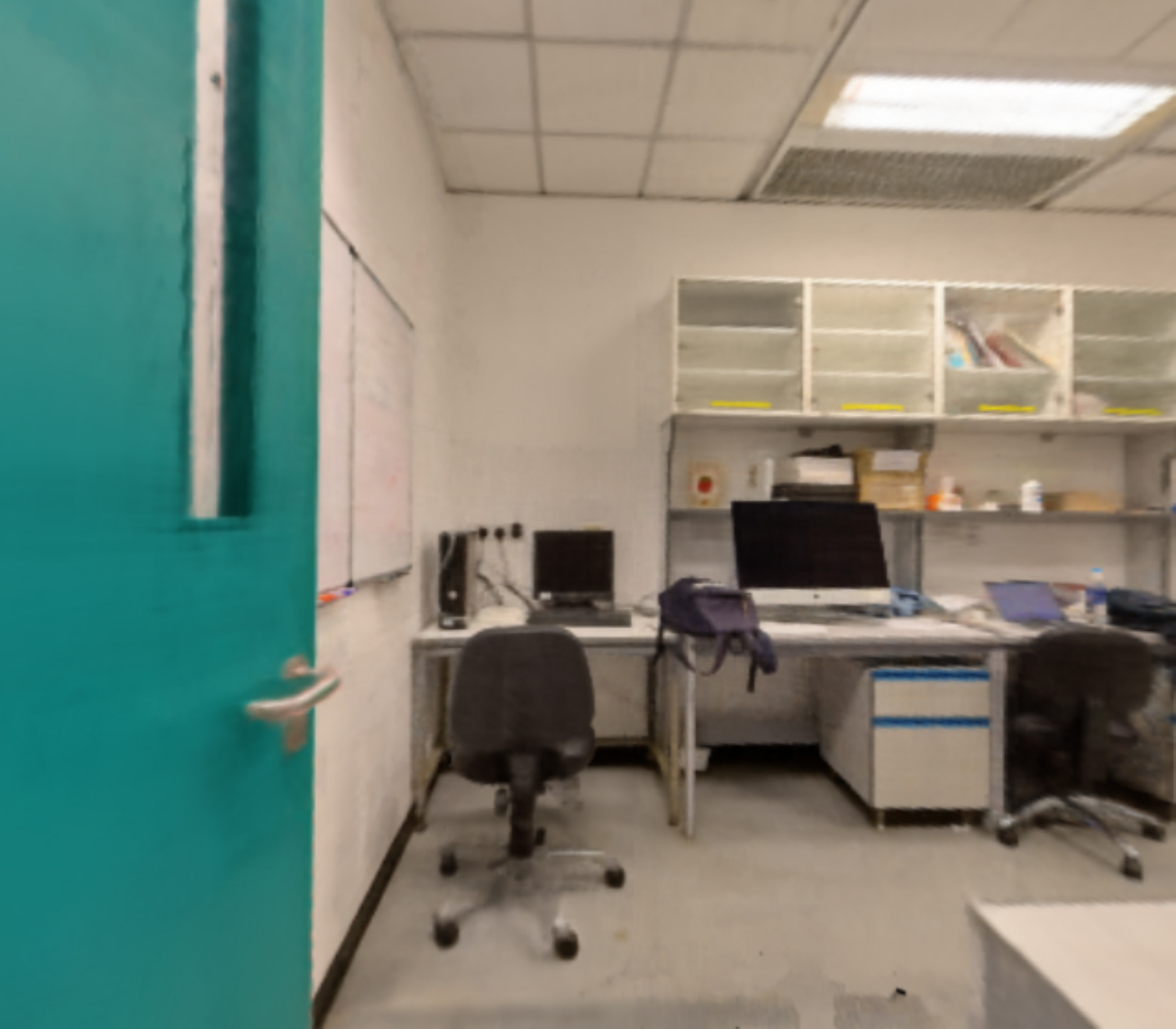}
	\includegraphics[width=\imgw\linewidth]{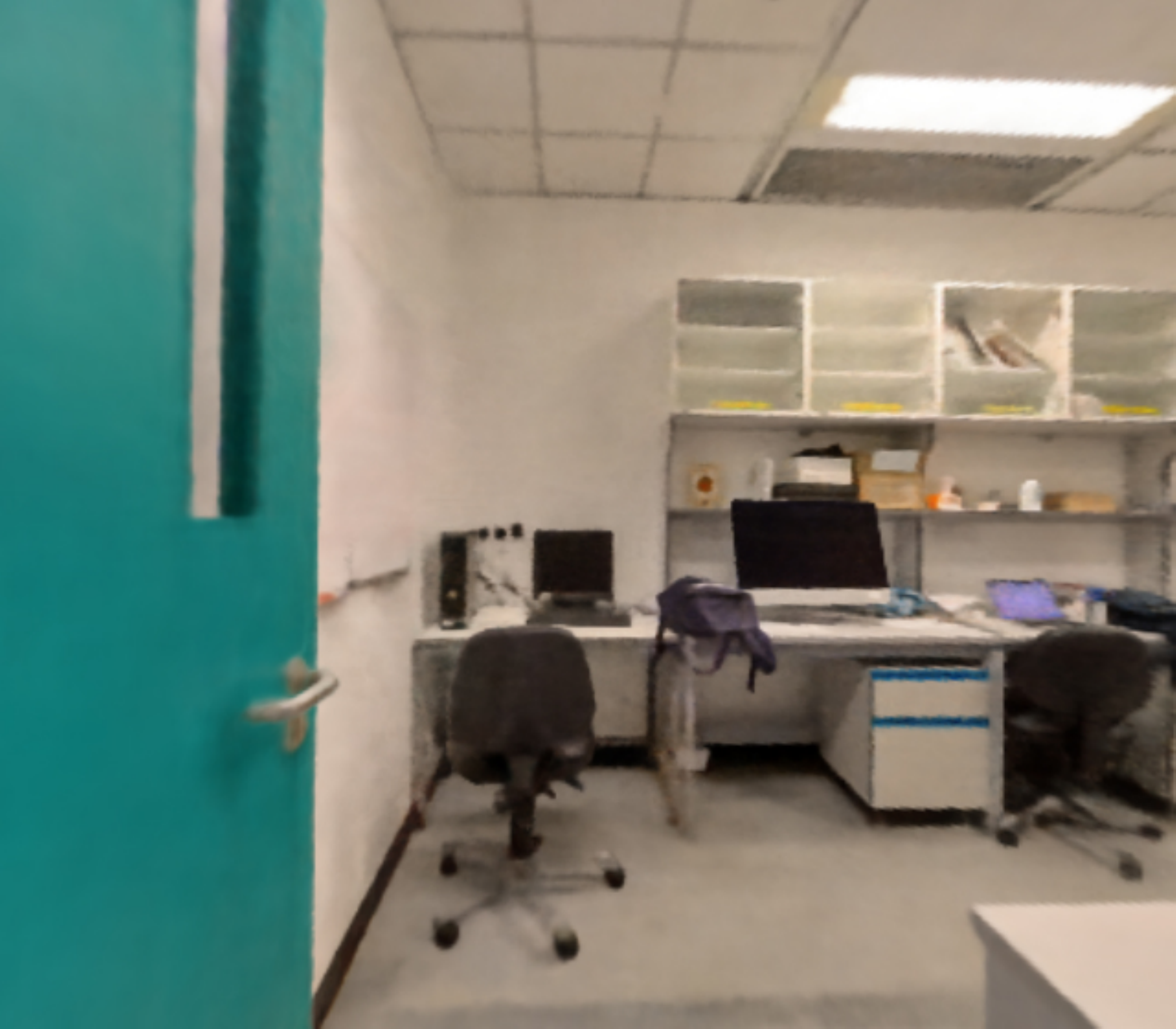}
	\includegraphics[width=\imgw\linewidth]{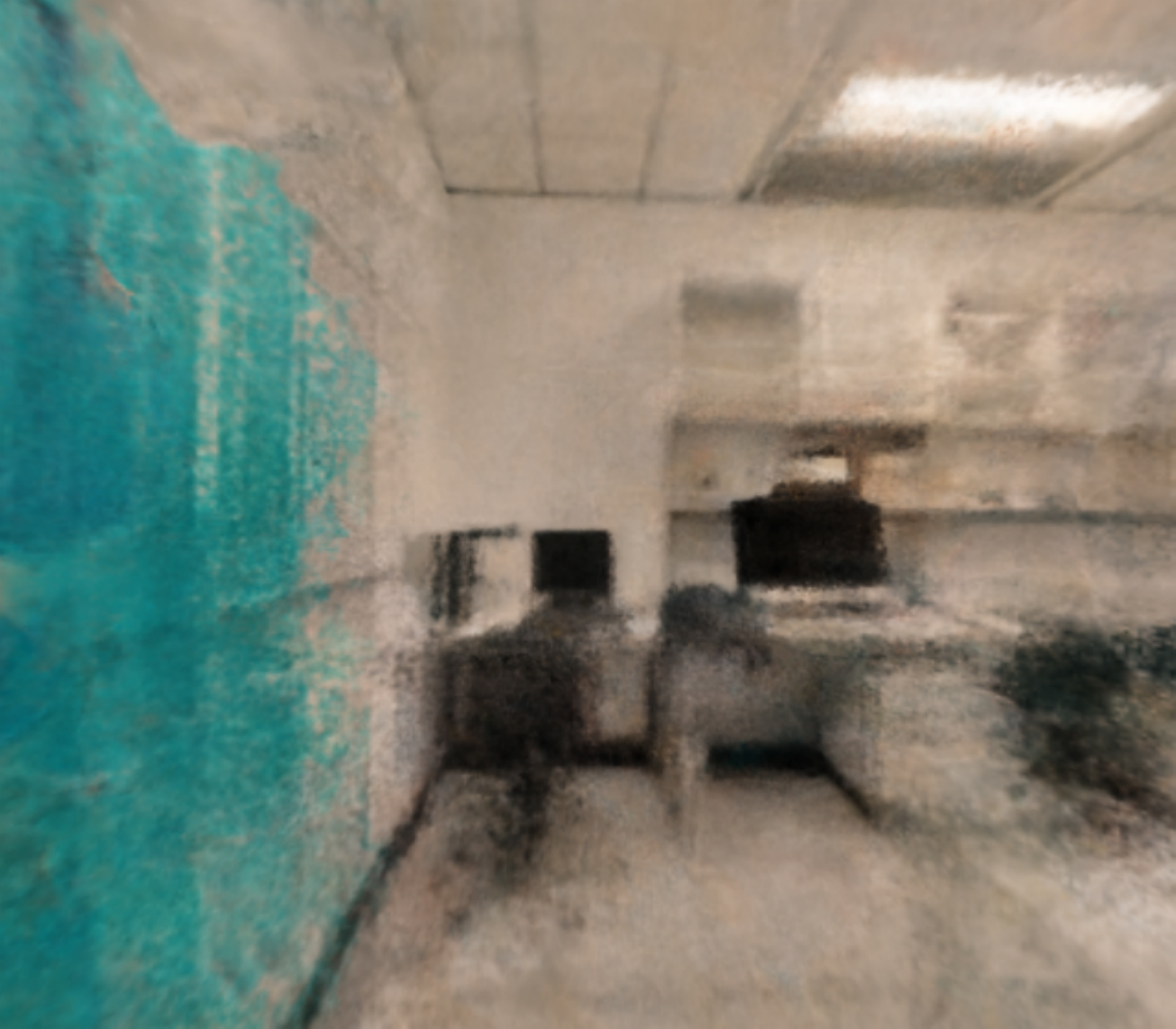}
	\includegraphics[width=\imgw\linewidth]{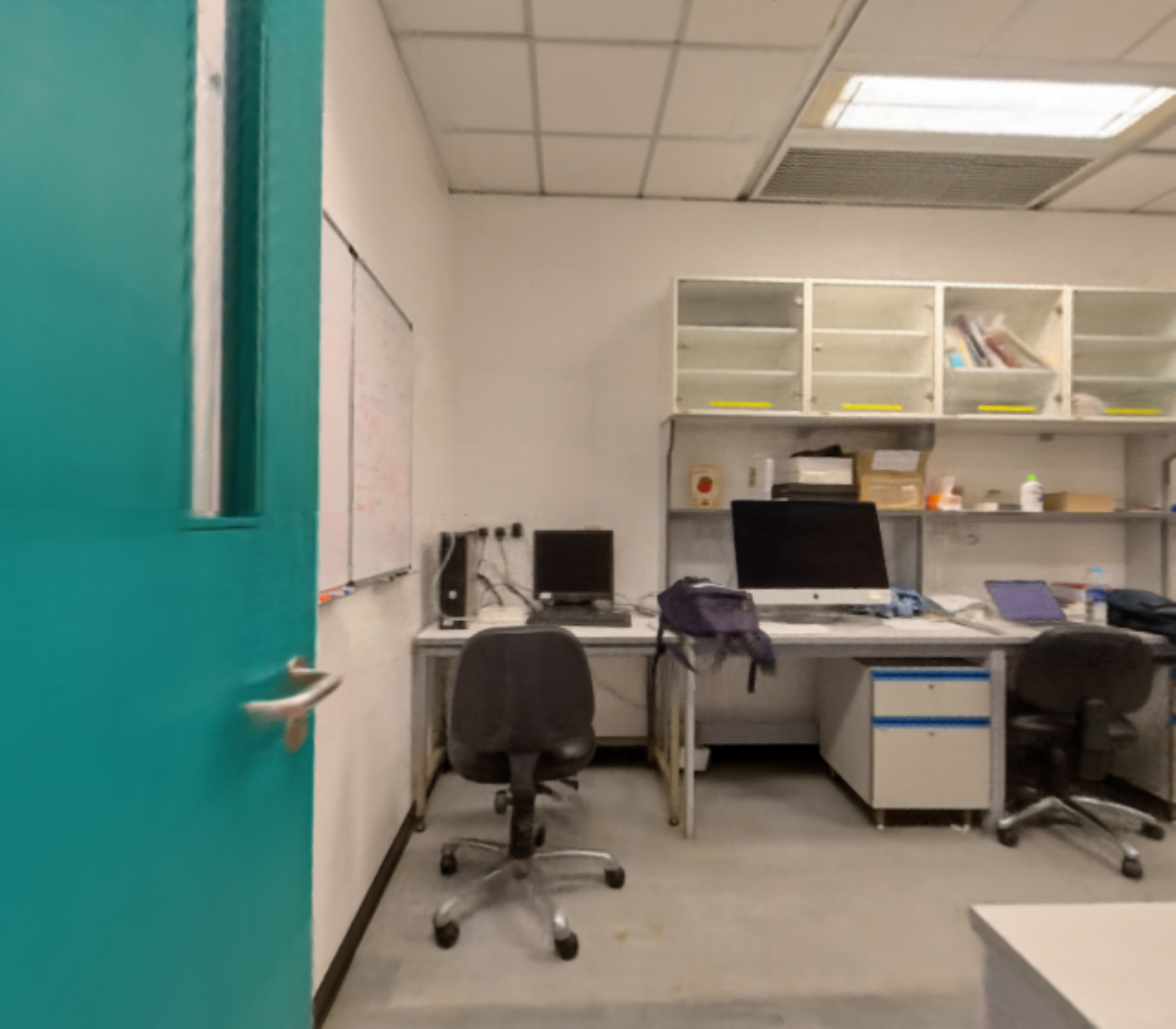}
	
	\begin{tikzpicture}
        \def\imgh{0.6}
        \def\citationH{0.3}
		\node[inner sep=0pt] () at (-0.42\linewidth,\imgh){\small Ground Truth};
  
		\node[inner sep=0pt] () at (-0.25\linewidth,\imgh){\small NeRF \cite{mildenhall2020nerf}};
  
		\node[inner sep=0pt] () at (-0.08\linewidth,\imgh){\small KiloNeRF \cite{reiser2021kilonerf}};
  
		\node[inner sep=0pt] () at (0.08\linewidth,\imgh){\small Mip-NeRF360 \cite{barron2022mip360}};
  
		\node[inner sep=0pt] () at (0.25\linewidth,\imgh){\small instant-NGP \cite{muller2022instant}};
  
		\node[inner sep=0pt] () at (0.42\linewidth,\imgh){\small Our approach};
    \end{tikzpicture}
    \caption{Qualitative comparison of novel view synthesis results with large translational motion on the scene \textsc{Lab}. Zoom in for a better view.}
	\label{fig:comparison}
\end{figure*}

\begin{table}[t]
    \tabcolsep=0.08cm
	\begin{tabular}{ccccc}
		\toprule
		& \footnotesize PSNR$\uparrow$ & \footnotesize SSIM$\uparrow$ & \footnotesize LPIPS$\downarrow$ & Inference Speed (s)\\
		\midrule
        NSVF & 17.050 & 0.516 & 0.758 & 3 \\
        NeRF & 22.443 & 0.672 & 0.339 & 60 \\
        KiloNeRF & 21.426 & 0.690 & 0.283 & 0.04 \\
        Mip-NeRF360 &24.579 &0.748 &0.269 & 14\\
        Mip-NeRF360 (*1024) &\textbf{25.464} &\textbf{0.789} &\textbf{0.198} & 39\\
        instant-NGP &17.018 &0.548 &0.532 & 0.25\\
        TensoRF &15.035 &0.531 &0.676 & 2.8\\
		\midrule
        360NeRF & 23.048 & 0.720 & 0.297 & 31 \\
        360Roam-100 & 23.918 & 0.692 & 0.315 & 0.7 \\
        360Roam-200 & 24.357 & 0.713 & 0.281 & 0.4 \\
        360Roam-u & 22.415 & 0.704 & 0.268 & \textbf{0.03} \\
        360Roam & \textbf{25.061} & \textbf{0.760} & \textbf{0.202} & \textbf{0.03}\\
		\bottomrule			

	\end{tabular}
	\caption{Comparison of novel view synthesis on 360Roam dataset. The approximate time to render one image is also reported. * indicates using a wider MLP network with 1024 channels in each layer. 
	}
	\label{tab:comparison}
\end{table}

\subsubsection{Quantitative Comparison} 
Quantitative measurements in terms of appearance similarity include Peak Signal-to-noise Ratio (PSNR), Structural Similarity Index Measure (SSIM) and Learned Perceptual Image Patch Similarity (LPIPS)~\cite{zhang2018unreasonable}. We also evaluated the inference speed of a panoramic view in 1520$\times$760 on a single RTX2080 Ti GPU. The average metrics are reported in Table~\ref{tab:comparison}.
From the results in Table~\ref{tab:comparison}, our 360Roam achieves the comparably best quality while the render speed is two thousand times faster than NeRF~\cite{mildenhall2020nerf} reaching real-time rendering. 
If we decompose the global radiance field uniformly and use a similar number of networks, the 360Roam-u degrades the significant performance. Another method relying on uniform decomposition, KiloNeRF~\cite{reiser2021kilonerf}, suffers the same phenomenon. This is because the layouts of real indoor scenes are more complicated and the density distribution of the radiance field is highly imbalanced. Uniform decomposition leads to the capacity of neural perceptrons wasted in insignificant regions. Obviously, the proposed adaptive slimming method made the model more stable. 
When there is no exploit of geometry to fine-tune the radiance field, 360NeRF performs similarly to the original NeRF.
Furthermore, with the increasing use of tiny MLPs, the rendering quality of the system is enhanced.
As a method focusing on object-centric or small-scale scenes, NSVF~\cite{liu2020neural} cannot converge on large-scale scenes. But we would like to note that the results of NSVF could possibly be improved by providing the depth to explicitly shape the neural voxels or using more powerful GPUs during training.
As the state-of-the-art NeRF method, Mip-NeRF360(*1024) with 1024-channel MLP achieves best reconstruction quality but with a very slow inference time that is over one thousand slower than 360Roam. Quick training approaches instant-NGP and TensoRF fail to reconstruct high-fidelity novel views for large-scale scenes with improved inference speed though slower than 360Roam.


\subsubsection{Qualitative Comparison} 
Visual roaming comparisons between some baselines and our method are displayed in Fig. \ref{fig:comparison}. 
The synthesized results of NeRF and Mip-NeRF360 have complete geometry, but they produce blurry objects in close proximity and a hazy effect in the distance area. Instant-NGP fails to reconstruct the accurate geometry of the entire scene, and so do NSVF and TensoRF though unshown here due to page limit. Conversely, KiloNeRF generates clearer details of the scene but parts of the reflective areas are replaced by black holes due to incorrect modeling.
Our method 360Roam outperforms other methods and the novel-view images always have more crystal fidelity of the texture at any distance. 360Roam is capable of rendering a proper view-dependent effect for non-Lambertian materials, e.g., the monitor on the first two rows of Fig.~\ref{fig:comparison}.
It proves that properly decomposing the single radiance field into a set of local radiance fields can exploit the capacity of positional encoding and neural networks achieving higher fidelity rendering. 

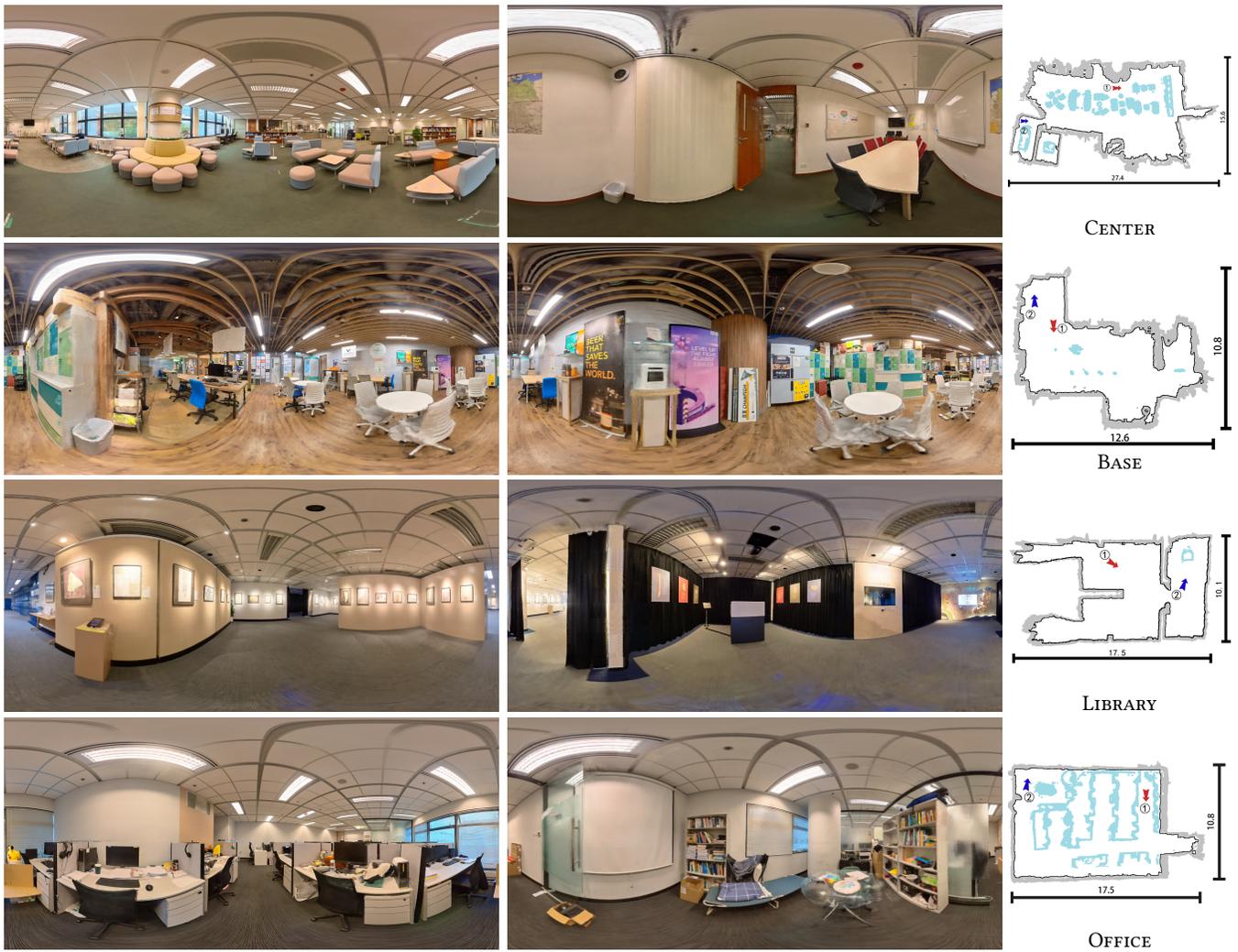
\begin{figure*}
	\centering
	\begin{tikzpicture}
		\node[inner sep=0pt] () at (-10.2,0){\includegraphics[width=0.4\linewidth]{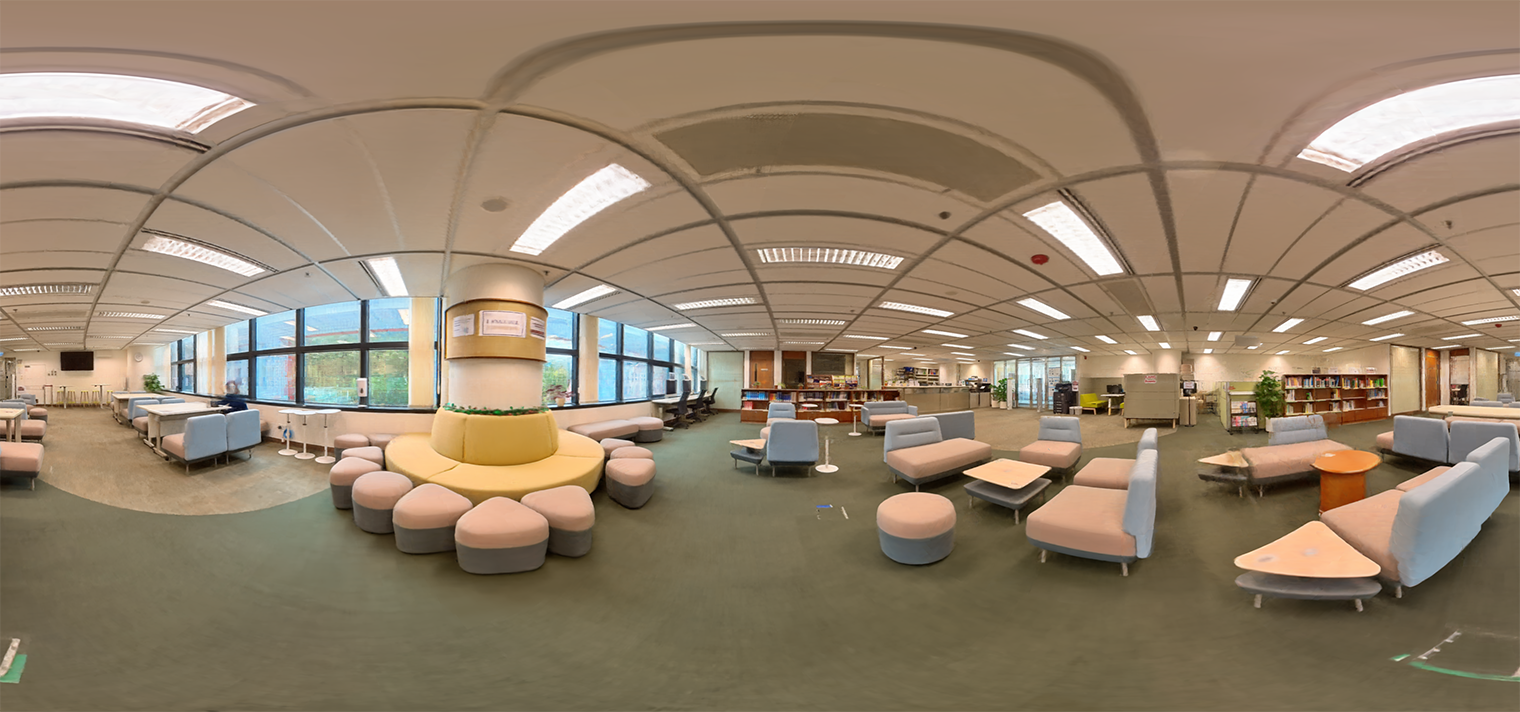}};
		\node[] () at (-2.9,0){\includegraphics[width=0.4\linewidth]{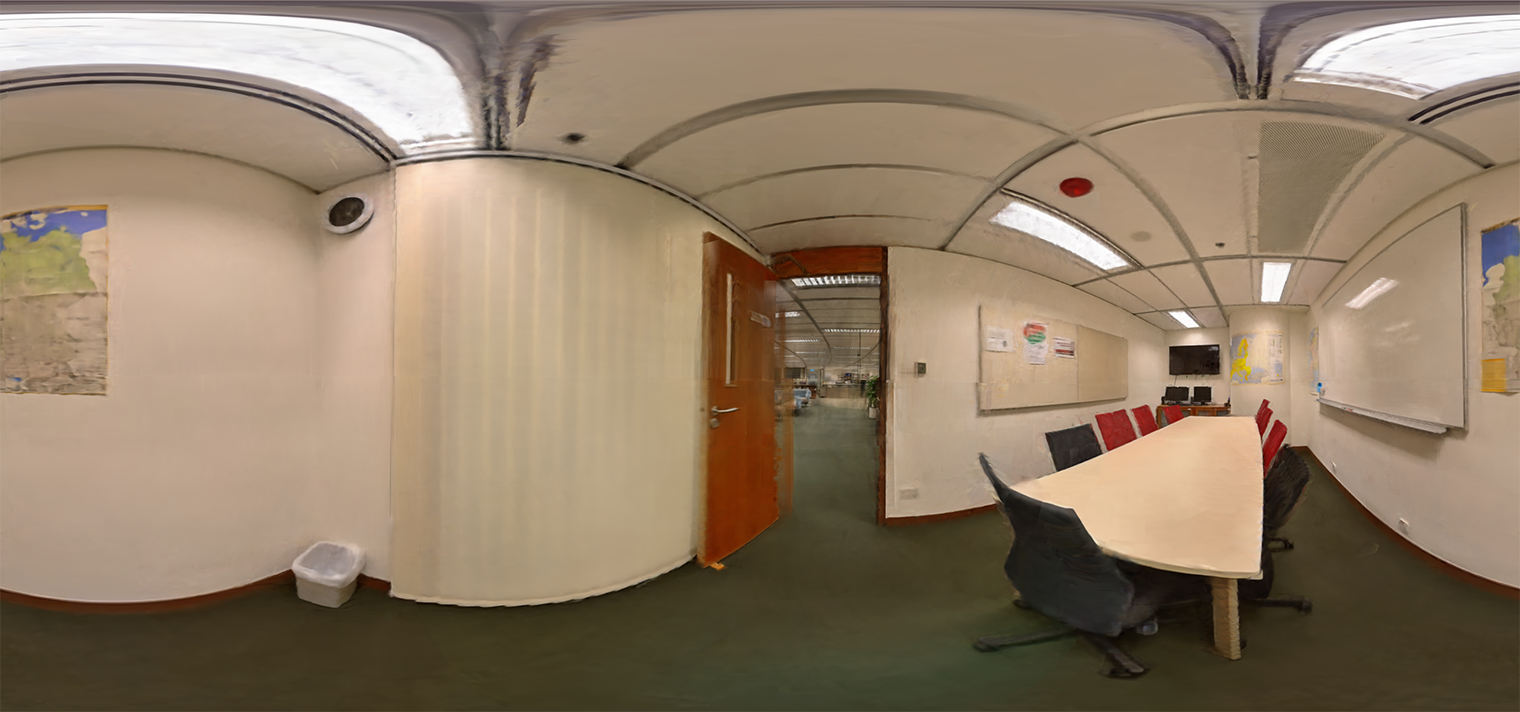}};
		\node[] () at (2.4,0){\includegraphics[width=0.18\linewidth]{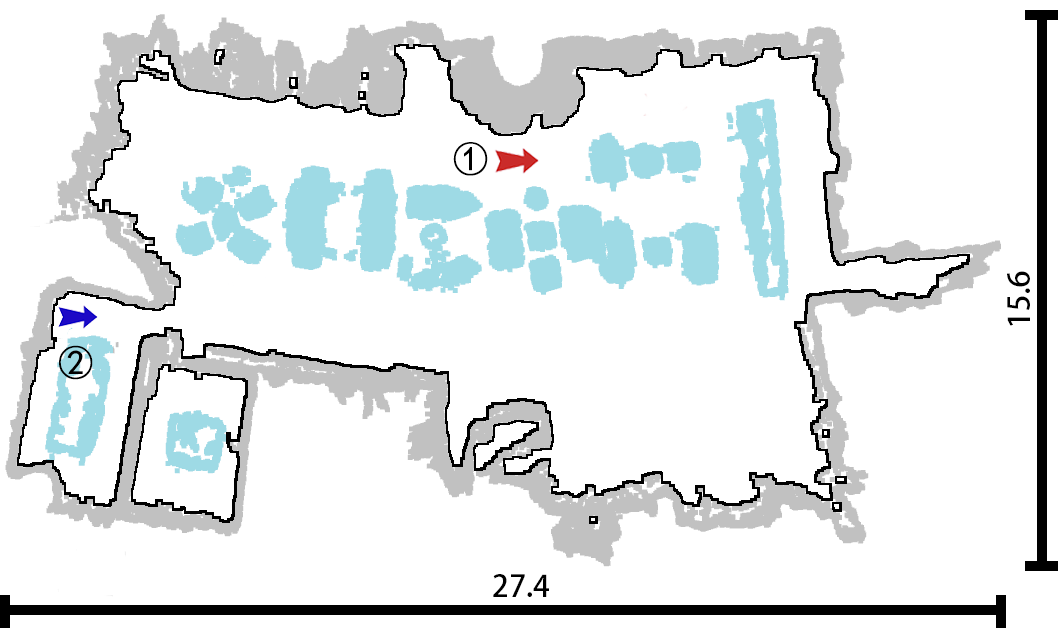}};
		\node[] () at (2.4, -1.55){\textsc{Center}};
		
		\node[inner sep=0pt] () at (-10.2,-3.45){\includegraphics[width=0.4\linewidth]{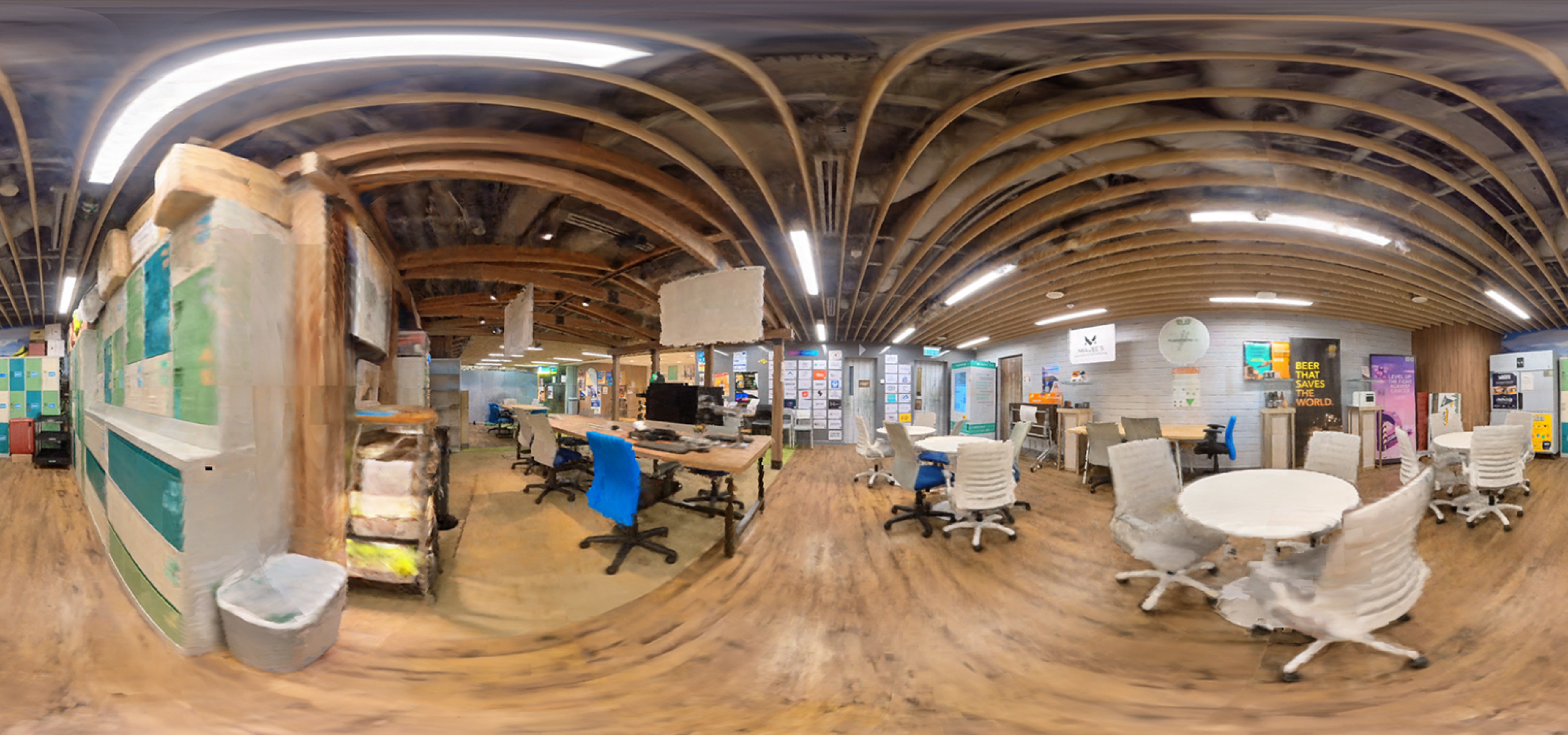}};
		\node[] () at (-2.9,-3.45){\includegraphics[width=0.4\linewidth]{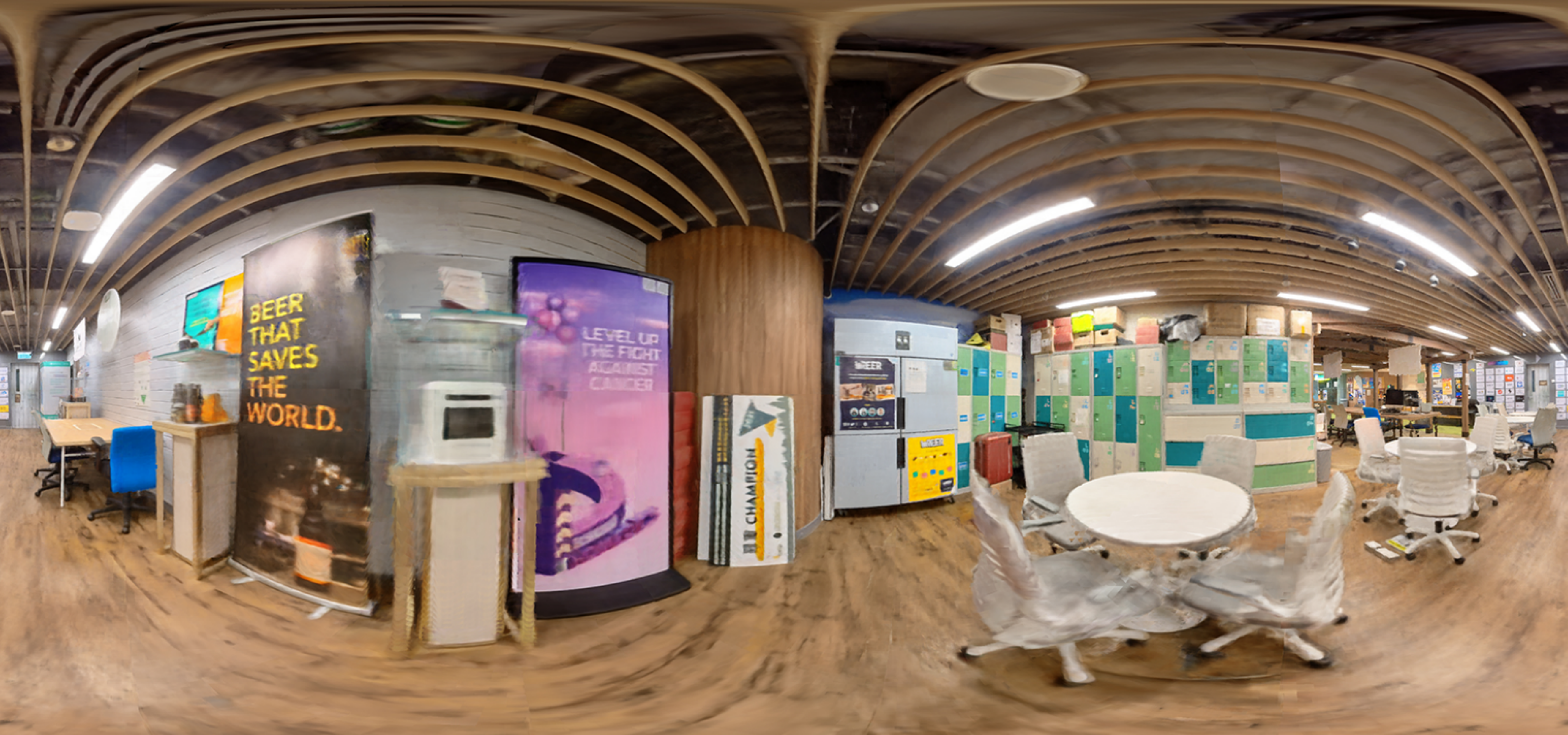}};
		\node[] () at (2.4,-3.45){\includegraphics[width=0.18\linewidth]{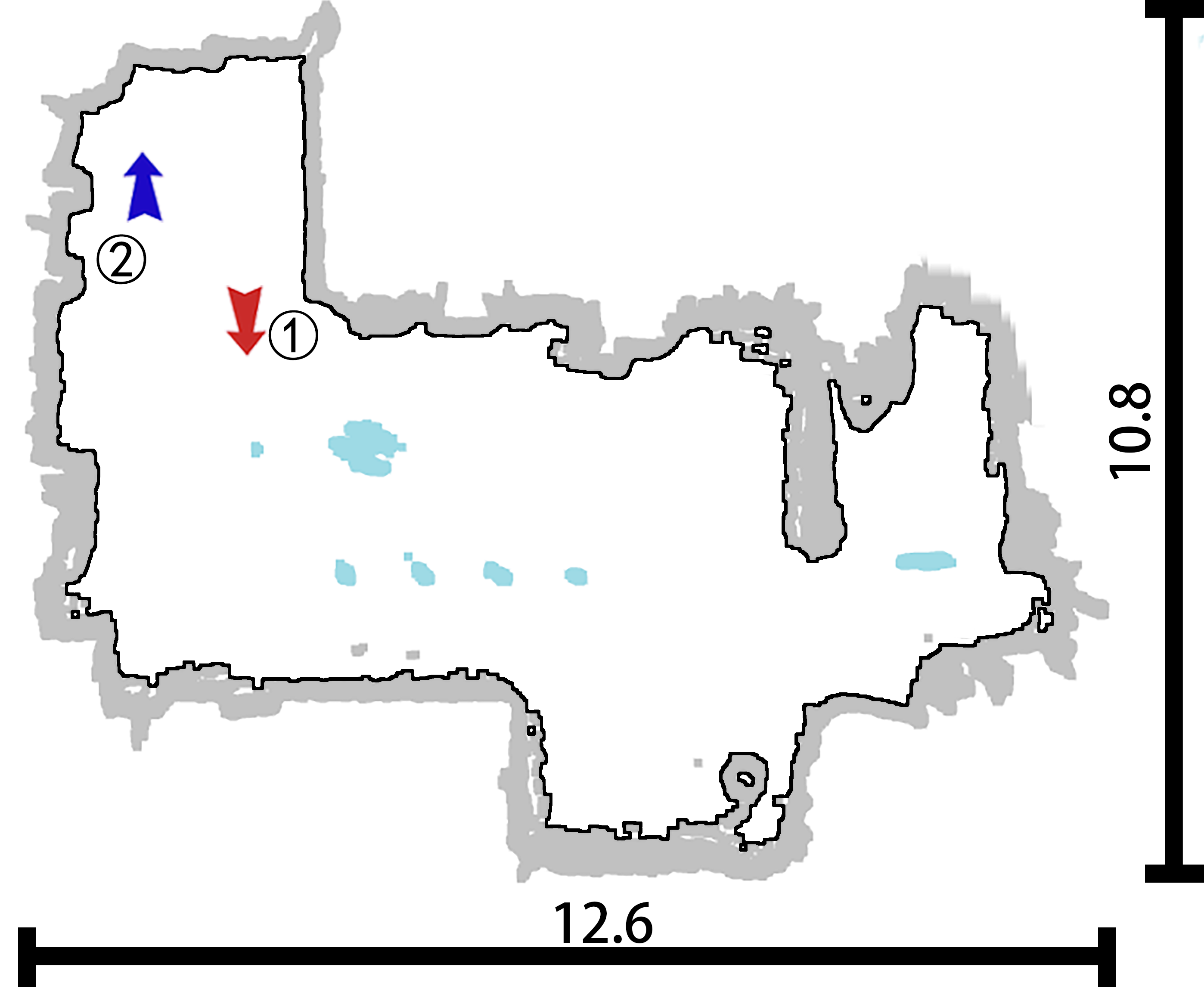}};
		\node[] () at (2.4, -4.95){\textsc{Base}};
		
		\node[inner sep=0pt] () at (-10.2,-6.9){\includegraphics[width=0.4\linewidth]{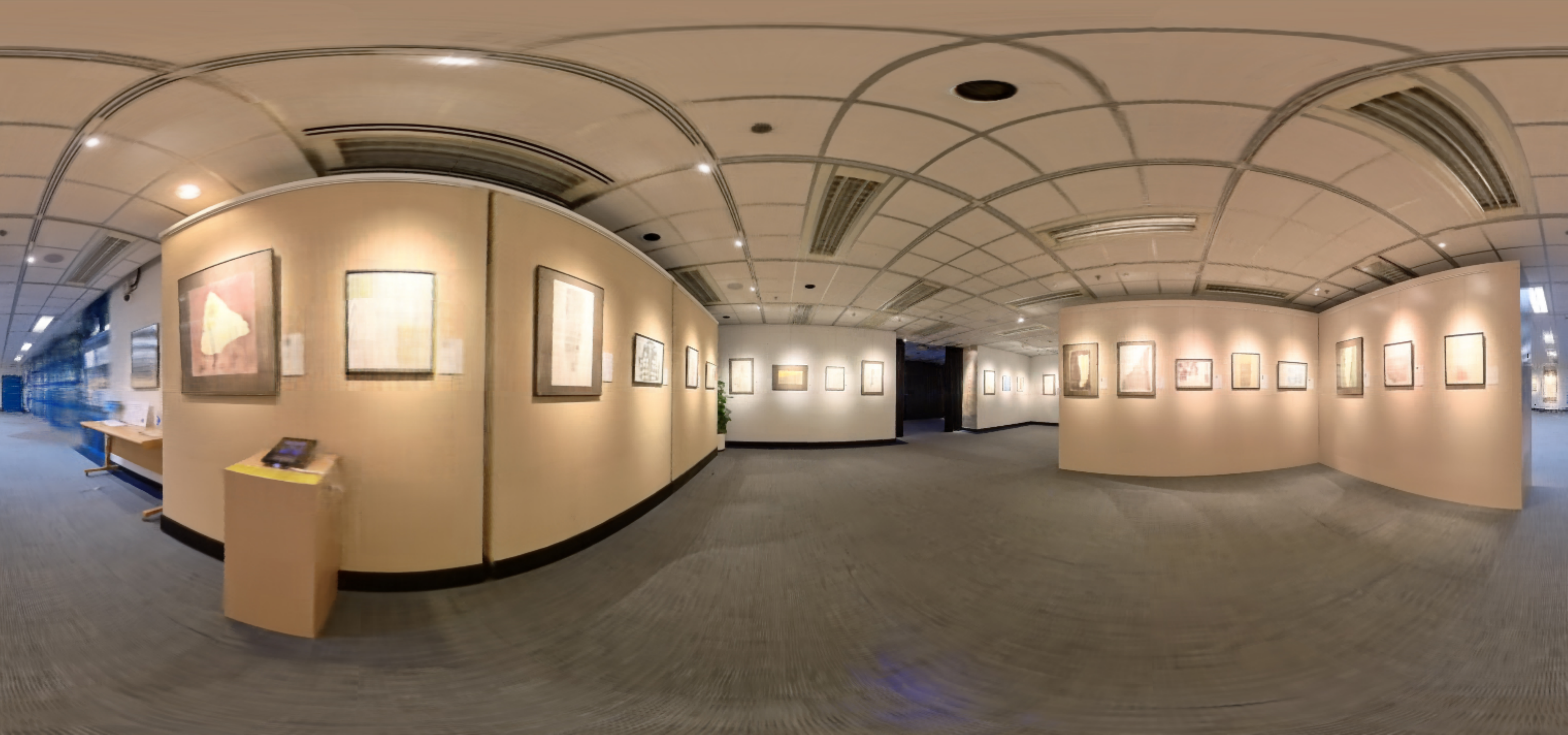}};
		\node[] () at (-2.9,-6.9){\includegraphics[width=0.4\linewidth]{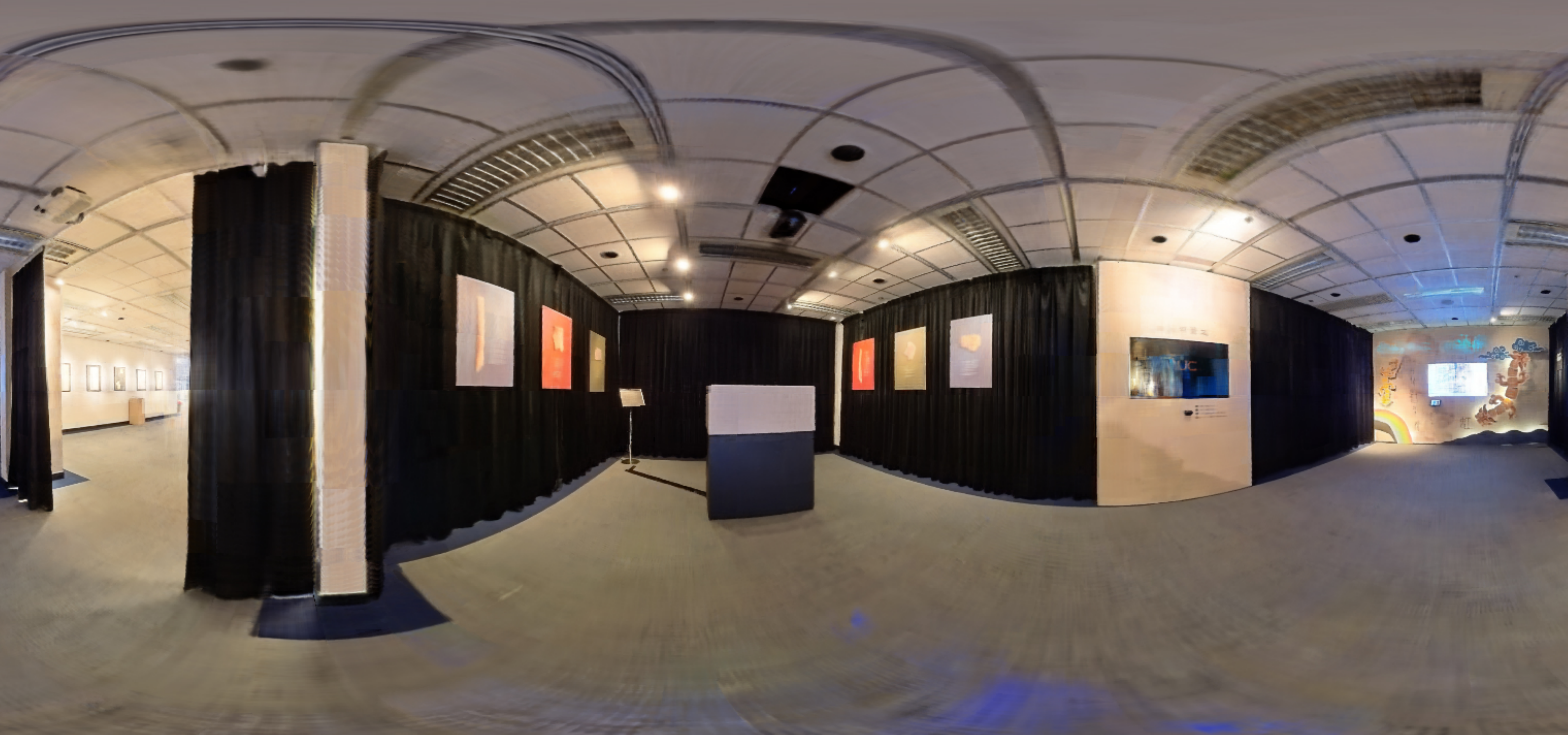}};
		\node[] () at (2.4,-6.9){\includegraphics[width=0.18\linewidth]{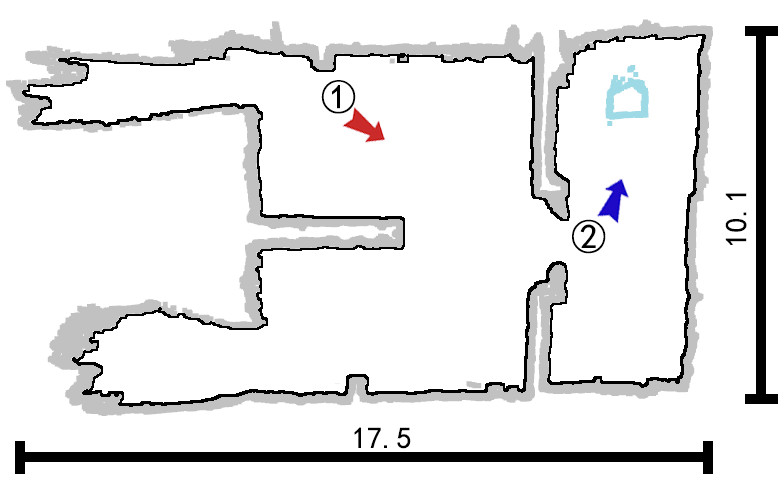}};
		\node[] () at (2.4, -8.45){\textsc{Library}};
		
		\node[inner sep=0pt] () at (-10.2,-10.35){\includegraphics[width=0.4\linewidth]{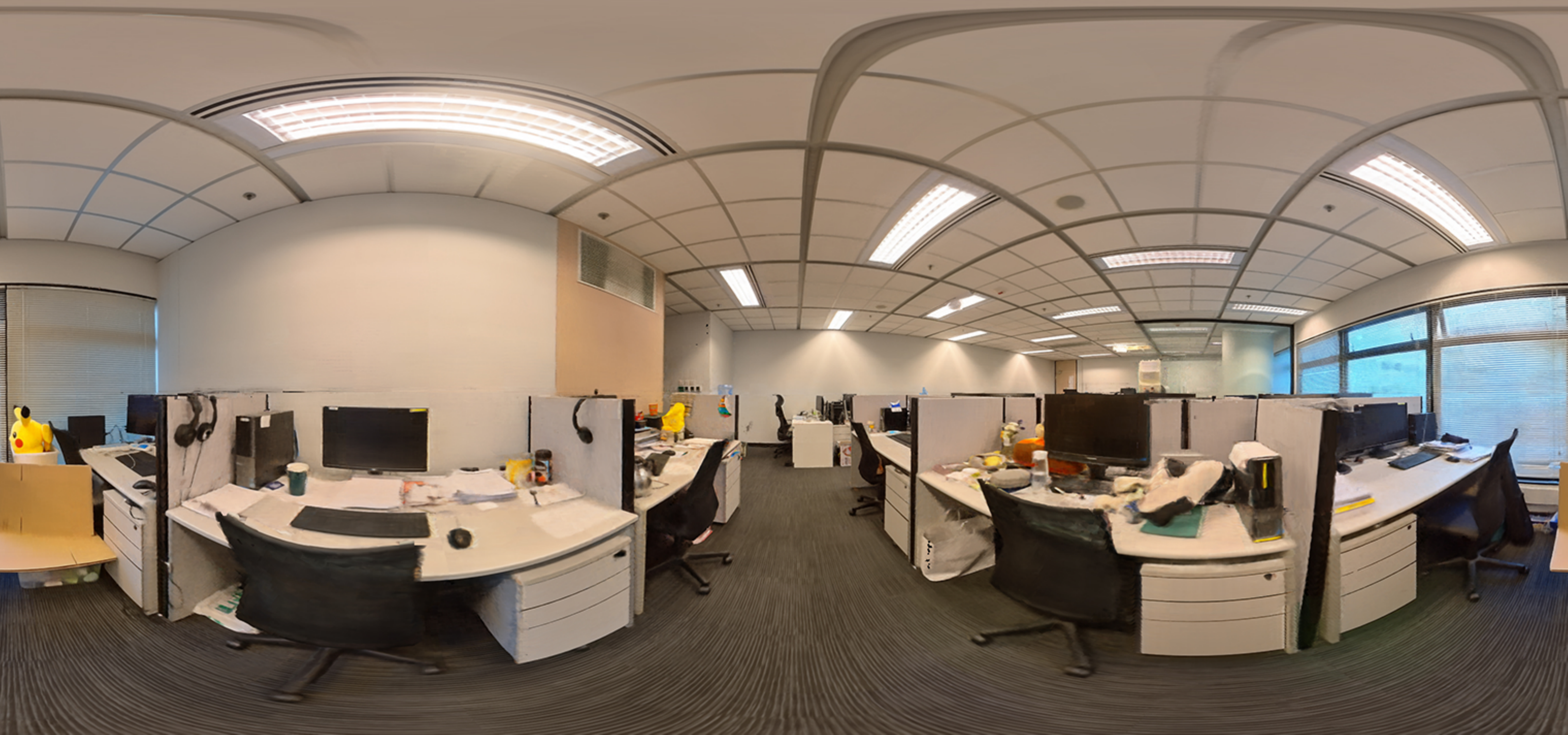}};
		\node[] () at (-2.9,-10.35){\includegraphics[width=0.4\linewidth]{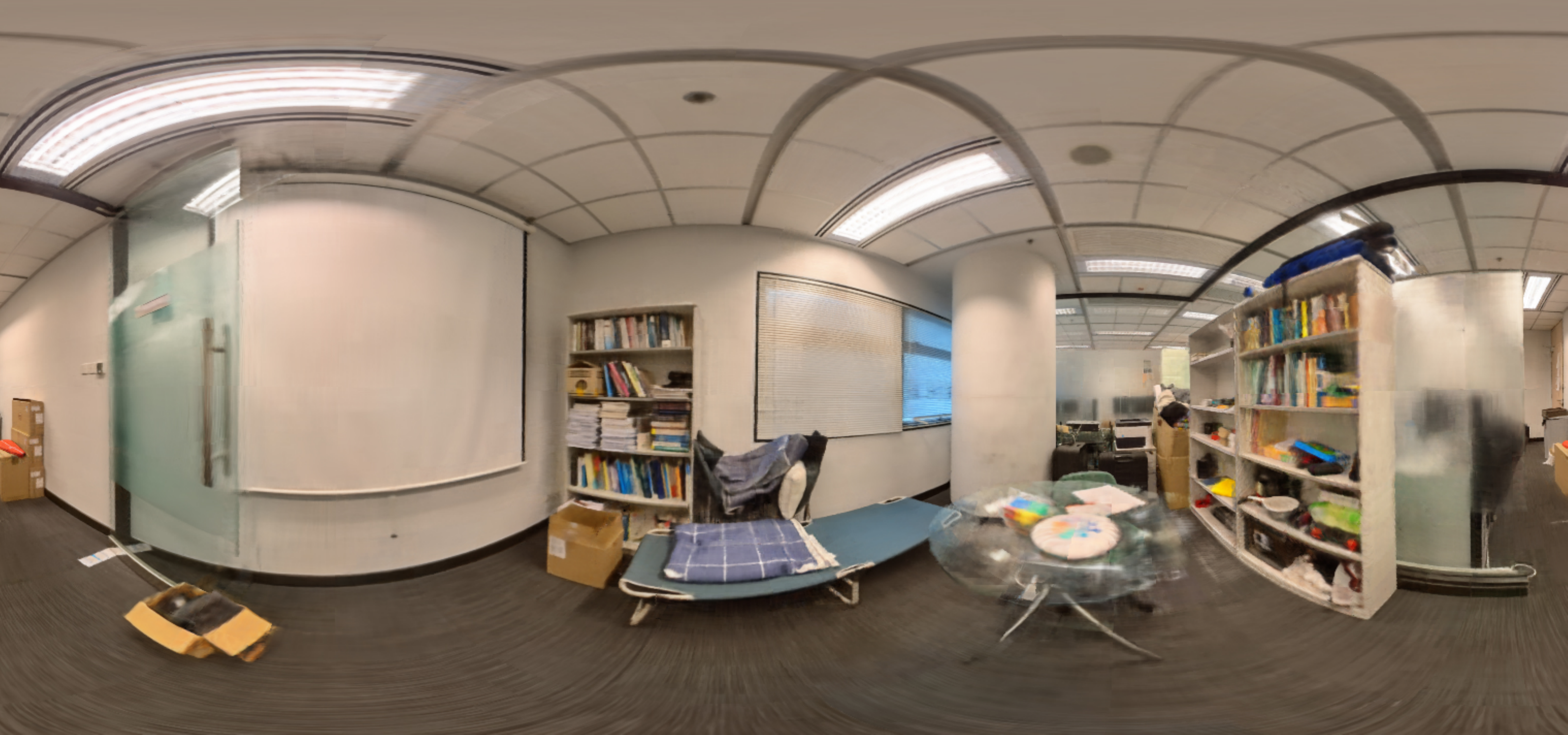}};
		\node[] () at (2.4,-10.35){\includegraphics[width=0.18\linewidth]{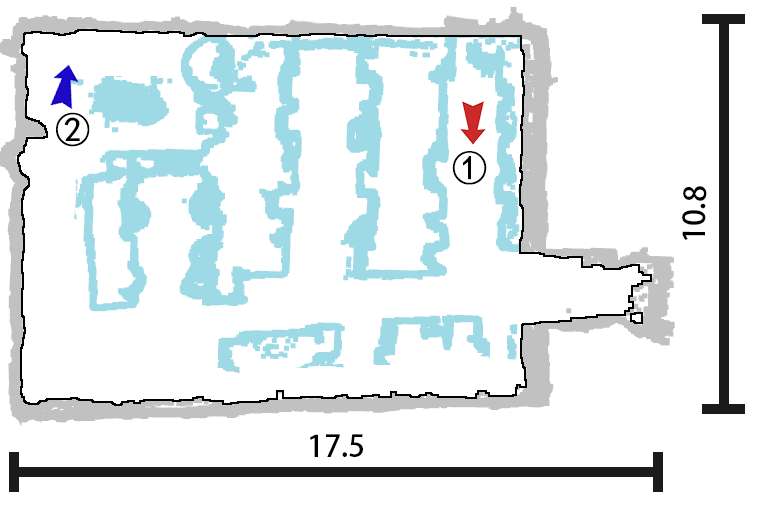}};
		\node[] () at (2.4, -11.9){\textsc{Office}};
		
	\end{tikzpicture} 
	\caption{Results of 360Roam on four real scenes. First two columns are synthesized novel views. The corresponding estimated floorplans are displayed in the rightmost column with circled numbers and arrows. Circled numbers indicate relative locations and in which column the rendered image is located, and arrows represent the viewing directions.}
	\label{fig:results}
\end{figure*}

Fig.~\ref{fig:results} is a gallery of some novel-view $360^\circ$ images of real scenes with their estimated floorplans on the right using our system. We show two novel views for each scene, and the relative locations and viewing directions are marked in the floorplan with circled numbers and arrows.

\begin{figure*}[]
    \def\imgw{0.48}
    \captionsetup[subfigure]{skip=1pt, justification=centering}
    \centering
    \subfloat[Mip-NeRF360(*1024) \\\footnotesize{$\left\langle PSNR=25.324, SSIM=0.674, LPIPS=0.255 \right\rangle$} ]{\includegraphics[width=\imgw\linewidth]{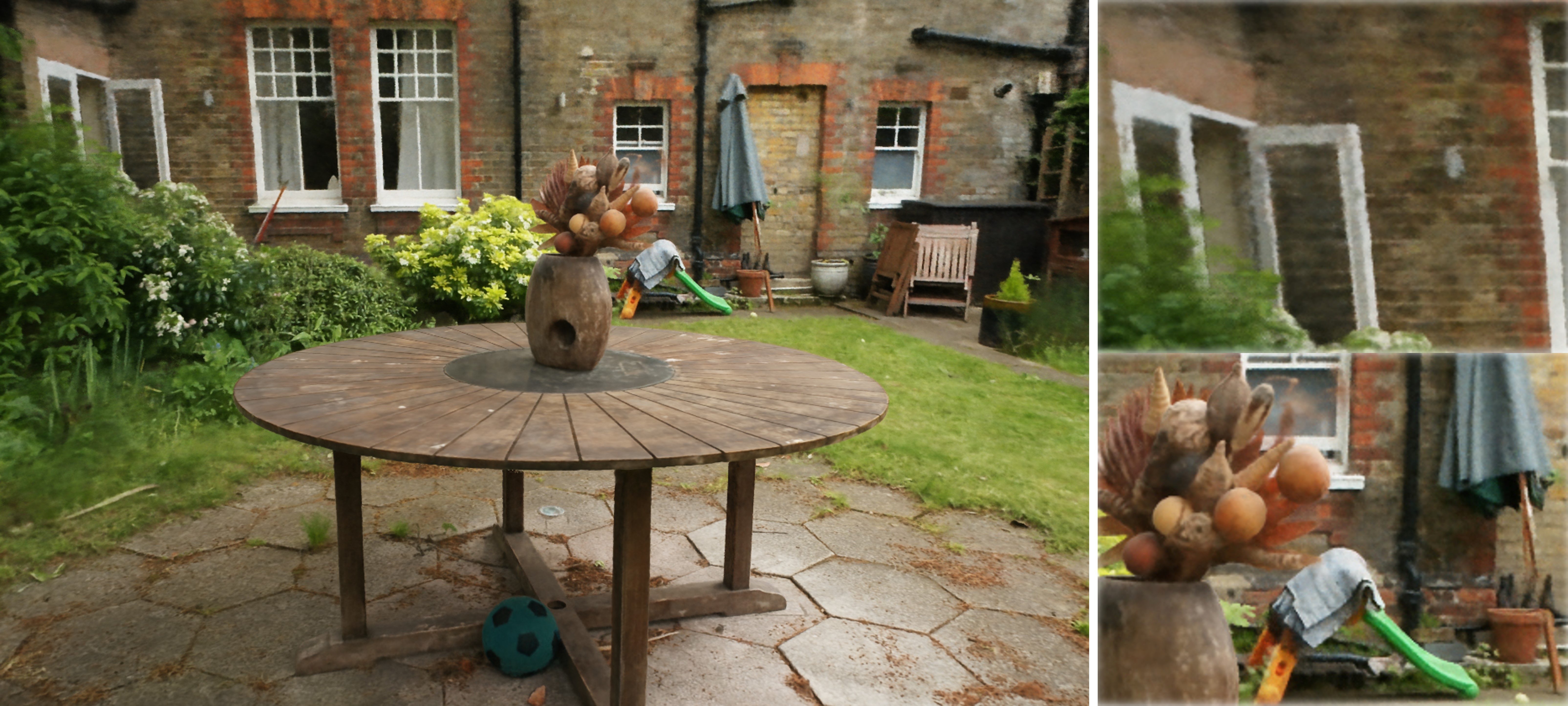}}\,
    \subfloat[360Roam$^\odot$\\\footnotesize{$\left\langle PSNR=25.550, SSIM=0.682, LPIPS=0.248 \right\rangle$} ]{\label{fig:360roam_center}\includegraphics[width=\imgw\linewidth]{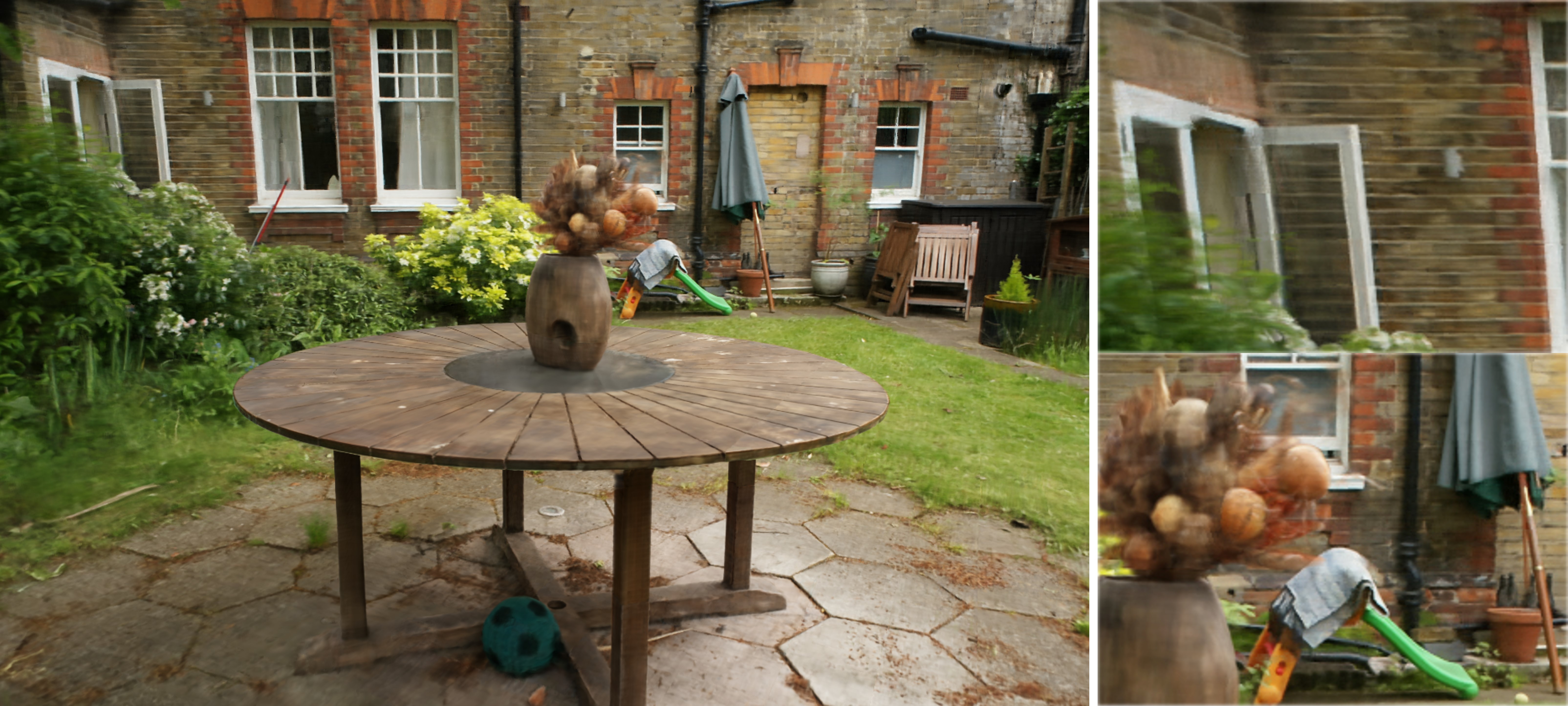}}
    \caption{Compared with Mip-NeRF360(*1024) on the object-centric scene \textsc{Garden} of the Mip-NeRF360 dataset. 
    }
    \label{fig:mip360}
\end{figure*}


\subsection{Entension to Object-centric Scenes}
The scenes we collected are indoor and commonly composed of large spaces and multiple sub-spaces, and the capturing positions of images are sparsely distributed in the scenes. Fully considering the characteristics of the scenes, we proposed an effective pipeline and achieved high rendering quality and fast rendering speed. To further verify the generalization of our proposed pipeline, apart from evaluating on our own 360Roam dataset, we also conducted additional experiments on Mip-NeRF360 outdoor scenes \textsc{Garden} which are object-centric scenarios. And the results are demonstrated in Fig.~\ref{fig:mip360}. 

Object-centric scenarios are represented by the training and test views rotating 360 degrees around a main object. These scenes are unbounded in all directions while the background would be far away from the foreground object. To enhance the performance in unbounded scenes, some methods~\cite{zhang2020nerf, neff2021donerf, barron2022mip360} explore diverse parameterization to model the foreground and background individually. Similarly, we take the region of interest into account. We further slim the foreground objects and assign more neural networks. 360Roam$^\odot$ then can achieve comparable performance on object-centric scenarios.


\section{Discussion and Future Work}
360Roam exhibits significant potential for VR roaming in expansive indoor real scenes while demanding fewer resources and manpower. Achieving rendering speeds at the level of stereoscopic VR is feasible by employing a more powerful computational machine and increasing the number of tiny networks. Nevertheless, further progress is required to attain superior quality and swifter rendering speeds for stereoscopic VR. Our system inherits certain limitations in panorama scene understanding and NeRF. For instance, we encounter stitching artifacts when dealing with panoramic views captured by consumer-grade 360-degree cameras. Since we use an ideal spherical camera model to describe the projection of the $360^\circ$ camera, the stitching artifacts will affect the rendering quality. Although we can exploit a professional $360^\circ$ camera for capturing, it is necessary to take distortion into account and optimize camera intrinsic parameters during training.
Akin to NeRF, 360Roam necessitates extensive training time and lacks the ability to generalize across arbitrary scenes. Our virtual roaming system exclusively supports the observation of static scenes devoid of dynamic objects. These aspects present intriguing directions for future research and enhancements in virtual roaming systems.

\section{Conclusion}

In this paper, we seek to extend neural radiance fields to handle large indoor scenes while achieving real-time performance for novel view synthesis. To accomplish this, we introduce a novel neural rendering pipeline called 360Roam. 360Roam takes $360^\circ$ images as input to initially learn an omnidirectional radiance field. Subsequently, the omnidirectional radiance field is further slimmed into multiple local radiance fields according to the estimated probabilistic occupancy map, followed by fine-tuning of these local radiance fields. Adaptively decomposing and sampling enables the system to make use of the capacity of positional encoding and neural networks. As a result, the geometry-aware 360$^\circ$ radiance fields not only accelerate the rendering speed but also improve the overall rendering quality. Our system can maintain fidelity even when the perspective of view undergoes a significant translational motion. With the guidance of a floorplan, our system offers an engaging and immersive indoor roaming experience that holds promise for a wide range of applications.



%
%
%
%

\bibliographystyle{IEEEtran}
\bibliography{bibliography}

\appendix
\clearpage

  This document illustrates data collection setup and floorplan details, network details, additional results on small-scale scenes, complete quantitative results, supplementary visual results, and finally more discussions.
  
  First of all, we display data capturing trajectories and device setup for all scenes along with estimated floorplans, and also conducted evaluation on estimated floorplans. Then Section \ref{sec:supp_network_detail} shows more details of network architectures. Next in Section \ref{sec:supp_eval_other_scenes} we applied our 360Roam pipeline to small-scale indoor scenes and our method achieved comparative performance. Additionally, in Section \ref{sec:supp_quan}, we present complete quantitative results of each scene. In Section \ref{sec:supp_qual}, we show more qualitative results, including more ablation study visual results, the panoramic input and some novel-view outputs of each real indoor scene in 360Roam dataset, comparisons of indoor roaming results using our system and other methods in the supplementary video.  Finally, we discuss limitations and future works in Section \ref{sec:supp_discuss}.

\section{Data Collection and Scene Structures} \label{sec:supp_floorplan}
\begin{figure}[!t]
    \centering
    \includegraphics[width=0.5\linewidth]{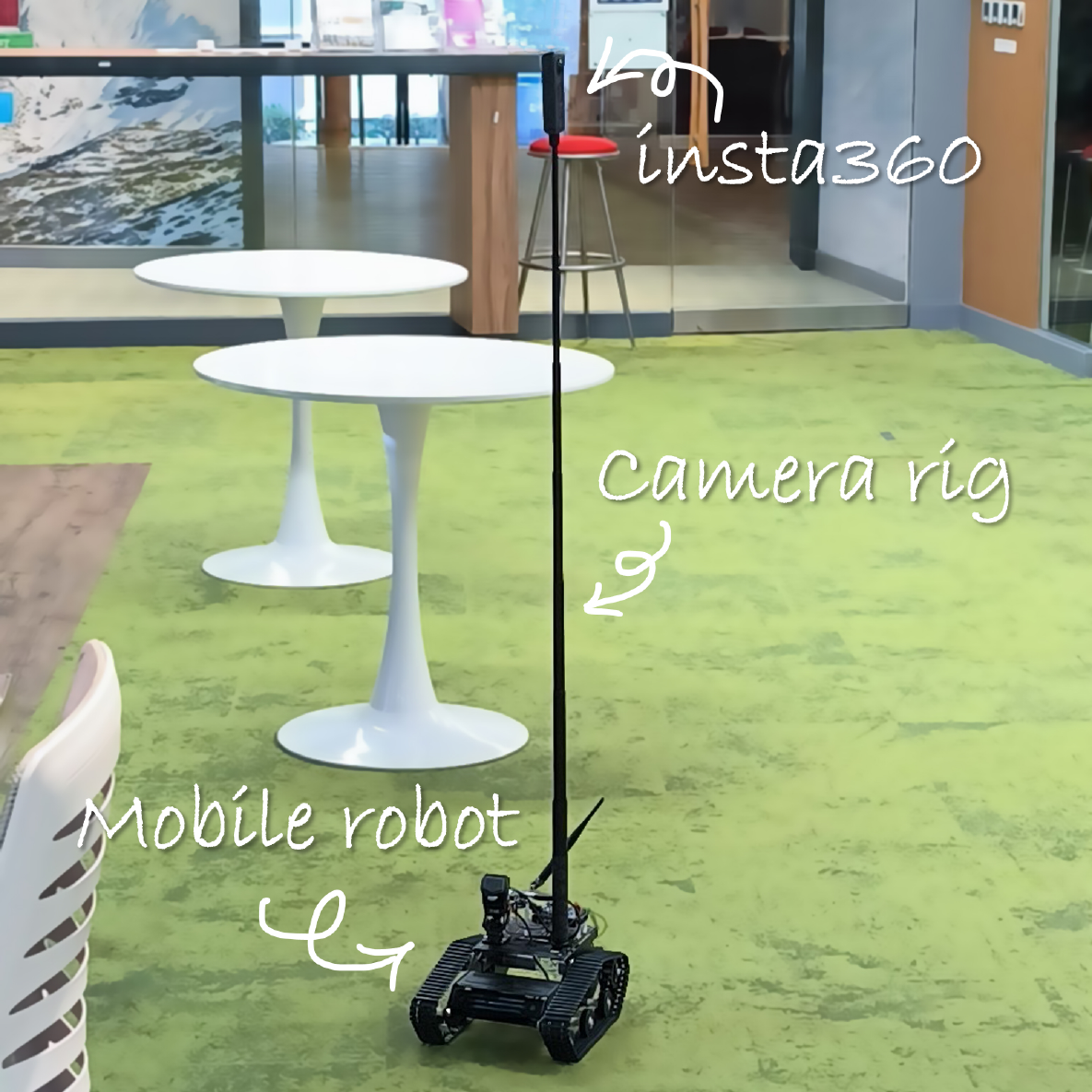}\\  
    \vspace{0.1em}
    \includegraphics[width=0.9\linewidth]{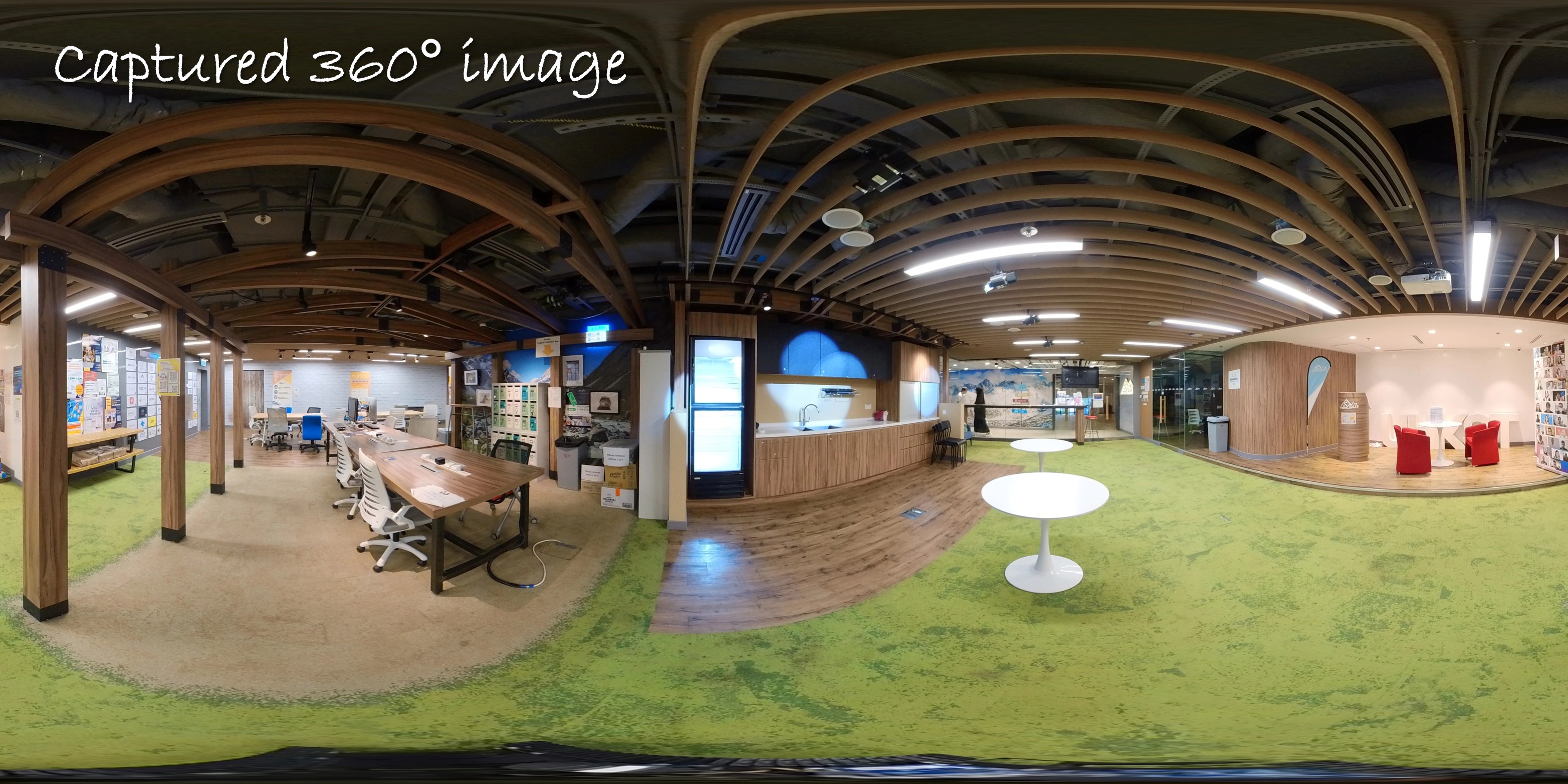} 
    \caption{Data capture and device setting for real scenes. }
    \label{fig:supp_data_collect}
\end{figure}

\begin{figure*} [t!]
    \def\wsc{0.19}
    \centering 
	\subfloat[\textsc{Bar}]{\includegraphics[width=\wsc\linewidth]{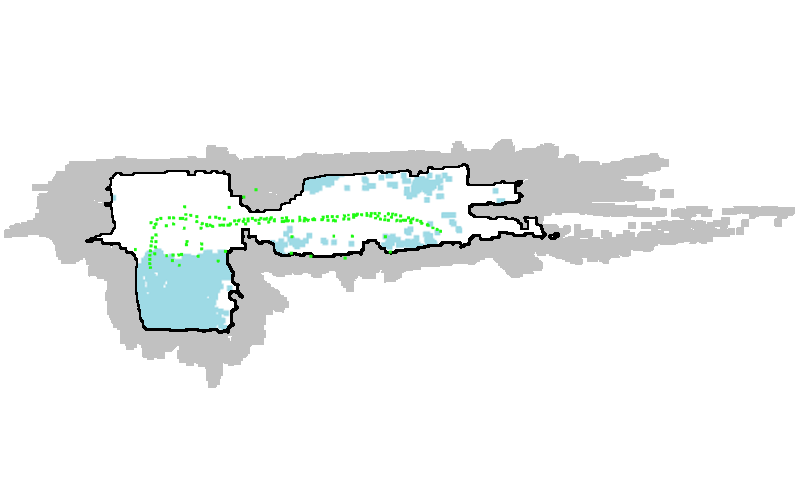}}\, 
    \subfloat[\textsc{Base}]{\includegraphics[width=\wsc\linewidth]{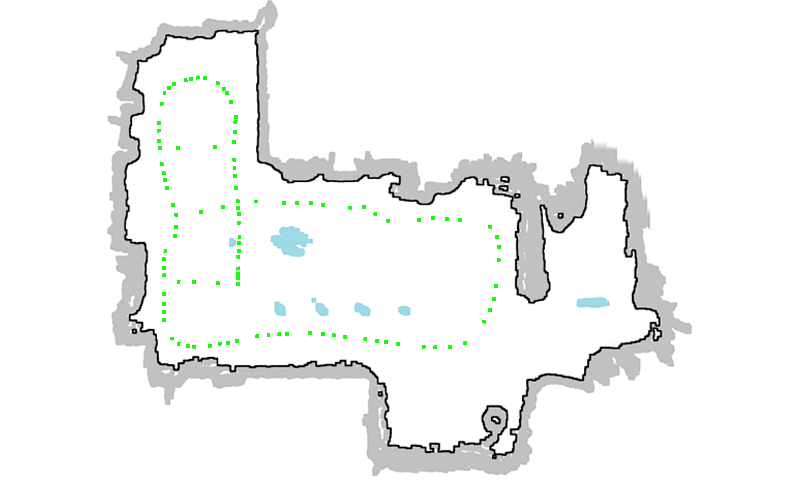}}\,
    \subfloat[\textsc{Cafe}]{\includegraphics[width=\wsc\linewidth]{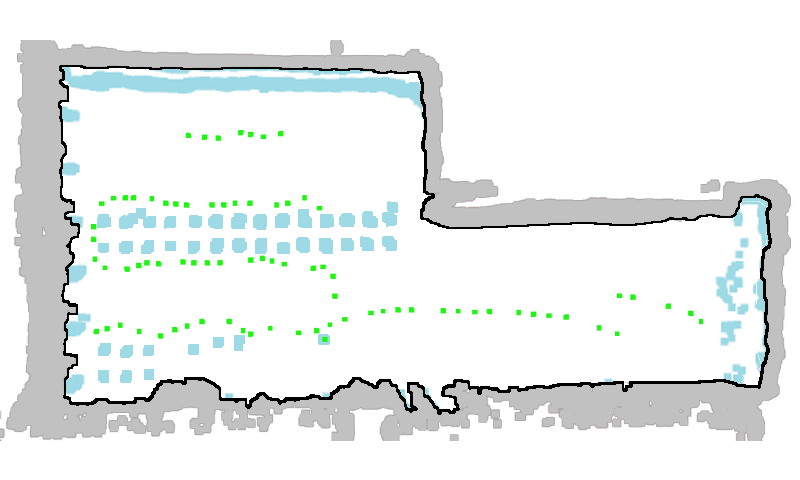}}\,
    \subfloat[\textsc{Canteen}]{\includegraphics[width=\wsc\linewidth]{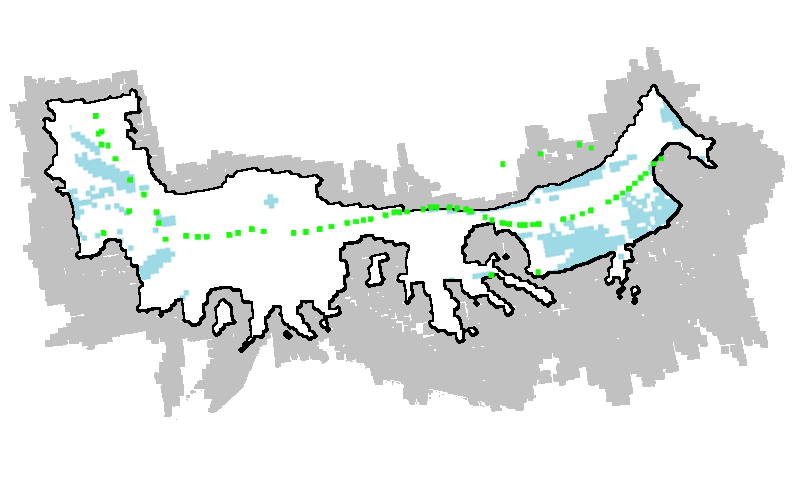}}\,
    \subfloat[\textsc{Center}]{\includegraphics[width=\wsc\linewidth]{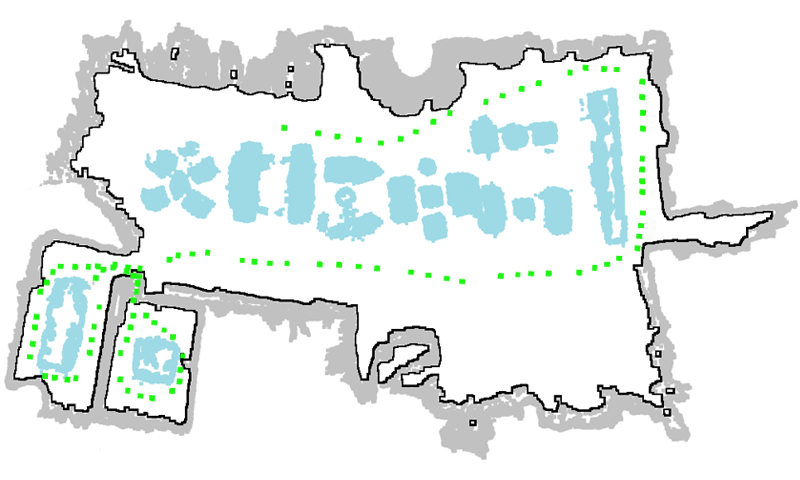}}\\
    \subfloat[\textsc{Corridor}]{\includegraphics[width=\wsc\linewidth]{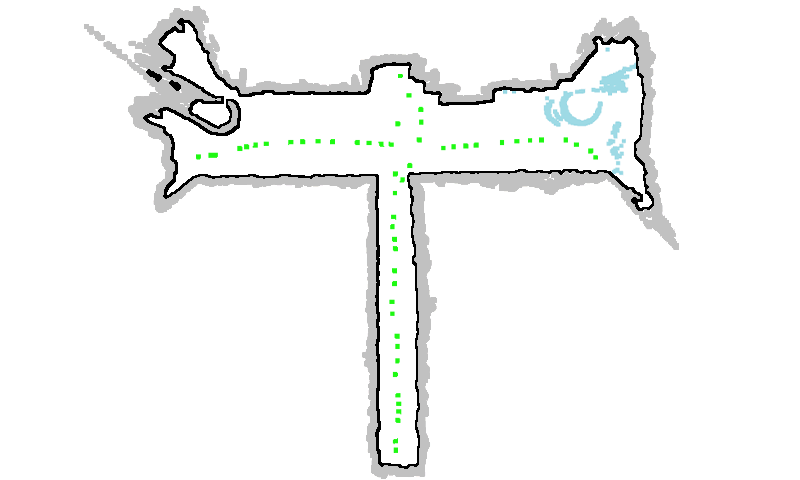}}\,
    \subfloat[\textsc{Innovation}]{\includegraphics[width=\wsc\linewidth]{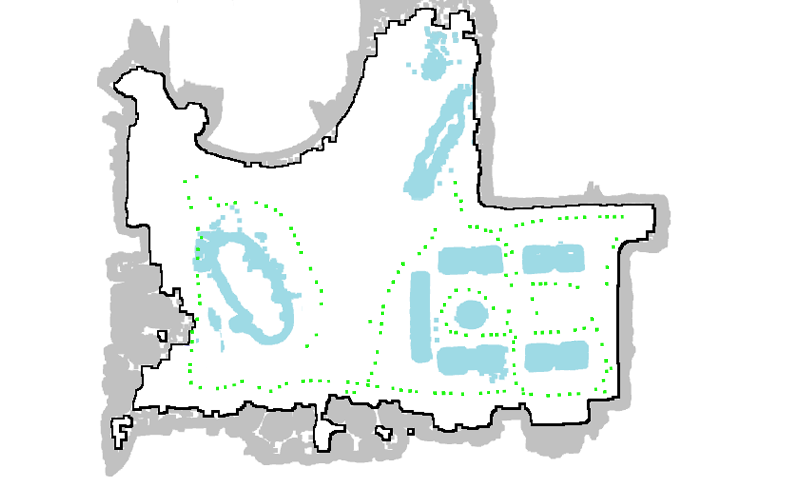}}\,
    \subfloat[\textsc{Lab}]{\includegraphics[width=\wsc\linewidth]{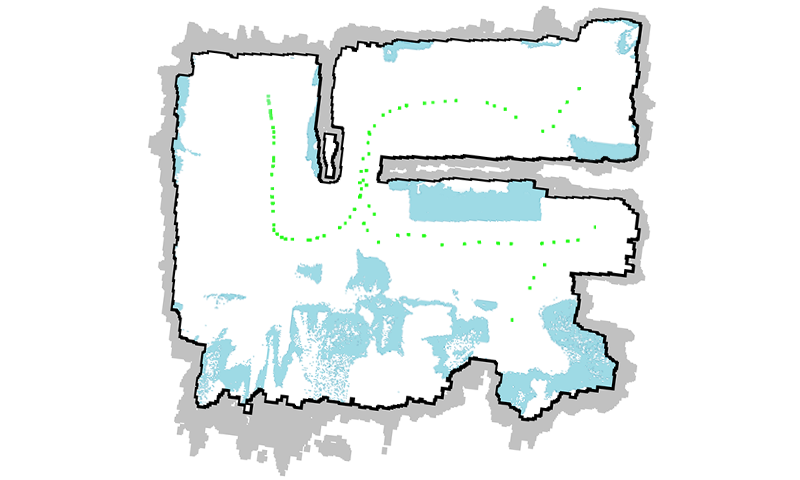}}\,
    \subfloat[\textsc{Library}]{\includegraphics[width=\wsc\linewidth]{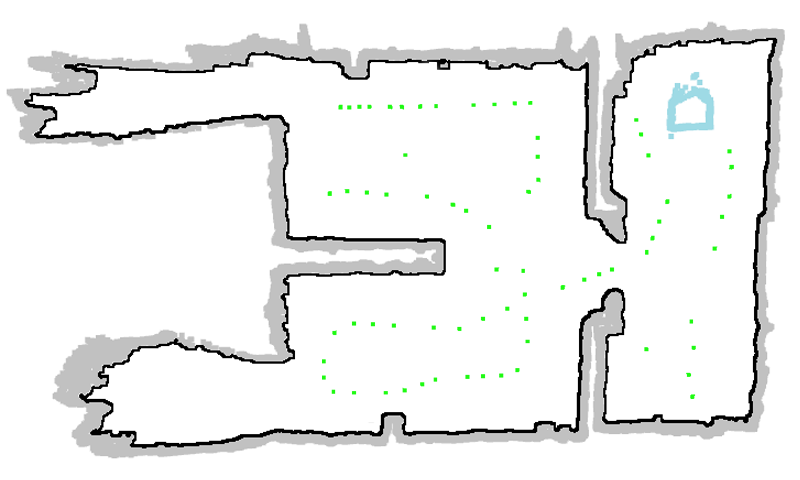}}\,
    \subfloat[\textsc{Office}]{\includegraphics[width=\wsc\linewidth]{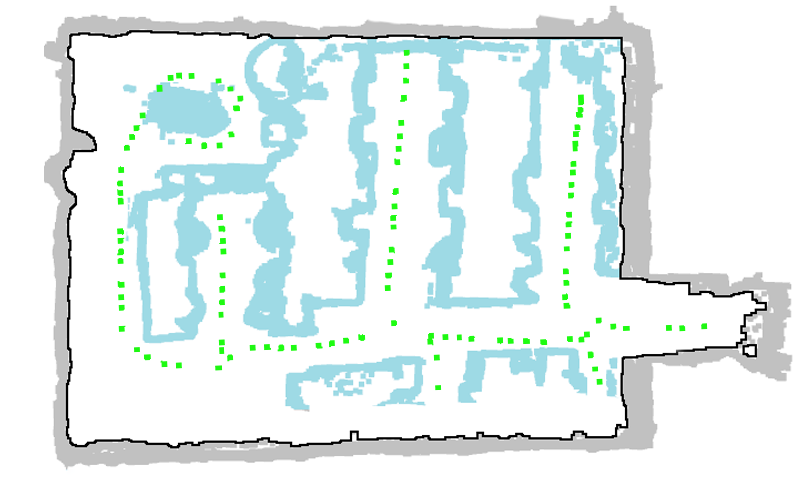}}\\
    \caption{Estimated floorplans of scenes in 360Roam dataset. \textcolor{cyan}{Cyan} blobs indicate indoor items, \textcolor{green}{green} dots represent the camera positions of the training panoramas. }
    \label{fig:supp_floorplan}
\end{figure*}

For real-world data capturing, as shown in Fig~\ref{fig:supp_data_collect}, we set up an `Insta360 ONE X2' 360$^\circ$ camera\footnote{https://www.insta360.com/product/insta360-onex} on the top of a camera rig on a tracked mobile robot with a height of 1 meter. Such a simple and tiny device setup allows flexible walking around indoor scenes. Fig.~\ref{fig:supp_floorplan} visualizes floorplans of scenes in 360Roam dataset. The camera capturing trajectories of training panoramas are visualized as green dots. From visualizations, our scenes cover diverse and complex space structures.

\begin{table}[t]
    \def\wsc{0.35}
    \centering
    \begin{tabular}{c|c}
        Estimated &  Ground truths \\
        \hline
        \includegraphics[width=\wsc\linewidth]{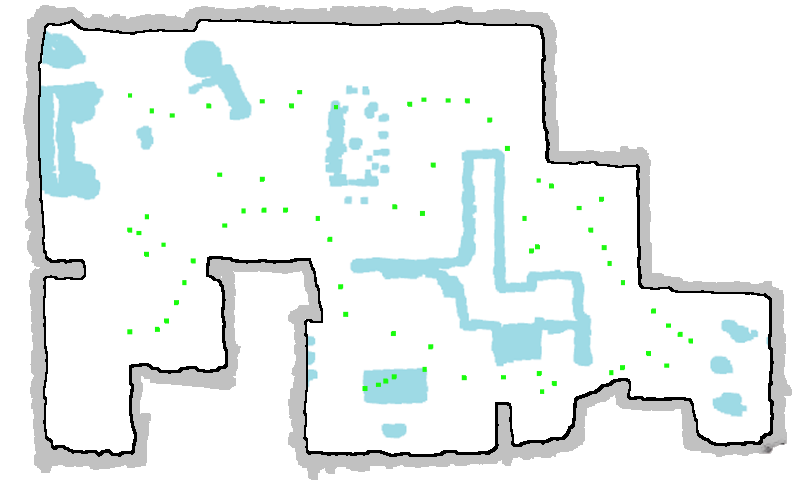}   & 
        \includegraphics[width=\wsc\linewidth]{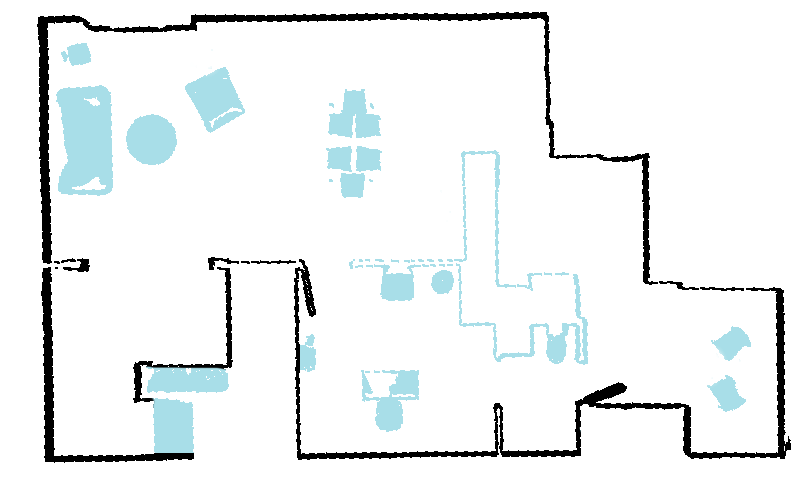} \\
        \includegraphics[width=\wsc\linewidth]{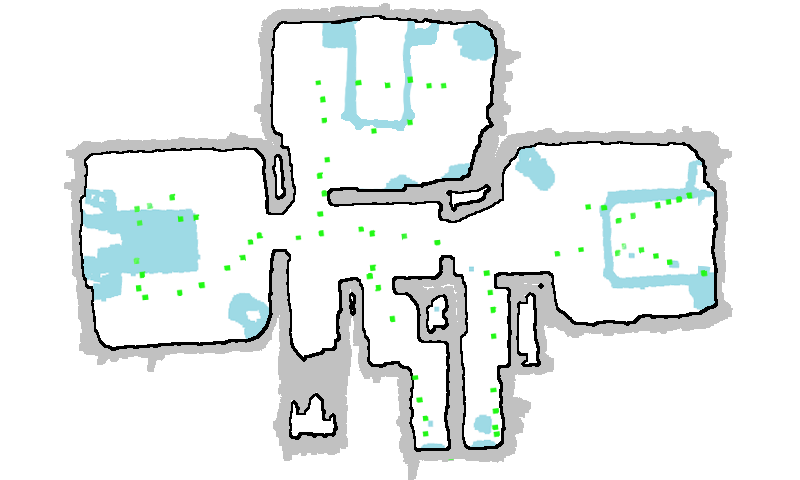} &
        \includegraphics[width=\wsc\linewidth]{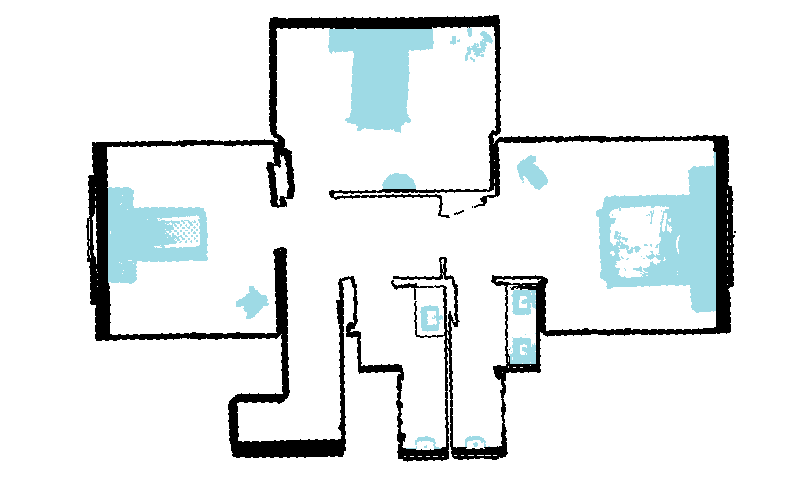} \\
        \includegraphics[width=\wsc\linewidth]{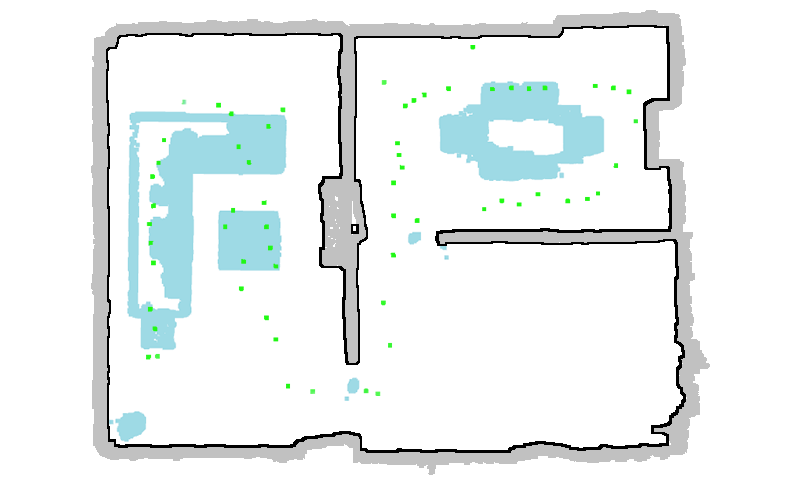} &
        \includegraphics[width=\wsc\linewidth]{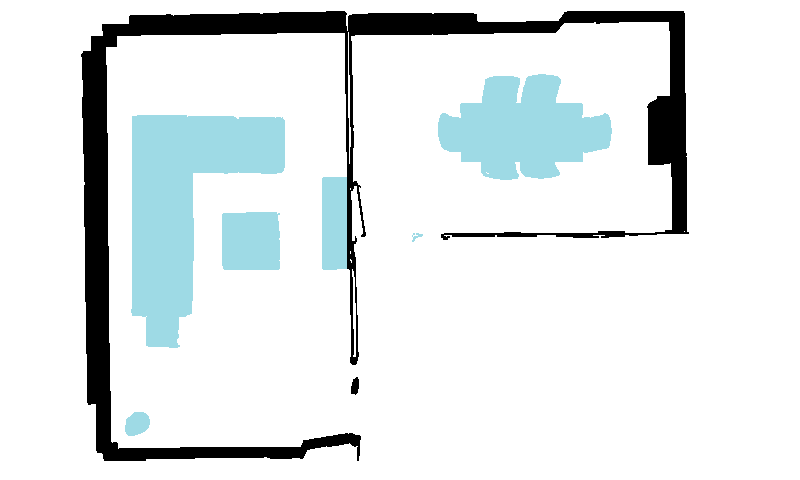} \\    
        \includegraphics[width=\wsc\linewidth]{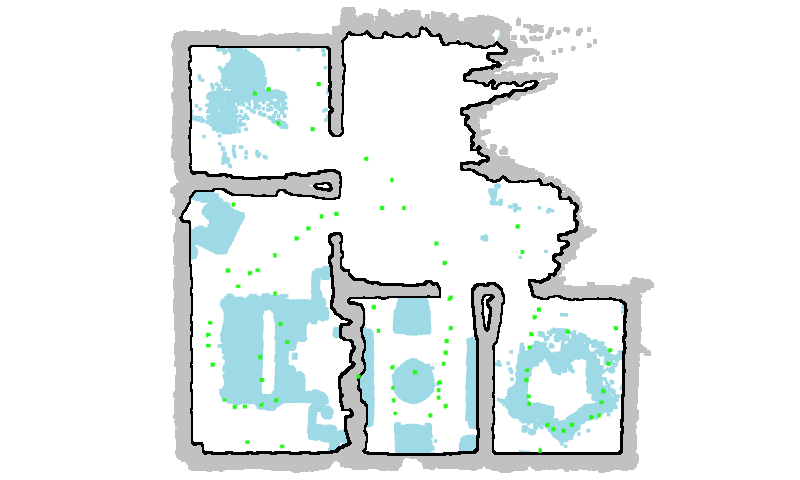} &
        \includegraphics[width=\wsc\linewidth]{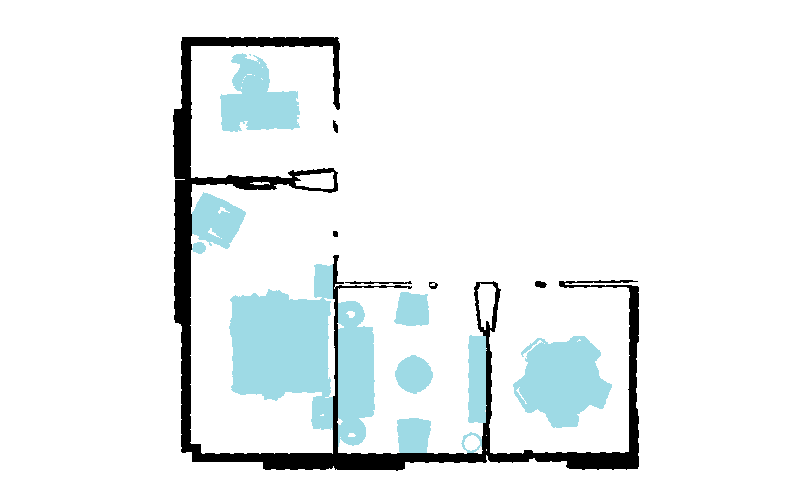} \\
        \includegraphics[width=\wsc\linewidth]{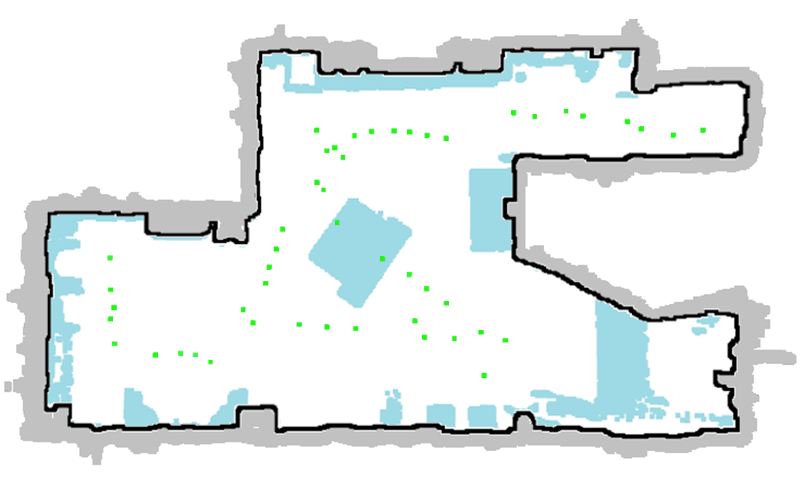} &
        \includegraphics[width=\wsc\linewidth]{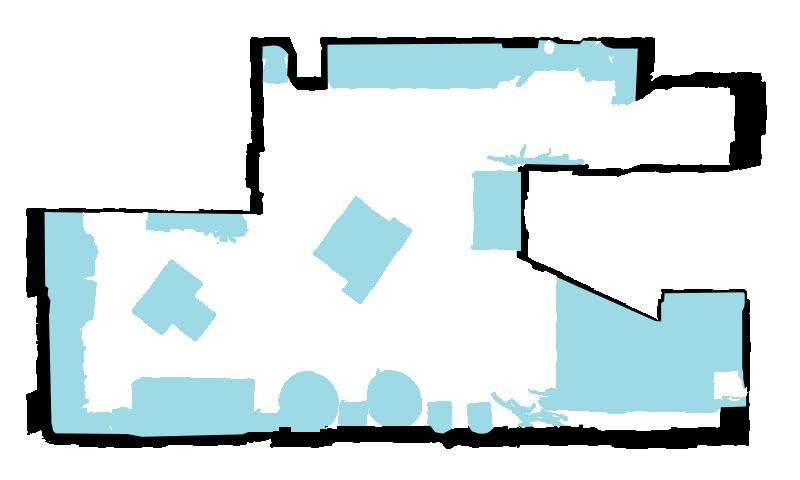} \\
    \end{tabular}
    \captionof{figure}{Estimated floorplans (left column) and ground-truth floorplans extracted from Replica meshes (right column). \textcolor{cyan}{Cyan} blobs indicate indoor items, \textcolor{green}{green} dots represent the camera positions of the training panoramas. }
    \label{fig:supp_floorplan_replica}
\end{table}
\begin{table}[]
	\small
    \setlength{\tabcolsep}{1pt} 
    \def\arraystretch{0.8} 
    \centering
    \begin{tabular}{c|c c c c c c}
        \toprule
         Scene & \textsc{Replica\_1} & \textsc{Replica\_2} & \textsc{Replica\_3} & \textsc{Replica\_4} & \textsc{Replica\_5} & mean\\
         \midrule
         Precise$\uparrow$ & 0.9778 & 0.9493 & 0.9177 & 0.9325 & 0.8734 & 0.9301\\
         Recall$\uparrow$ & 0.9196 & 0.9554 & 0.9670 & 0.9465 & 0.9748 & 0.9527\\
         Acc$\uparrow$ & 0.9103 & 0.9175 & 0.9011 & 0.8941 & 0.8799 & 0.9006\\
         F1$\uparrow$ & 0.9478 & 0.9523 & 0.9417 & 0.9395 & 0.9213 & 0.9405\\
         IoU$\uparrow$ & 0.8296 & 0.8625 & 0.8601 & 0.8734 & 0.7833 & 0.8418\\
         \bottomrule
    \end{tabular}
    \caption{Quantitative evaluation on estimation floorplans of Replica synthetic scenes.}
    \label{tab:supp_floorplan_quan}
\end{table}

\section{Floorplan Evaluation}



\subsection{Estimation Accuracy}
To qualitatively and quantitatively evaluate the estimated 2D floorplans, we employed the synthetic data with ground-truth floorplans. We applied the off-the-shelf Replica~\cite{replica19arxiv} dataset to render images from reconstructed 3D models of indoor spaces and project ground-truth floorplans from the top-down view. The related SDK provides a user interface to generate perspective images. We rendered a cube map containing 6 perspective images at each capturing location and transformed the cube maps into equirectangular images with a resolution of $3840\times 1920$. We have collected 5 sets of synthetic data, named as \textsc{Replica\_1, Replica\_2, Replica\_3, Replica\_4, Replica\_5}. We collected around 80 panoramas per scene for training 360NeRF and estimating floorplans  with on average 66 for training and 16 for testing. 

Fig.~\ref{fig:supp_floorplan_replica} illustrates visual comparisons between estimated and ground-truth floorplans for 5 synthetic scenes from Replica~\cite{replica19arxiv} dataset. By qualitative comparison, our estimated floorplans of Replica resemble ground truths in terms of both scene boundaries and items. 

We also calculated Precise, Recall, Accuracy, F1-score, and IoU (Intersection over Union) between aligned ground-truth and estimated floorplans. We only evaluated the correctness of estimated wall boundaries and interior items within the scene. ``Occupied" pixels such as items and boundaries (cyan and black pixels as shown in Fig.~\ref{fig:supp_floorplan_replica}) are regarded as the positive class, while ``empty" pixels (other pixels such as white and gray pixels in Fig.~\ref{fig:supp_floorplan}) as the negative class. For IoU, we calculated the intersection and union of items and boundaries. Detailed results are reported in Table \ref{tab:supp_floorplan_quan}. On average our estimated floorplans reach over 0.94 F1-score and 0.84 IoU, which illustrates our reliability in floorplan estimation. Our estimated floorplan can work as a dependable visual guide in the roaming system.

\subsection{User Study Detail}
There are 11 participants in the user study.
Before the user study, all participants practiced with a small dummy scene to get familiar with controls in 360Roam system using the mouse to control movement and rotation.
After practice, each participant roamed randomly some of 10 scenes and finished a questionnaire. Each scene was presented with full function or floorplan ablation, and participants can compare different roaming experiences. With the floorplan, users can view the scene map with real-time location and viewing direction indicator, and the floorplan can hinder users from walking through walls or interior items with collision detection.

We evaluated floorplan effects on the roaming system in terms of ``usage preference", ``immersiveness", ``camera clipping". First two metrics were collected from the subjective questionnaire, and the last metric was recorded by auto-detection from the system.
``Usage preference" indicates if users prefer using floorplan. ``Immersiveness" indicates if floorplan assists immersive roaming. 
``Camera clipping" indicates the occurrence of camera clipping through walls or objects.

During roaming, the occurrence of camera clipping in each scene was recorded, and we calculated the average clipping percentage per scene per participant for evaluation.
After roaming all assigned scenes, the participants filled out questionnaires about user experience. There are two main multiple-choice questions with options ``\textit{Yes}", ``\textit{Not Sure}", and ``\textit{No}":
\begin{itemize}
    \item \textit{Q: ``Do you prefer using floorplan?"}
    \item \textit{Q: ``Do you think floorplan can enhance roaming immersiveness?"}
\end{itemize}
Then we accumulated the number of each option and reported the average numbers in percentage form.
In addition, we also collected additional feedback on user experience from participants. The major comments are summarized as follows:
\begin{itemize}
    \item Floorplan shows available subspaces and orients yourself in an unfamiliar complex place.
    \item Floorplan may distract my focus on touring.
    \item Floorplan helps avoid obstacles.
\end{itemize}


\section{Network Architecture Details} \label{sec:supp_network_detail}
Our proposed method is implemented in Pytorch and CUDA. For 360NeRF, we applied a single MLP network detailed in Table \ref{tab:nerf_net} to model the scene. All layers are fully-connected layers with ReLU activation unless specified otherwise. The input position vector is transformed into a positional encoding vector and then passed through 8 layers. A skip connection is used in the fifth layer to preserve information from former layers by concatenating the positional encoding and the output of the fourth layer. The ninth layer outputs a density rectified by a ReLU to guarantee non-negativity and a feature vector. The viewing direction with positional encoding is concatenated with the feature vector to contribute to producing radiance. The final layer is activated by the sigmoid function to restrict the values within range $[0,1]$. The count of network parameters is about 0.596M.
In terms of tiny MLP networks in the final system, the network architecture is detailed in Table \ref{tab:slim_net}. The skip connection is excluded from the network. Compared to the normal MLP, the width and the depth of the tiny network are significantly compressed while the number of network parameters is 6212. 

\begin{table}[t]
    \centering
    \begin{tabular}{c c c c c}
    \hline
    layer   &input & channels & activation \\
    \hline
    pe1 & position vector & 3/60 & - \\        
    layer1  &pe1 & 60/256 & ReLU \\ 
    layer2  &layer1 & 256/256 & ReLU \\
    layer3  &layer2 & 256/256 & ReLU \\
    layer4  &layer3 & 256/256 & ReLU \\
    layer5  &layer4 + pe1 & 256/256 & ReLU \\
    layer6  &layer5 & 256/256 & ReLU \\
    layer7  &layer6 & 256/256 &  ReLU \\
    layer8  &layer7 & 256/256 & - \\
    layer9-density  &layer8 & 256/1 & ReLU \\
    layer9-feature  &layer8 & 256/256 & ReLU \\
    pe2 & viewing direction & 3/24 & - \\     
    layer10  &layer9-feature+pe2 & (256+24)/128 & ReLU \\
    layer11  &layer10 & 128/3 & sigmoid \\
    \hline
    \end{tabular}
    \caption{The MLP network architecture of 360NeRF.}
    \label{tab:nerf_net}
    \vspace{-0.3cm}
\end{table}
\begin{table}[]
    \centering
    \begin{tabular}{c c c c c}
    \hline
    layer    &input & channels & activation \\
    \hline
    pe1 & position vector & 3/60 & - \\        
    layer1  &pe1 & 60/32 & ReLU \\ 
    layer2-density  &layer1 & 32/1 & ReLU \\
    layer2-feature  &layer1 & 32/32 & - \\
    pe2 & viewing direction & 3/24 & - \\     
    layer3  &layer2-feature+pe2 & (32+24)/32 & ReLU \\
    layer4  &layer3 & 32/3 & sigmoid \\
    \hline
    \end{tabular}
    \caption{The MLP network architecture of slimming radiance fields.}
    \label{tab:slim_net}
    \vspace{-0.1cm}
\end{table}

\begin{figure}[t]
    \def\imgw{0.49}
    \captionsetup[subfigure]{skip=0.6pt}
    \def\gaph{0.3em}
    \centering
    \subfloat[Philip et al. \cite{PMGD21}]{\includegraphics[width=\imgw\linewidth]{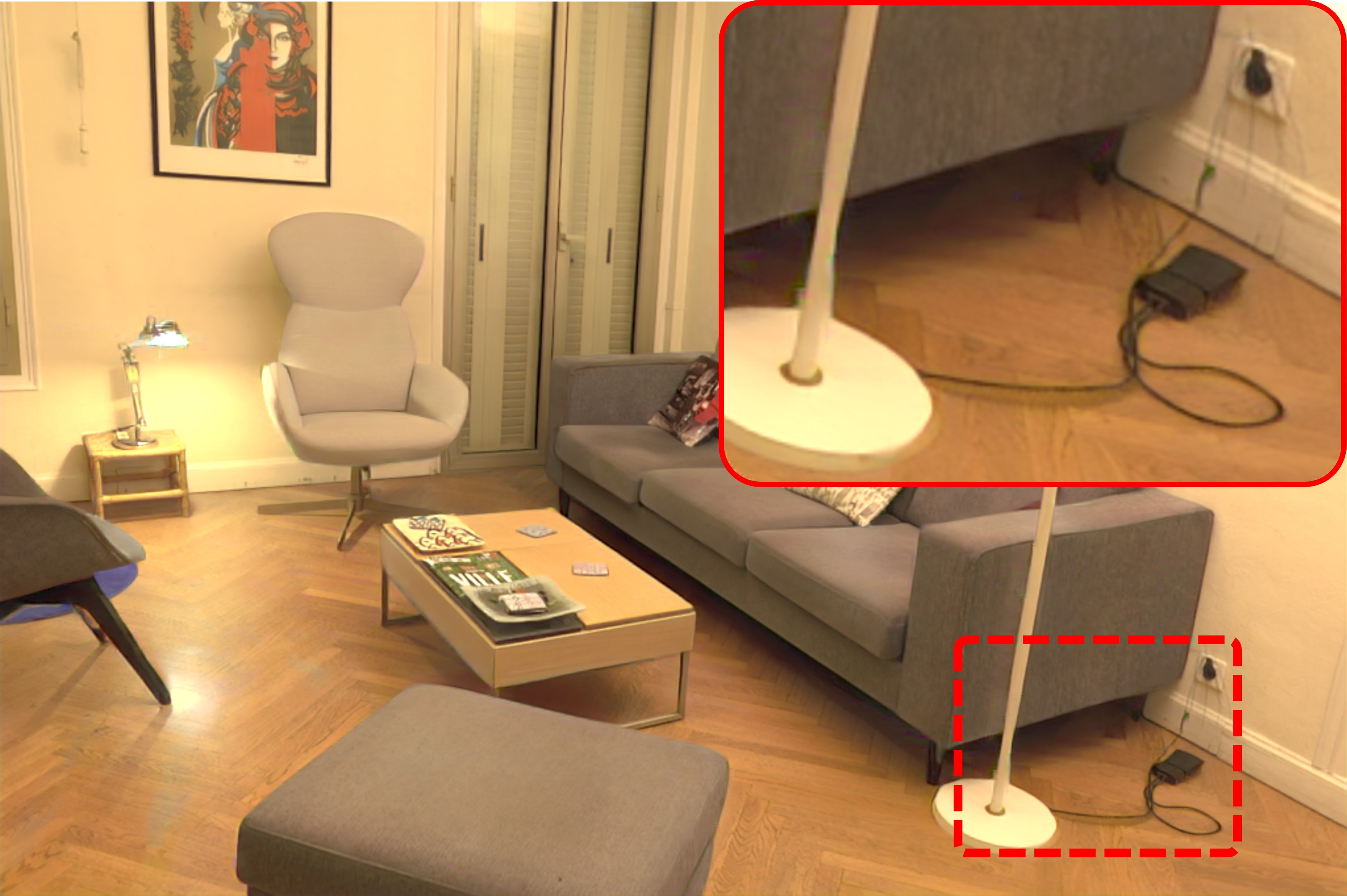}}\,
    \subfloat[Wu et al. \cite{wu2022snisr}]{\includegraphics[width=\imgw\linewidth]{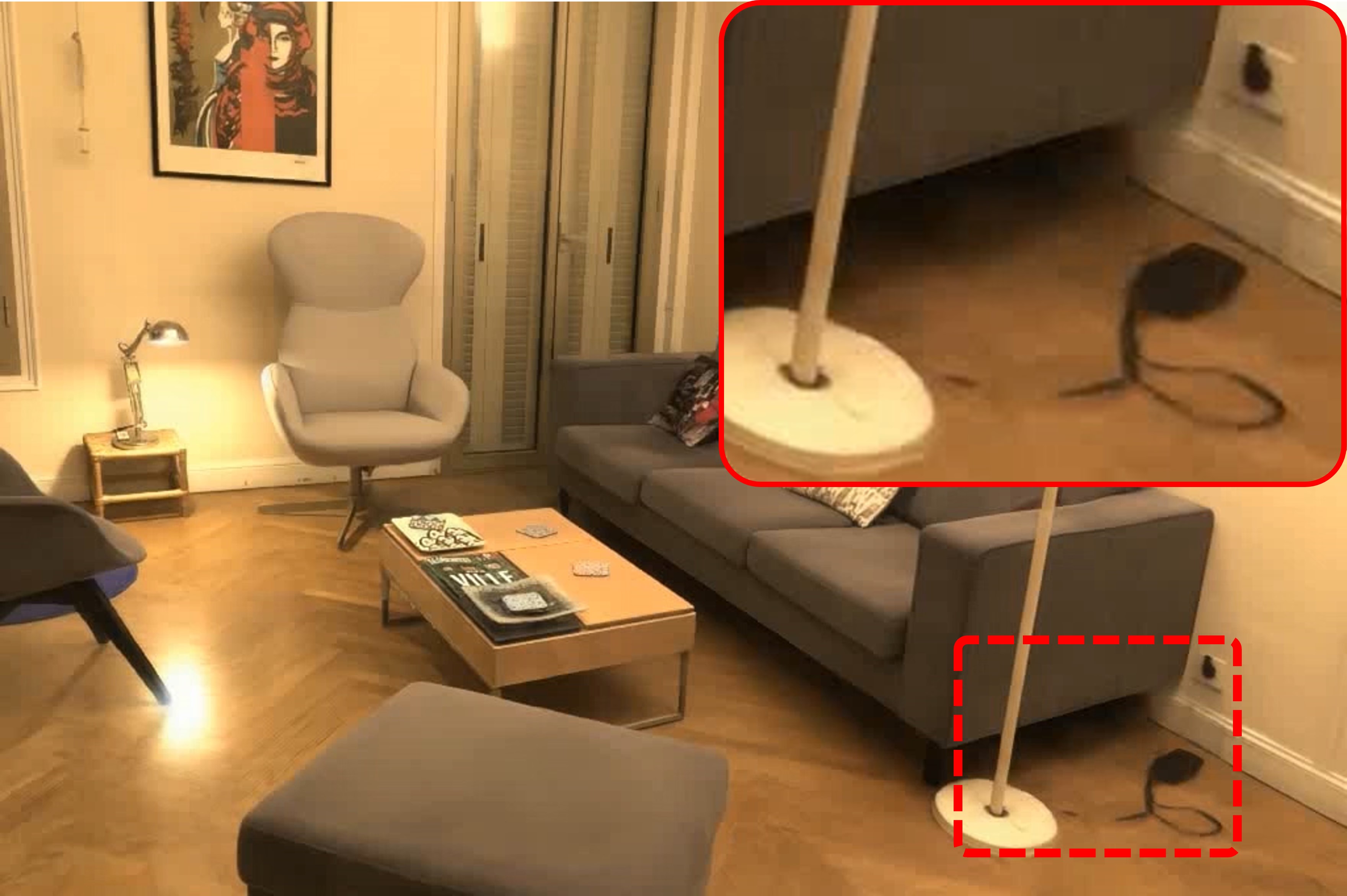}}\\
    \subfloat[360NeRF]{\includegraphics[width=\imgw\linewidth]{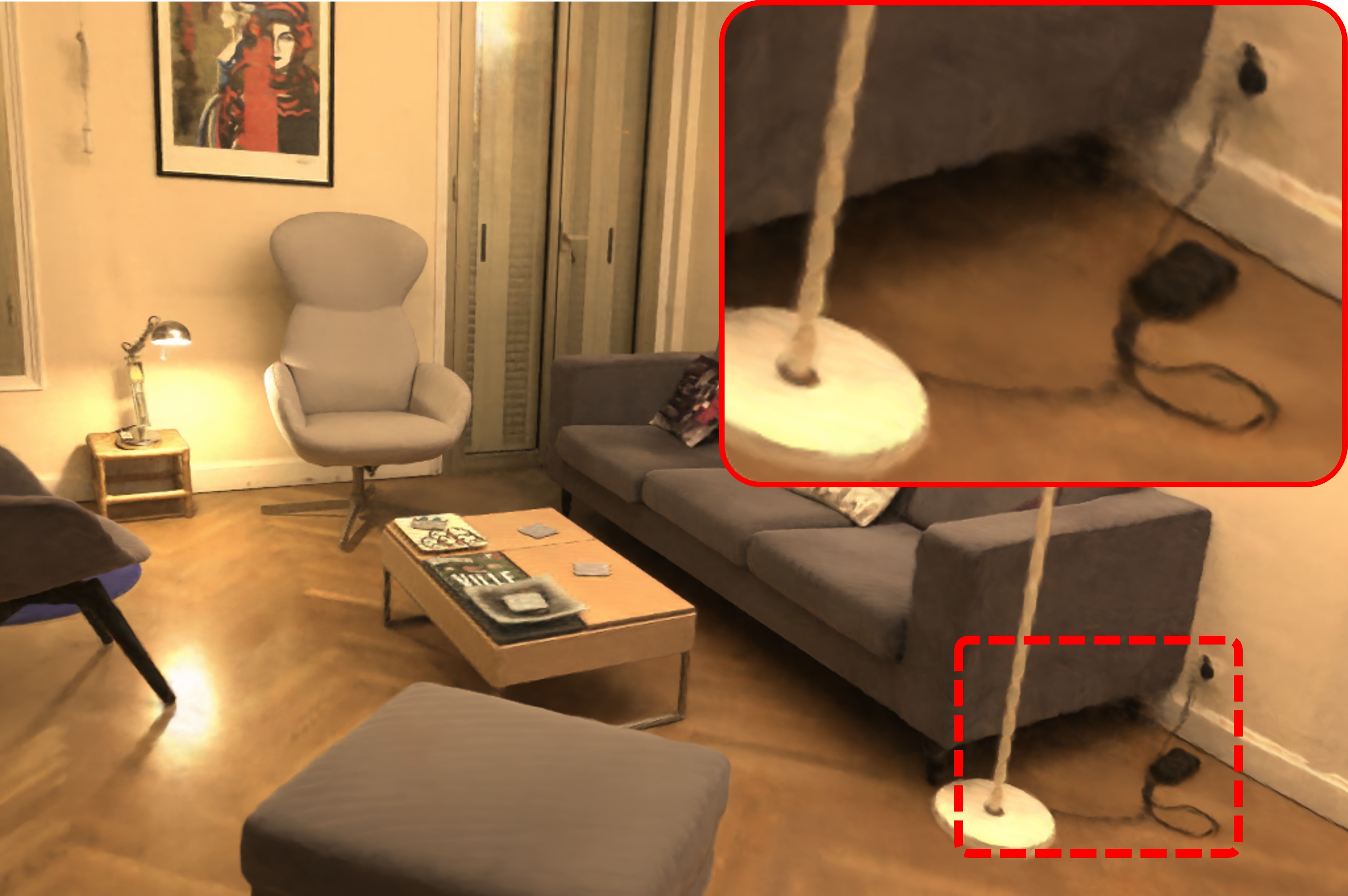}}\,
    \subfloat[360Roam]{\includegraphics[width=\imgw\linewidth]{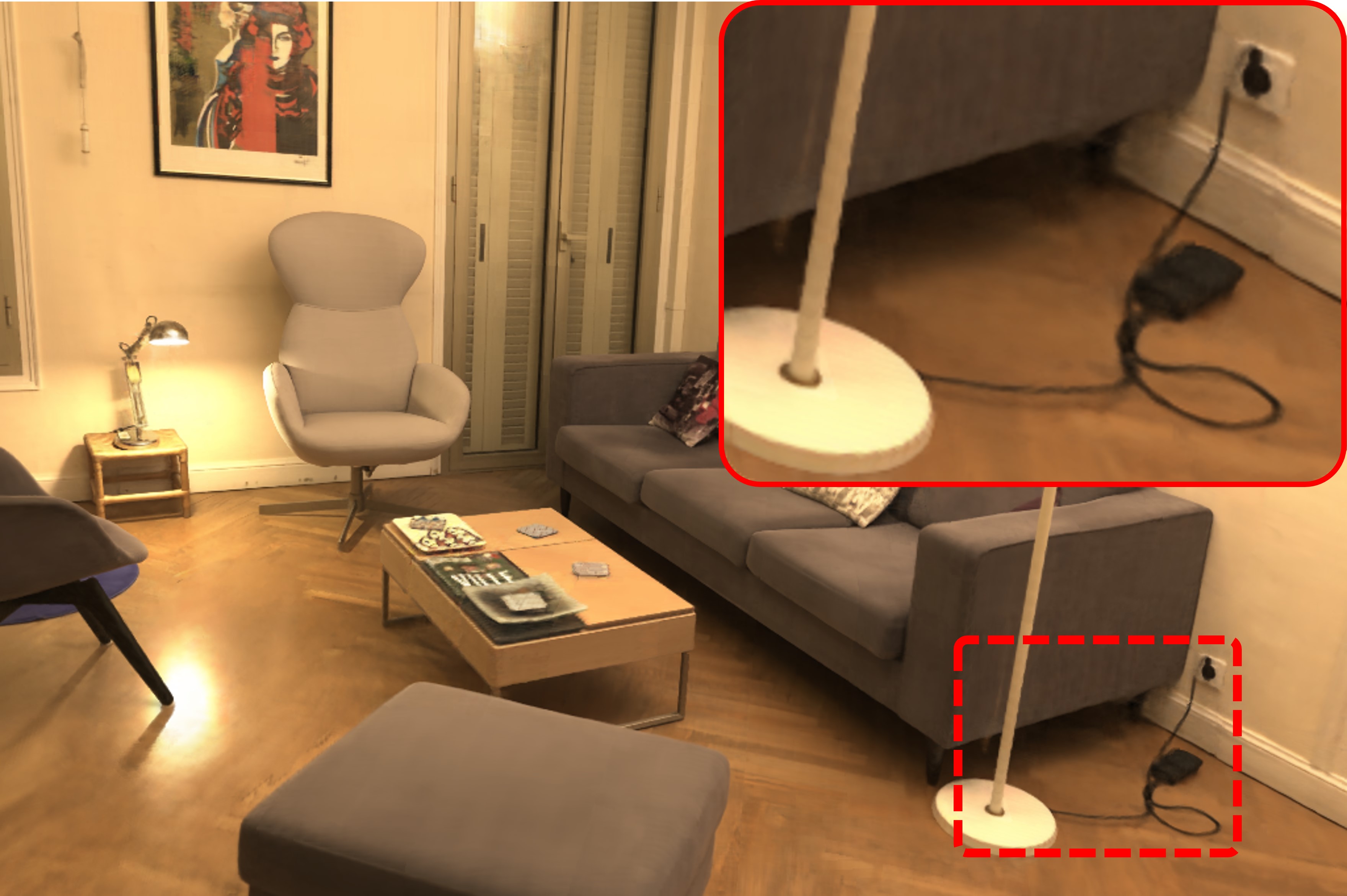}}
    \caption{Visual results on the small-scale scene \textsc{Living Room} of \cite{PMGD21} dataset. Compared to the methods of \cite{PMGD21} and \cite{wu2022snisr} requiring a 3D mesh model as input, our system is more practical and achieves comparable performance.}
    \label{fig:finr}
\end{figure}
\section{Additional Results on Small-scale Scenes} \label{sec:supp_eval_other_scenes}

Apart from evaluating our own 360Roam dataset, we also conducted additional experiments on existing datasets which are small-scale indoor scenarios.
We compared our system with the two extra methods, FINR~\cite{PMGD21} and SNISR~\cite{wu2022snisr}, in the  \textsc{Living room}. This room-level scene contains 322 perspective images collected by \cite{PMGD21}. FINR introduces a neural relighting algorithm using heuristic image-based rendering (IBR) and allows interactive free-viewpoint navigation. SNISR is a recent work supporting scalable neural scene rendering. Both of them take 3D mesh as input for ray sampling while our practical pipeline only requires the posed images. As illustrated in Fig.~\ref{fig:finr}, the rendering quality is competitive. Compared with SNISR, 360Roam can generate the complete structure of tiny objects.

\begin{table*}[t!]
    \centering
    \setlength{\tabcolsep}{1pt} 
	\resizebox{\linewidth}{!}{
	\begin{tabular}{rc cccc cccc|ccccc}
		\toprule
		Scene& & NSVF &NeRF &Mip-NeRF &KiloNeRF & Mip-NeRF360 &Mip-NeRF360* & instant-NGP & TensoRF & 360NeRF &360Roam-100 &360Roam-200 &360Roam-u& 360Roam\\
		\midrule
		\multirow{3}*{\small \textsc{Bar}}&\footnotesize PSNR$\uparrow$& 15.413 &19.049 &19.180 &19.581 & 21.112 &22.082 &15.163 &21.517 & 19.133 &20.770 &21.092 &19.376 &21.676 \\
		&\footnotesize SSIM$\uparrow$&0.438 &0.601 &0.595 &0.632 &0.683  &0.736 &0.488 &0.707 &0.628 &0.636 &0.660 &0.637 &0.711 \\
		&\footnotesize LPIPS$\downarrow$&0.811 &0.409 &0.405 &0.318 &0.324 &0.243 &0.571 &0.294 &0.383 &0.348	&0.311 &0.327 &0.235 \\

		\multirow{3}*{\small \textsc{Base}}&\footnotesize PSNR$\uparrow$& 15.511 &21.255 &21.262 &20.400 &23.178  &24.434 &15.668 &13.256 & 20.802 &22.527 &23.050 &20.837 & 24.093 \\
		&\footnotesize SSIM$\uparrow$&0.427 &0.598 &0.595 &0.596 &0.692 &0.761 &0.472 &0.412 &0.638  &0.615	&0.648 &0.600 &0.725 \\
		&\footnotesize LPIPS$\downarrow$&0.896 &0.400 &0.408 &0.366 &0.309 &0.206 &0.616 &0.791 & 0.359 &0.374	&0.327 &0.373 &0.210 \\
			
		\multirow{3}*{\small \textsc{café}}&\footnotesize PSNR$\uparrow$& 16.435 &20.732 &21.034 &18.324 &23.508 &24.477 &17.051 &13.699 & 20.275 &21.318 &21.752 &19.734& 21.969 \\
		&\footnotesize SSIM$\uparrow$& 0.493 &0.656 &0.659 &0.665 &0.746 &0.798 &0.554 &0.506 & 0.728 &0.646 &0.671 &0.703 & 0.720 \\
		&\footnotesize LPIPS$\downarrow$& 0.803 &0.378 &0.359 &0.308 &0.277 &0.197 &0.521 &0.608 & 0.290 &0.350	&0.311 &0.276 & 0.230 \\

		\multirow{3}*{\small \textsc{canteen}}&\footnotesize PSNR$\uparrow$& 15.933 &19.941 &19.914 &19.891 &21.851 &22.351 &15.851 &17.886 & 22.049 &21.063 &21.444 &19.952 & 21.984 \\
		&\footnotesize SSIM$\uparrow$& 0.491 &0.613 &0.609 &0.628 &0.690 &0.722 &0.506 &0.605 & 0.682 &0.625 &0.640 &0.630 & 0.680 \\
		&\footnotesize LPIPS$\downarrow$& 0.769 &0.427 &0.414 &0.364 &0.356 &0.280 &0.549 &0.498 & 0.378 &0.416	&0.387 &0.364 & 0.303 \\
		
		\multirow{3}*{\small \textsc{center}}&\footnotesize PSNR$\uparrow$& 16.328 &22.439 &22.540 &22.915 &24.841 &25.518 &16.566 &14.391 & 24.462 &24.427 &24.823 &23.220& 25.109 \\
		&\footnotesize SSIM$\uparrow$& 0.554 &0.699 &0.692 &0.731 &0.771 &0.795 &0.590 &0.530 & 0.749 &0.733 &0.748 &0.734 & 0.775 \\
		&\footnotesize LPIPS$\downarrow$& 0.822 &0.358 &0.362 &0.282 &0.280 &0.228 &0.597 &0.783 & 0.298 &0.322	&0.289 &0.279 & 0.226 \\
					
		\multirow{3}*{\small \textsc{corridor}}&\footnotesize PSNR$\uparrow$& 17.777 &25.342 &25.435 &25.825 &28.442 &29.134 &17.722 &14.523 & 27.786 &27.921&28.199 &25.936 & 28.812 \\
		&\footnotesize SSIM$\uparrow$& 0.613 &0.754 &0.753 &0.774 &0.834 &0.863 &0.618 &0.588 & 0.810 &0.791	&0.803 &0.776 & 0.832 \\
		&\footnotesize LPIPS$\downarrow$& 0.607 &0.238 &0.240 &0.204 &0.166 &0.107 &0.425 &0.676 & 0.194 &0.218	&0.197 &0.205 & 0.145 \\
		
		\multirow{3}*{\small \textsc{innovation}}&\footnotesize PSNR$\uparrow$& 16.681 &22.482 &22.558 &21.700 &25.433 &26.398 &17.121 &12.716 & 23.822 &24.729 &25.197 &23.024 & 26.191 \\
		&\footnotesize SSIM$\uparrow$& 0.476 &0.667 &0.660 &0.679 &0.754 &0.796 &0.519 &0.428 & 0.718 &0.694 &0.715 &0.706 & 0.771 \\
		&\footnotesize LPIPS$\downarrow$& 0.800 &0.324 &0.322 &0.265 &0.253 &0.185 &0.543 &0.799 & 0.300 &0.305	&0.276 &0.243 & 0.187 \\
			
		\multirow{3}*{\small \textsc{lab}}&\footnotesize PSNR$\uparrow$& 21.331 &24.135 &24.018 &21.072 &25.830  &27.005 &19.075  &15.542 & 21.264 &25.518 &26.269 &23.680 & 27.667 \\
		&\footnotesize SSIM$\uparrow$& 0.651 &0.763 &0.767 &0.797 &0.833 &0.870 &0.651 &0.593 & 0.801 &0.773	&0.802 &0.817 & 0.855 \\
		&\footnotesize LPIPS$\downarrow$& 0.504 &0.239 &0.231 &0.185 &0.192 &0.129 &0.414 &0.642 & 0.224 &0.243	&0.198 &0.144 & 0.116 \\
		
		\multirow{3}*{\small \textsc{library}}&\footnotesize PSNR$\uparrow$& 16.794 &23.909 &23.872 &21.301 &25.971 &26.540 &18.450 &13.409 & 25.134 &25.418 &25.745 &23.752 &26.127 \\
		&\footnotesize SSIM$\uparrow$& 0.497 &0.656 &0.651 &0.652 &0.714 &0.742 &0.519 &0.468 & 0.702 &0.668	&0.685 &0.683 & 0.722 \\
		&\footnotesize LPIPS$\downarrow$& 0.832 &0.326 &0.330 &0.317 &0.288 &0.232 &0.558 &0.836 & 0.289 &0.322	&0.294 &0.276 &0.225 \\

		\multirow{3}*{\small \textsc{office}}&\footnotesize PSNR$\uparrow$& 18.293 &25.149 &25.491 &23.251 &25.619 &26.707 &17.521 &13.409  & 25.749 &25.486 &26.002 &24.637 &26.977 \\
		&\footnotesize SSIM$\uparrow$& 0.525 &0.714 &0.716 &0.744 &0.763 &0.802 &0.563 &0.468 & 0.741 &0.735	&0.758 &0.755 &0.811 \\
		&\footnotesize LPIPS$\downarrow$& 0.741 &0.290 &0.286 &0.219 &0.245 &0.174 &0.524 &0.836 & 0.255 &0.255	&0.218	 &0.189 &0.145 \\

		\bottomrule
	\end{tabular}
	}
	\caption{Complete quantitative comparison of novel view synthesis among different scenes. * indicates Mip-NeRF360 with 1024-channel MLP network.}
	\label{tab:supp_complete_quan}
\end{table*}

\section{Complete Quantitative Results} \label{sec:supp_quan}
In this section, we report detailed quantitative evaluation statistics for all 10 evaluated scenes in Table~\ref{tab:supp_complete_quan}. Compared with baselines and 360Roam variants, our proposed 360Roam system generally is more reliable in view synthesis of real-world large scenes.

\section{More Qualitative Results} \label{sec:supp_qual}

\subsection{Ablation Study Visual Results} \label{sec:supp_ablation_results}

\begin{figure*}[t]
    \def\imgw{0.16}
	\centering
	
	\includegraphics[width=\imgw\linewidth]{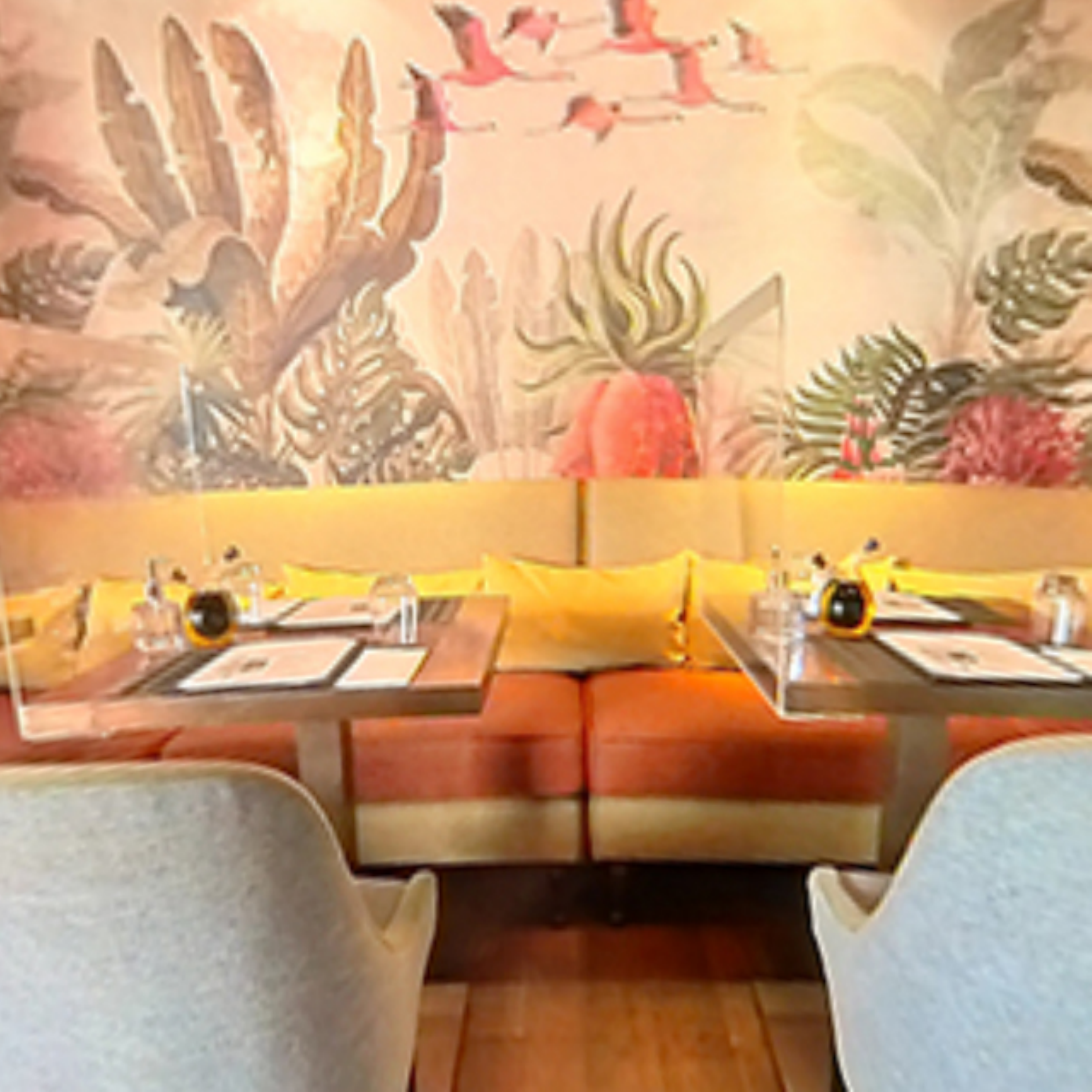}
	\includegraphics[width=\imgw\linewidth]{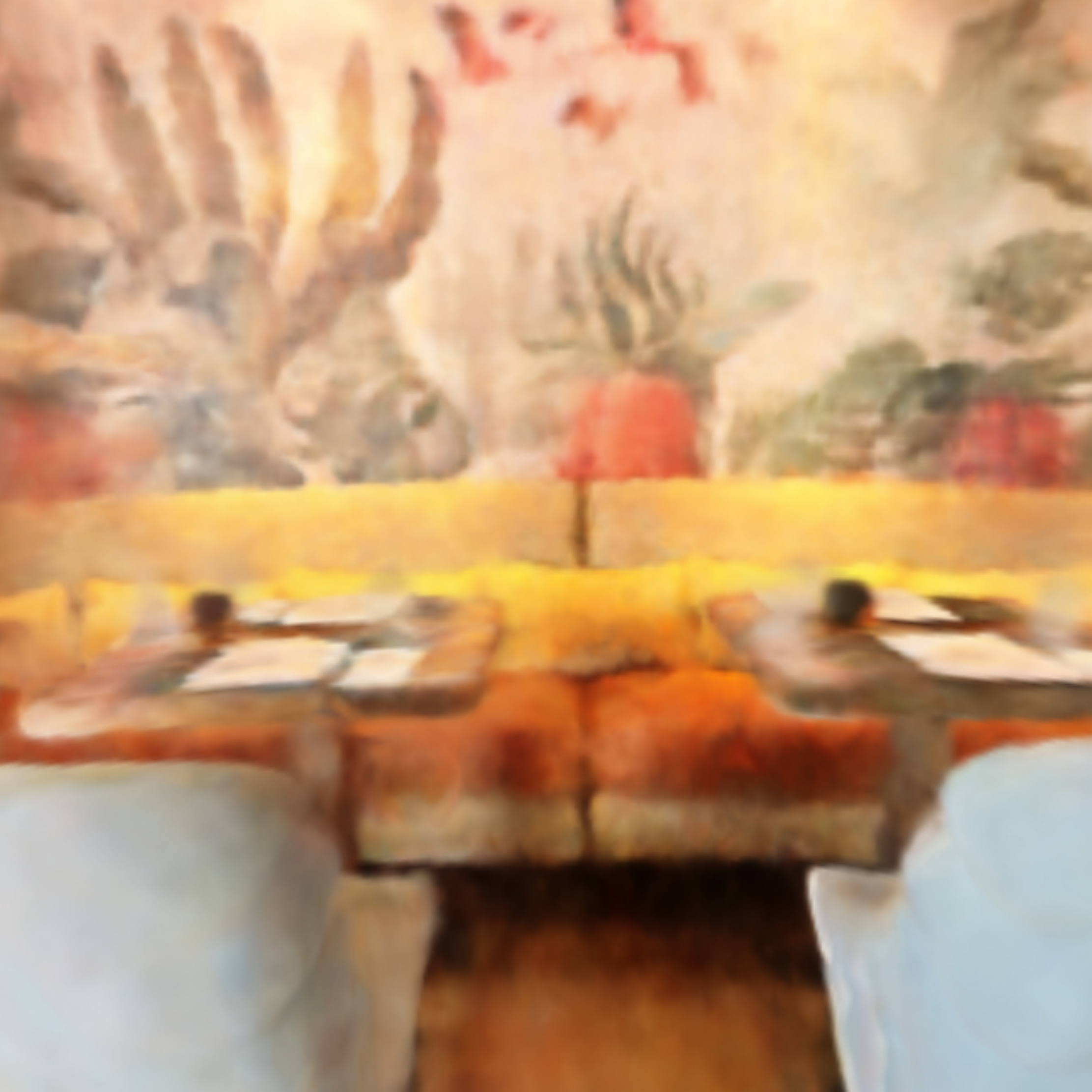}
	\includegraphics[width=\imgw\linewidth]{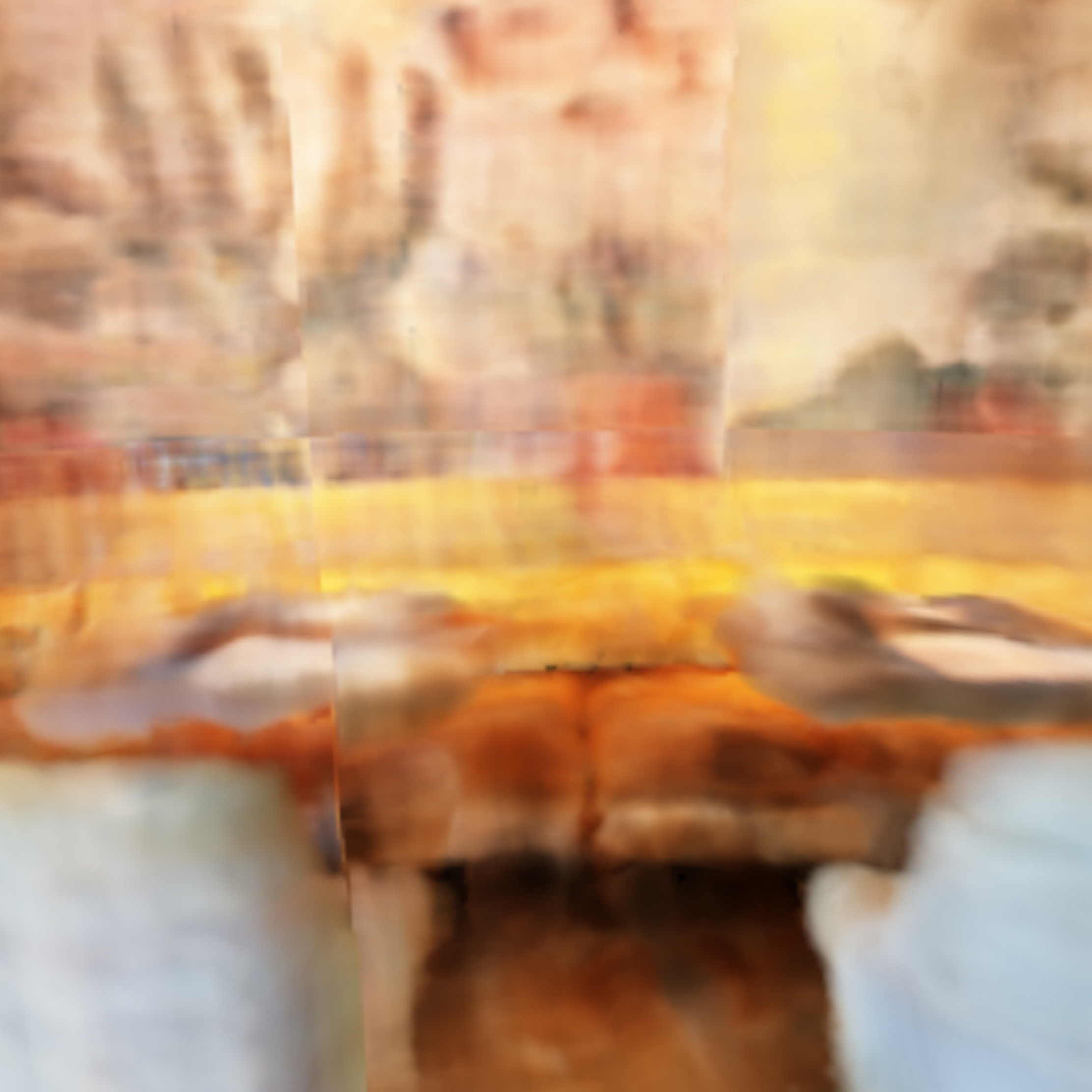}
	\includegraphics[width=\imgw\linewidth]{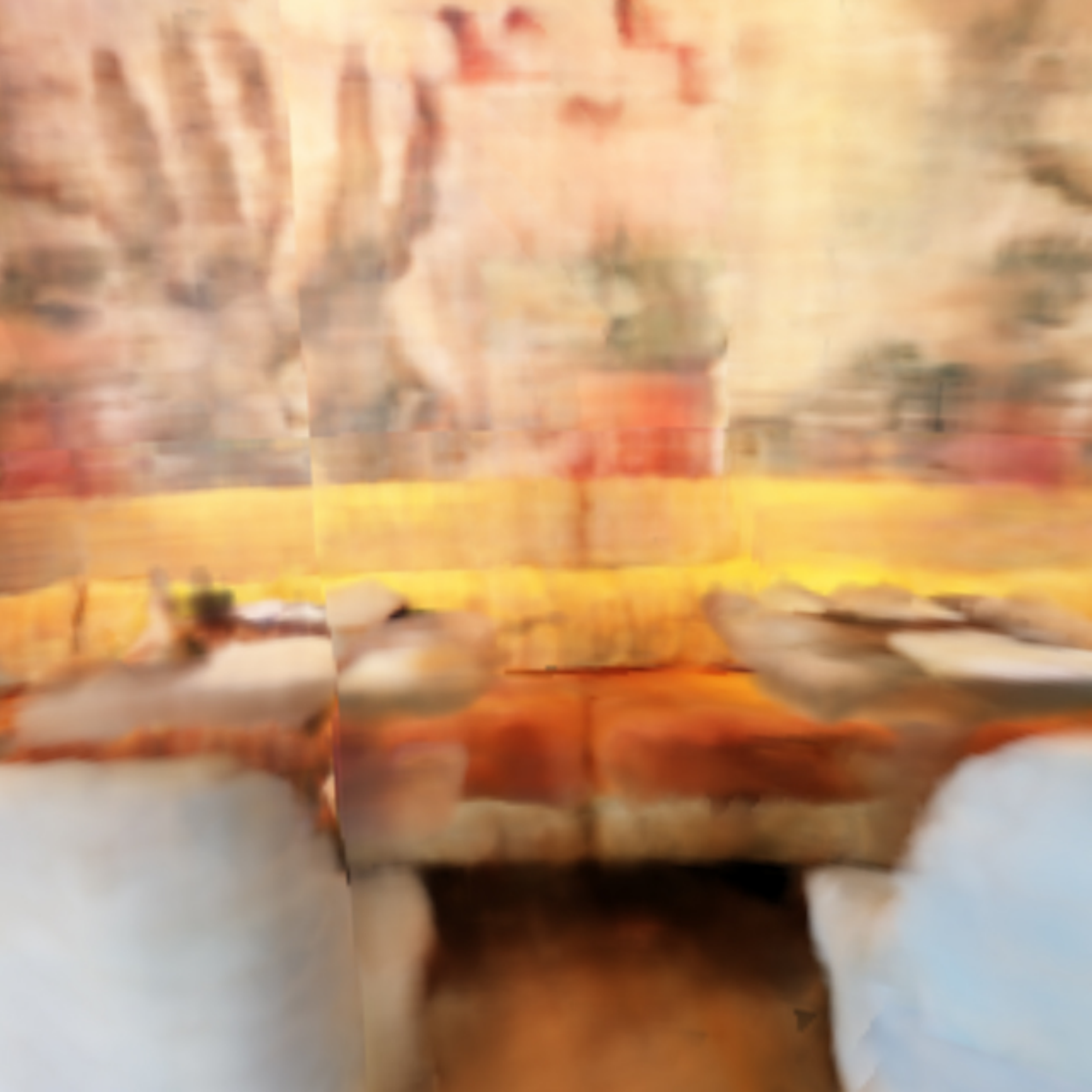}
	\includegraphics[width=\imgw\linewidth]{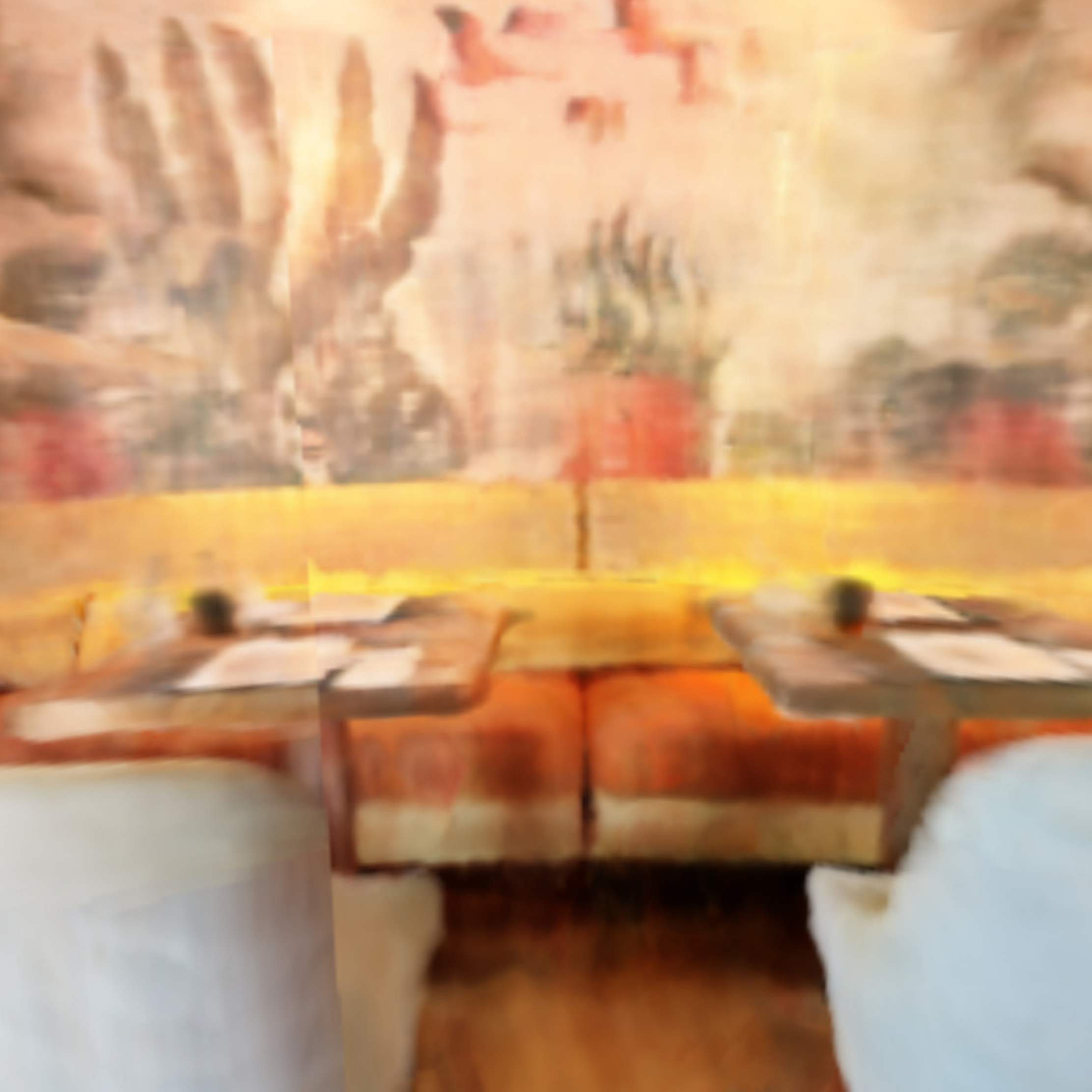}
	\includegraphics[width=\imgw\linewidth]{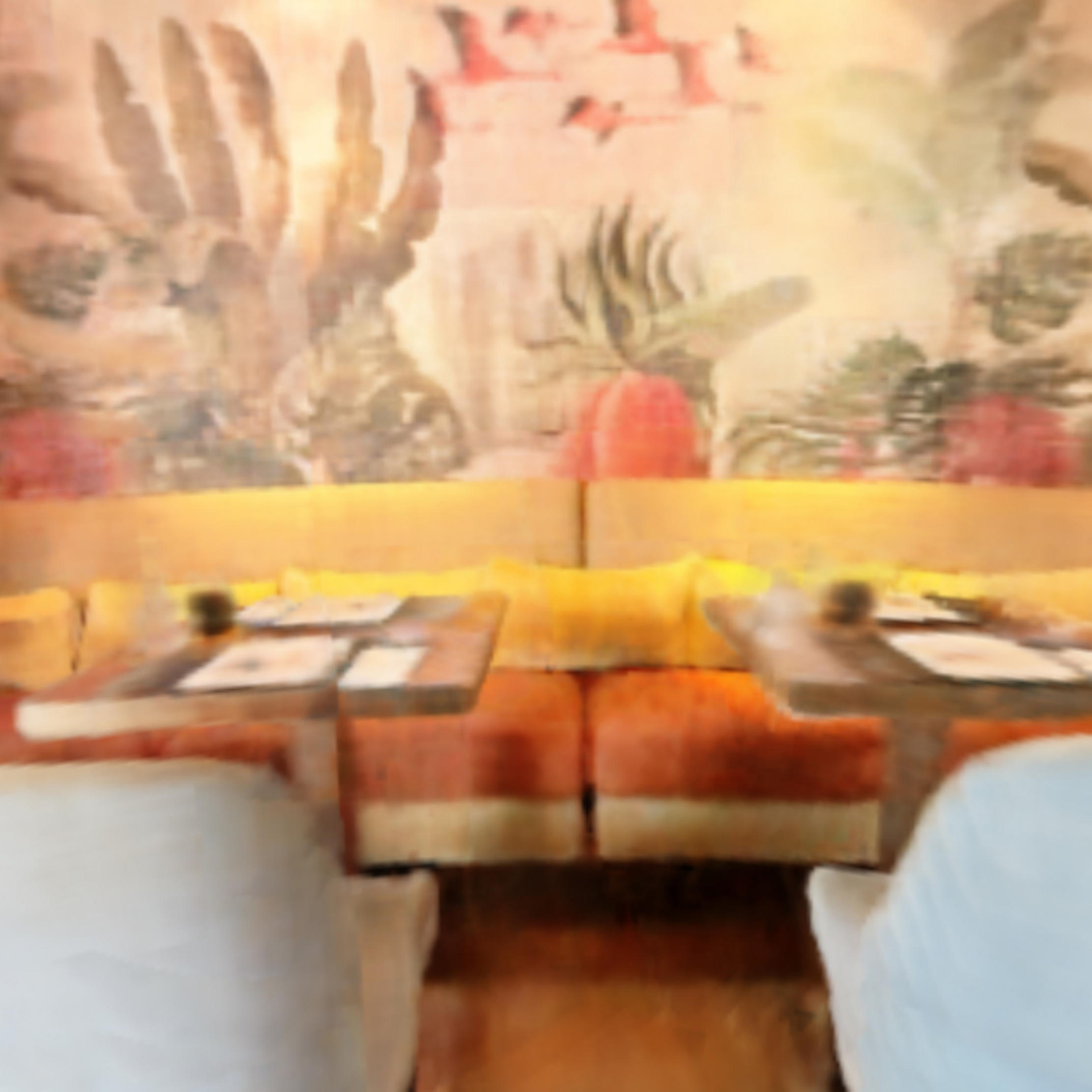}
	
    \includegraphics[width=\imgw\linewidth]{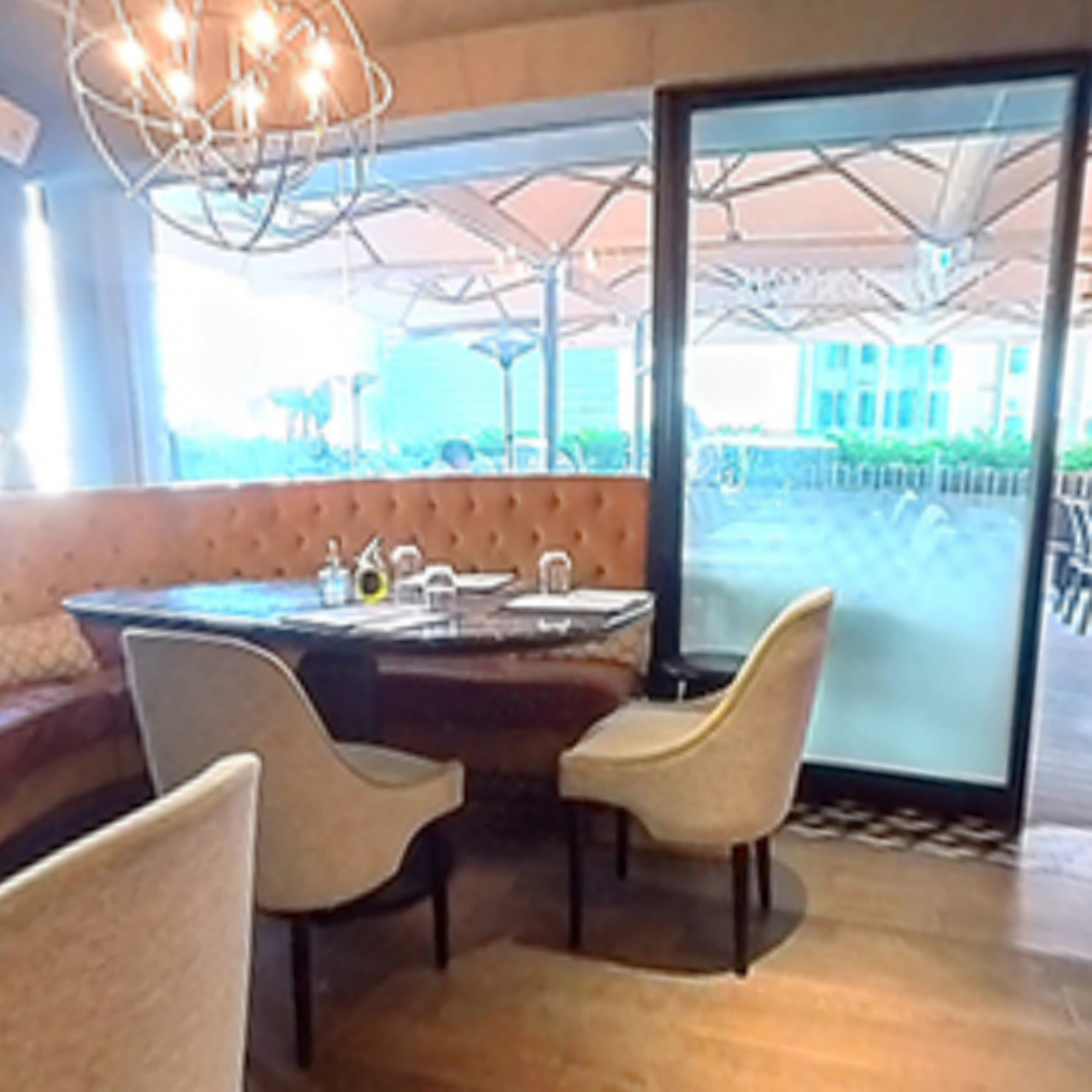}
	\includegraphics[width=\imgw\linewidth]{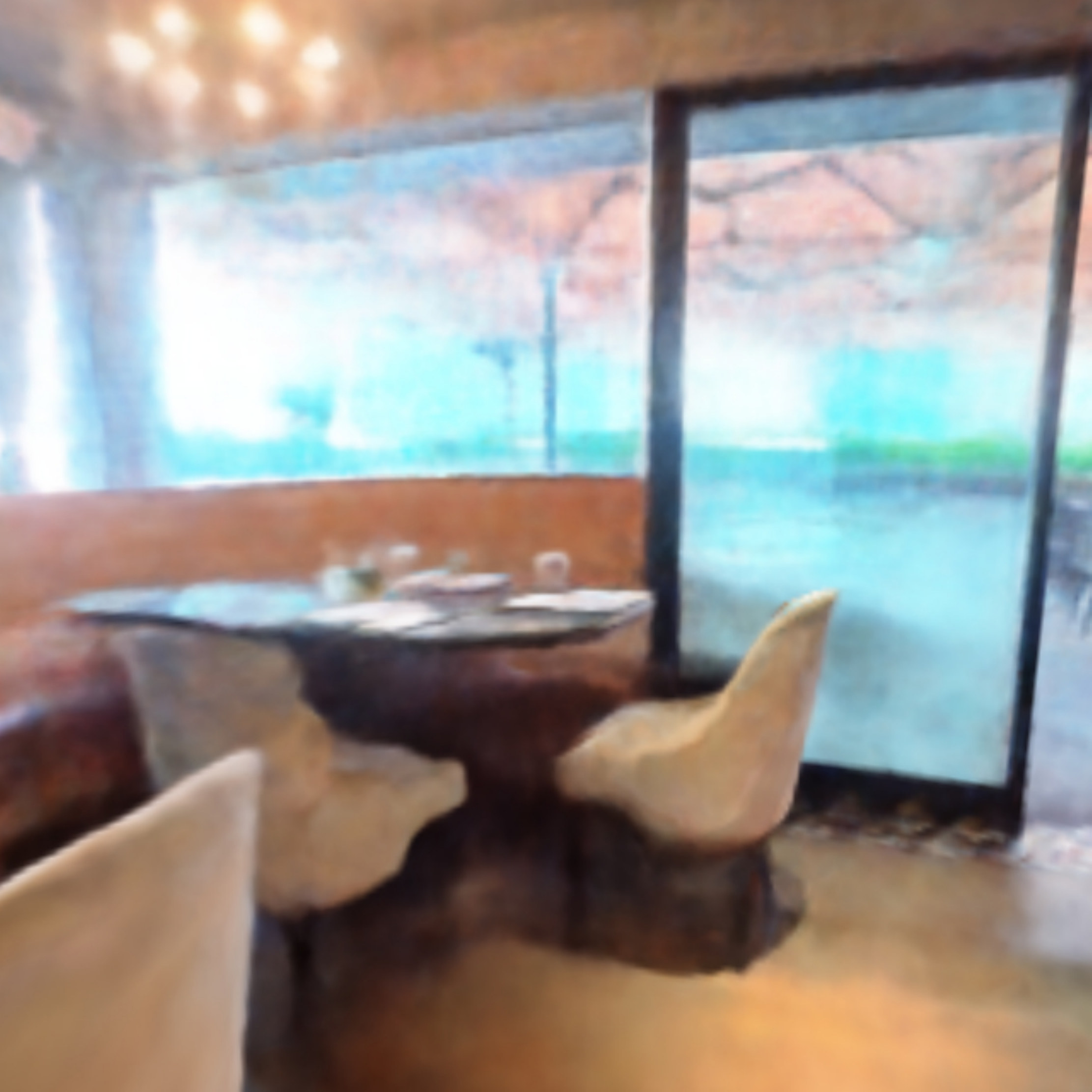}
	\includegraphics[width=\imgw\linewidth]{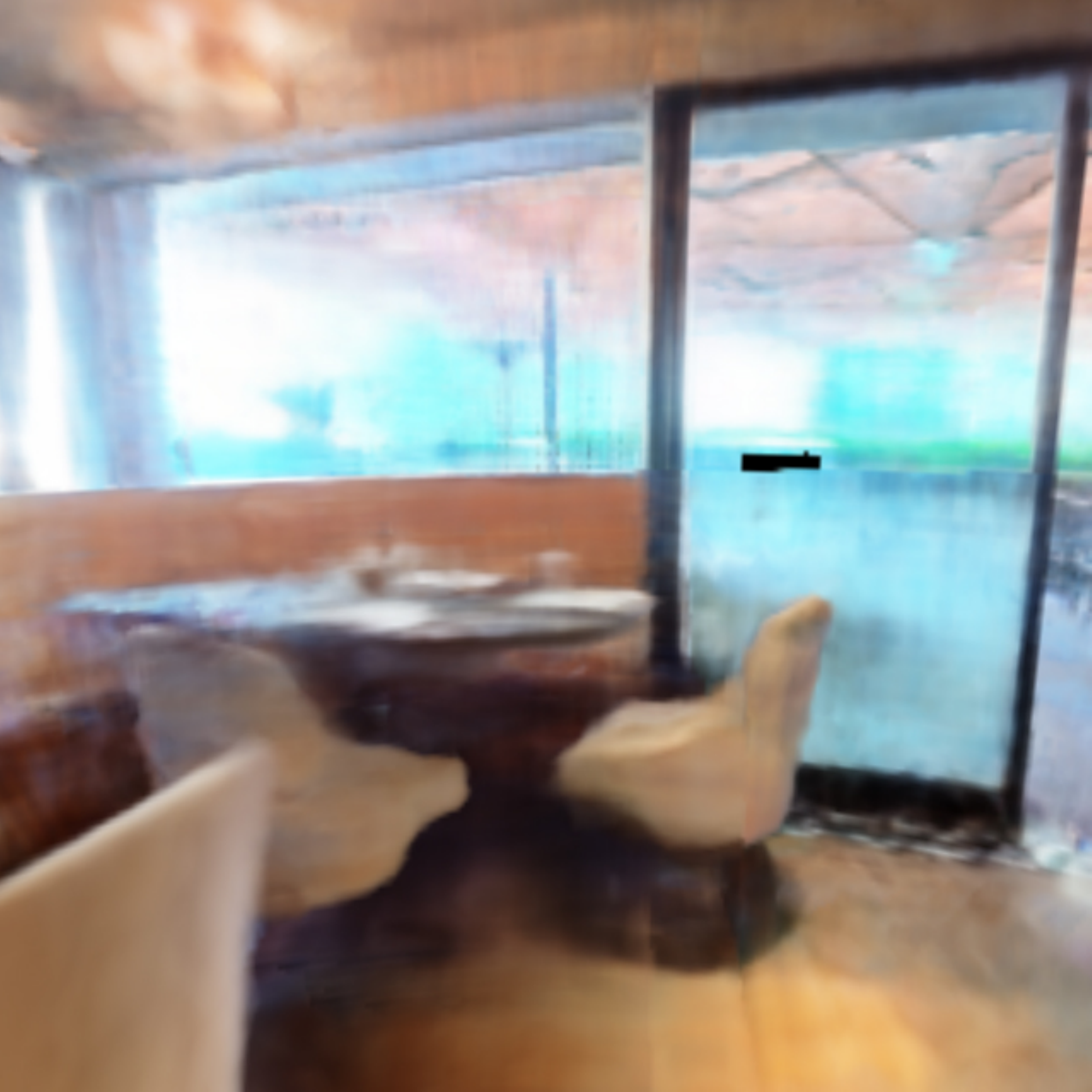}
	\includegraphics[width=\imgw\linewidth]{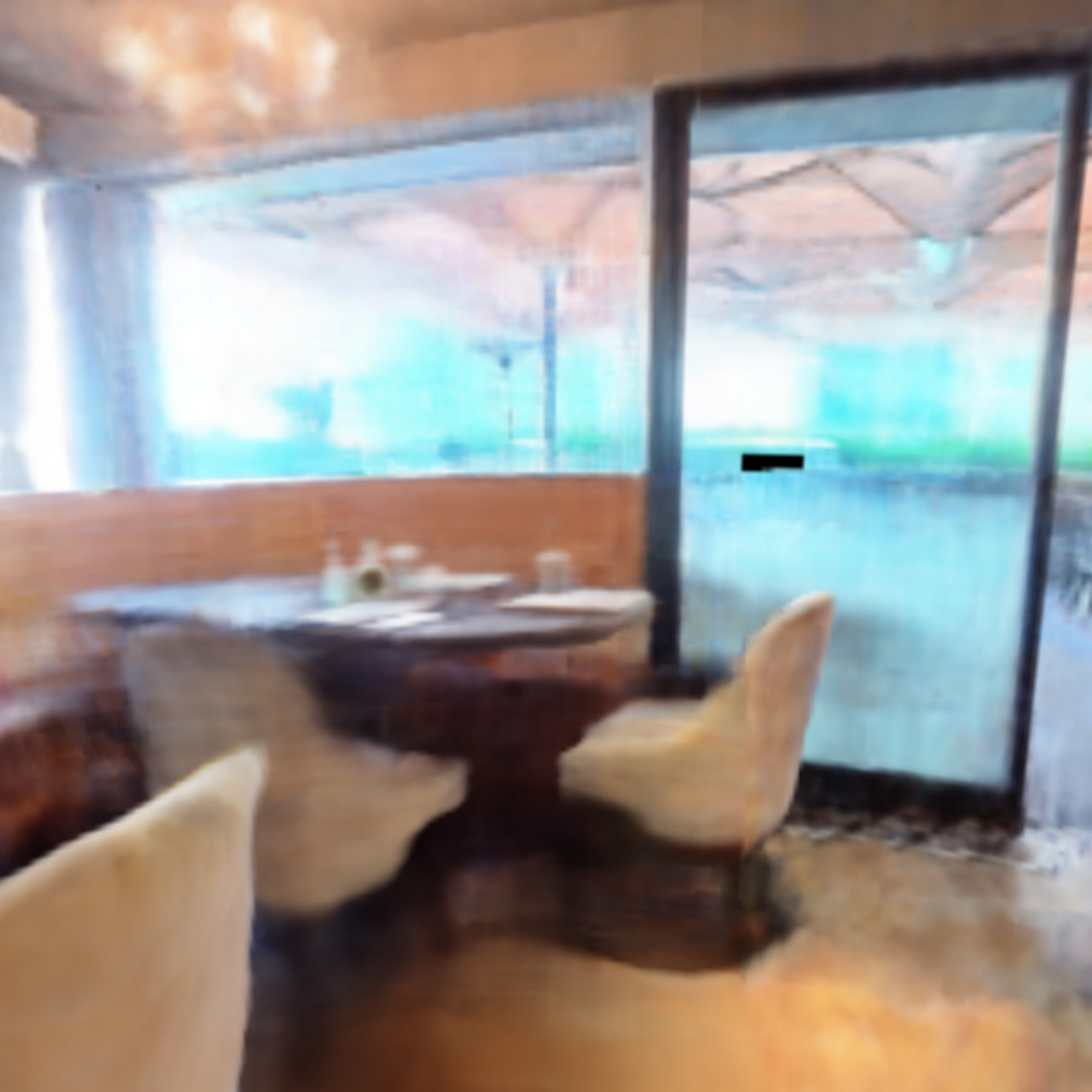}
	\includegraphics[width=\imgw\linewidth]{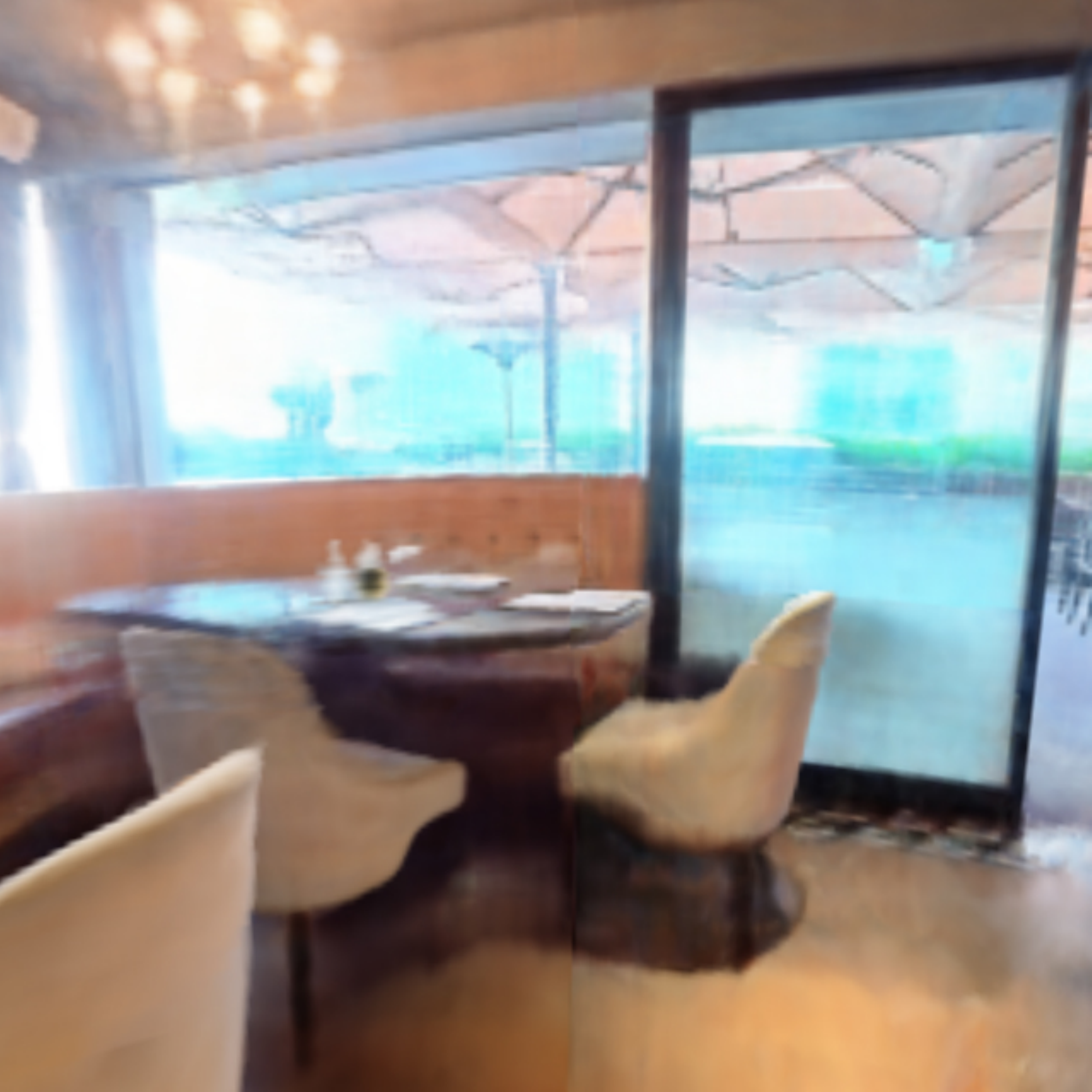}
	\includegraphics[width=\imgw\linewidth]{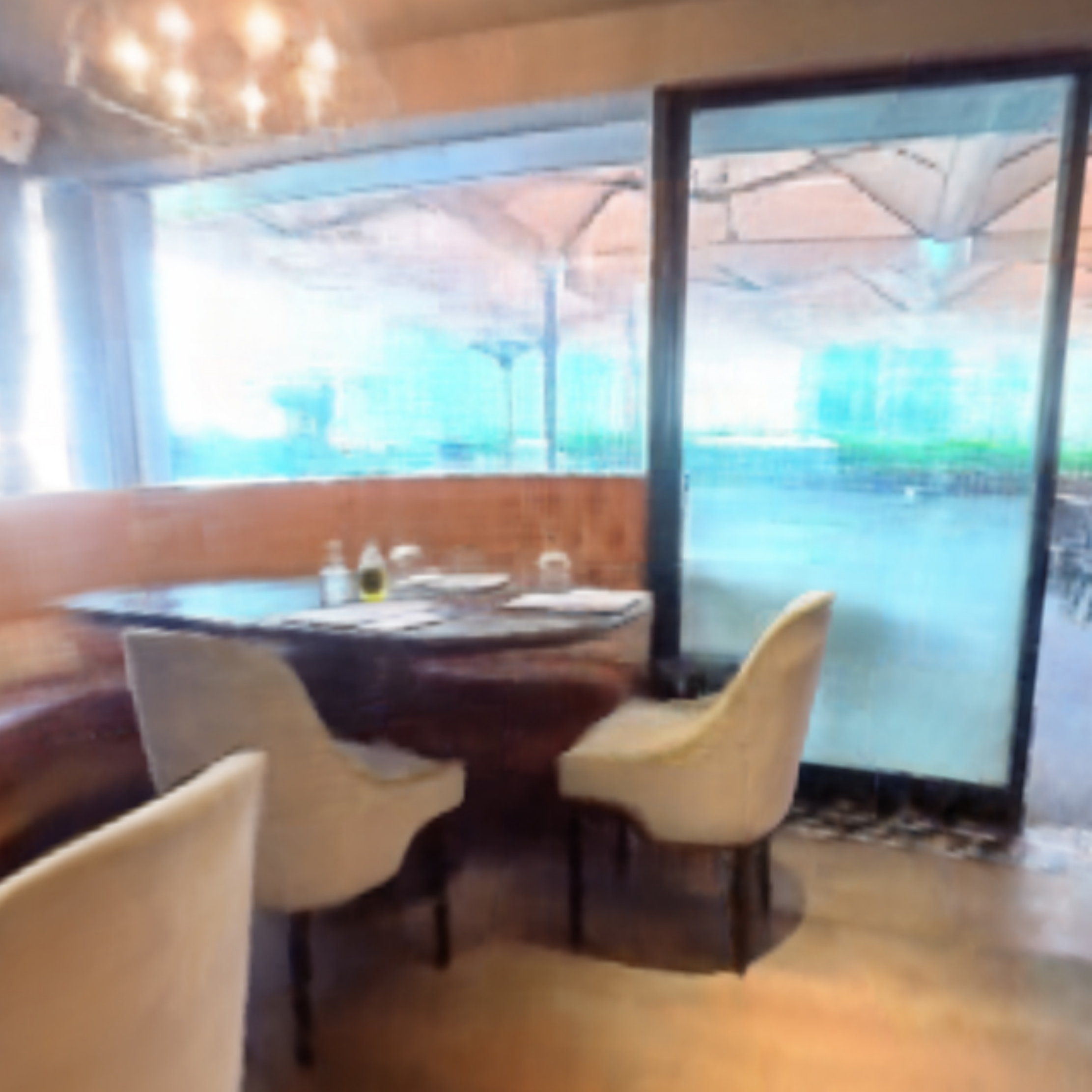}

	\begin{tikzpicture}
        \def\imgh{0.6}
        \def\citationH{0.3}
		\node[inner sep=0pt] () at (-0.42\linewidth,\imgh){Ground Truth};
		\node[inner sep=0pt] () at (-0.25\linewidth,\imgh){360NeRF};
		\node[inner sep=0pt] () at (-0.08\linewidth,\imgh){360Roam-100};
		\node[inner sep=0pt] () at (0.08\linewidth,\imgh){360Roam-200};
		\node[inner sep=0pt] () at (0.25\linewidth,\imgh){360Roam-u};
		\node[inner sep=0pt] () at (0.42\linewidth,\imgh){360Roam};
		
    \end{tikzpicture}
    \caption{Novel view synthesis results of our system variants. Zoom in for a better view.} 
    \label{fig:supp_ablation}
\end{figure*}

In the main paper, we quantitatively evaluate the performance of 360Roam variants, and Table~\ref{tab:supp_tiny_network} reports detailed evaluation numbers. Fig. ~\ref{fig:supp_ablation} additionally supplements visual comparison. Novel view images of the representative 360Roam always have higher fidelity of the texture at any distance. For example, the complex wall texture in the first row, and the tableware in the second row are generated with clear and recognizable details. 
The visual comparisons also intuitively demonstrate that the rendering images get clearer and the more high-frequency details are recovered as the number of tiny networks used increases.



More novel view comparisons are composed in the supplementary video.

\begin{table*}[t!]
    \centering
    \begin{tabular}{rcccccccccc}
        \toprule
		Scene&& 100 &200 &300 &400 & 500 &1024 &2048 & 4096 & 8192\\
		\midrule
		\multirow{3}*{\textsc{Bar}}&\footnotesize   PSNR$\uparrow$& 20.770&	21.092&	21.185&	21.343&	21.409&	21.572&	21.676&	21.741&	21.749 \\
		&\footnotesize SSIM$\uparrow$&0.636	&0.660	&0.672	&0.681	&0.686	&0.699	&0.711	&0.718	&0.723	  \\  
		&\footnotesize LPIPS$\downarrow$&0.348	&0.311	&0.294	&0.281	&0.272	&0.251	&0.235	&0.223	&0.216	  \\
		
		\multirow{3}*{\textsc{Base}}&\footnotesize  PSNR$\uparrow$&22.527	&23.050	&23.323	&23.499	&23.565	&23.874	&24.093	&24.164	&24.190  \\  
		&\footnotesize SSIM$\uparrow$&0.615	&0.648	&0.666	&0.678	&0.685	&0.708	&0.725	&0.734	&0.741 \\  
		&\footnotesize LPIPS$\downarrow$&0.374	&0.327	&0.299	&0.282	&0.270	&0.234	&0.210	&0.194	&0.182\\
		
		\multirow{3}*{\textsc{Café}}&\footnotesize  PSNR$\uparrow$&21.318	&21.752	&21.901	&21.854	&21.897	&22.158	&21.969	&22.184	&22.353 \\  
		&\footnotesize SSIM$\uparrow$&0.646	&0.671	&0.686	&0.690	&0.696	&0.713	&0.720	&0.730	&0.736 \\  
		&\footnotesize LPIPS$\downarrow$&0.350	&0.311	&0.290	&0.281	&0.271	&0.245	&0.230	&0.215	&0.204 \\
	
		\multirow{3}*{\textsc{Canteen}}&\footnotesize  PSNR$\uparrow$&21.063	&21.444	&21.620	&21.637	&21.527	&21.828	&21.984	&22.045	&22.070 \\
		&\footnotesize SSIM$\uparrow$  &0.625	&0.640	&0.651	&0.654	&0.657	&0.669	&0.680	&0.685	&0.689 \\
		&\footnotesize LPIPS$\downarrow$&0.416	&0.387	&0.363	&0.355	&0.348	&0.322	&0.303	&0.285	&0.273  \\

		\multirow{3}*{\textsc{Center}}&\footnotesize PSNR$\uparrow$&24.427	&24.823	&24.952	&24.914	&25.023	&25.084	&25.109	&25.183	&25.247  \\  
		&\footnotesize SSIM$\uparrow$ &0.733	&0.748	&0.756	&0.758	&0.762	&0.770	&0.775	&0.778	&0.781 \\
		&\footnotesize LPIPS$\downarrow$&0.322	&0.289	&0.272	&0.267	&0.258	&0.238	&0.226	&0.216	&0.211 \\

		\multirow{3}*{\textsc{Corridor}}&\footnotesize PSNR$\uparrow$ &27.921	&28.199	&28.374	&28.489	&28.555	&28.733	&28.812	&28.808	&28.843\\
		&\footnotesize SSIM$\uparrow$&0.791	&0.803	&0.810	&0.814	&0.818	&0.826	&0.832	&0.835	&0.837  \\
		&\footnotesize LPIPS$\downarrow$&0.218	&0.197	&0.184	&0.179	&0.172	&0.156	&0.145	&0.138	&0.131 \\

		\multirow{3}*{\textsc{Innovation}}&\footnotesize PSNR$\uparrow$&24.729	&25.197	&25.474	&25.634	&25.675	&26.020	&26.191	&26.290	&26.301  \\
		&\footnotesize SSIM$\uparrow$&0.694	&0.715	&0.729	&0.738	&0.743	&0.761	&0.771	&0.779	&0.785  \\
		&\footnotesize LPIPS$\downarrow$&0.305	&0.276	&0.255	&0.241	&0.234	&0.207	&0.187	&0.174	&0.164 \\

		\multirow{3}*{\textsc{Lab}}&\footnotesize PSNR$\uparrow$&25.518	&26.269	&26.654	&26.973	&27.125	&27.429	&27.667	&28.012	&28.118  \\  
		&\footnotesize SSIM$\uparrow$&0.773	&0.802	&0.816	&0.825	&0.831	&0.845	&0.855	&0.862	&0.865 \\
		&\footnotesize LPIPS$\downarrow$&0.243	&0.198	&0.176	&0.161	&0.152	&0.132	&0.116	&0.106	&0.100\\

		\multirow{3}*{\textsc{Library}}&\footnotesize PSNR$\uparrow$&25.418	&25.745	&25.961	&25.961	&26.097	&26.161	&26.127	&26.201	&26.199   \\  
		&\footnotesize SSIM$\uparrow$&0.668	&0.685	&0.697	&0.700	&0.706	&0.715	&0.722	&0.727	&0.728  \\
		&\footnotesize LPIPS$\downarrow$&0.322	&0.294	&0.274	&0.267	&0.258	&0.238	&0.225	&0.217	&0.210  \\

		\multirow{3}*{\textsc{Office}}&\footnotesize PSNR$\uparrow$  &25.486	&26.002	&26.261	&26.463	&26.414	&26.767	&26.977	&27.082	&27.090  \\
		&\footnotesize SSIM$\uparrow$&0.735	&0.758	&0.772	&0.781	&0.785	&0.801	&0.811	&0.817	&0.820  \\
		&\footnotesize LPIPS$\downarrow$&0.255	&0.218	&0.199	&0.187	&0.180	&0.159	&0.145	&0.136	&0.130  \\

		\bottomrule
    \end{tabular}
    \caption{Quantitative results on each scene using different numbers of the tiny networks. }
    \label{tab:supp_tiny_network}
\end{table*}

\begin{figure}[t]
    \def\imgw{0.45}
    \centering
    \subfloat[\small COLMAP]{\label{fig:colmap_inno}\includegraphics[width=\imgw\linewidth]{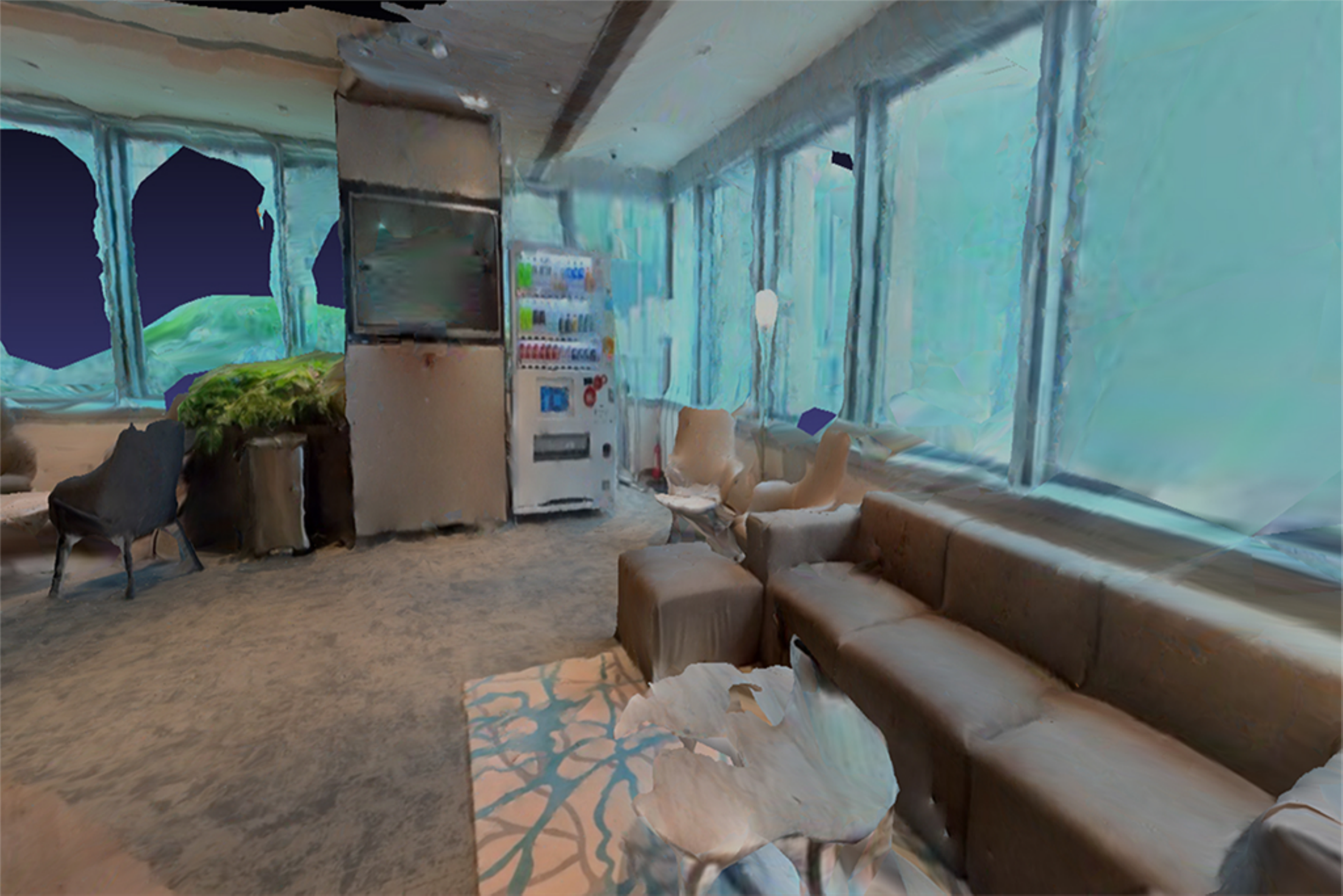}}\quad
    \subfloat[\small Ours]{\label{fig:our_inno}\includegraphics[width=\imgw\linewidth]{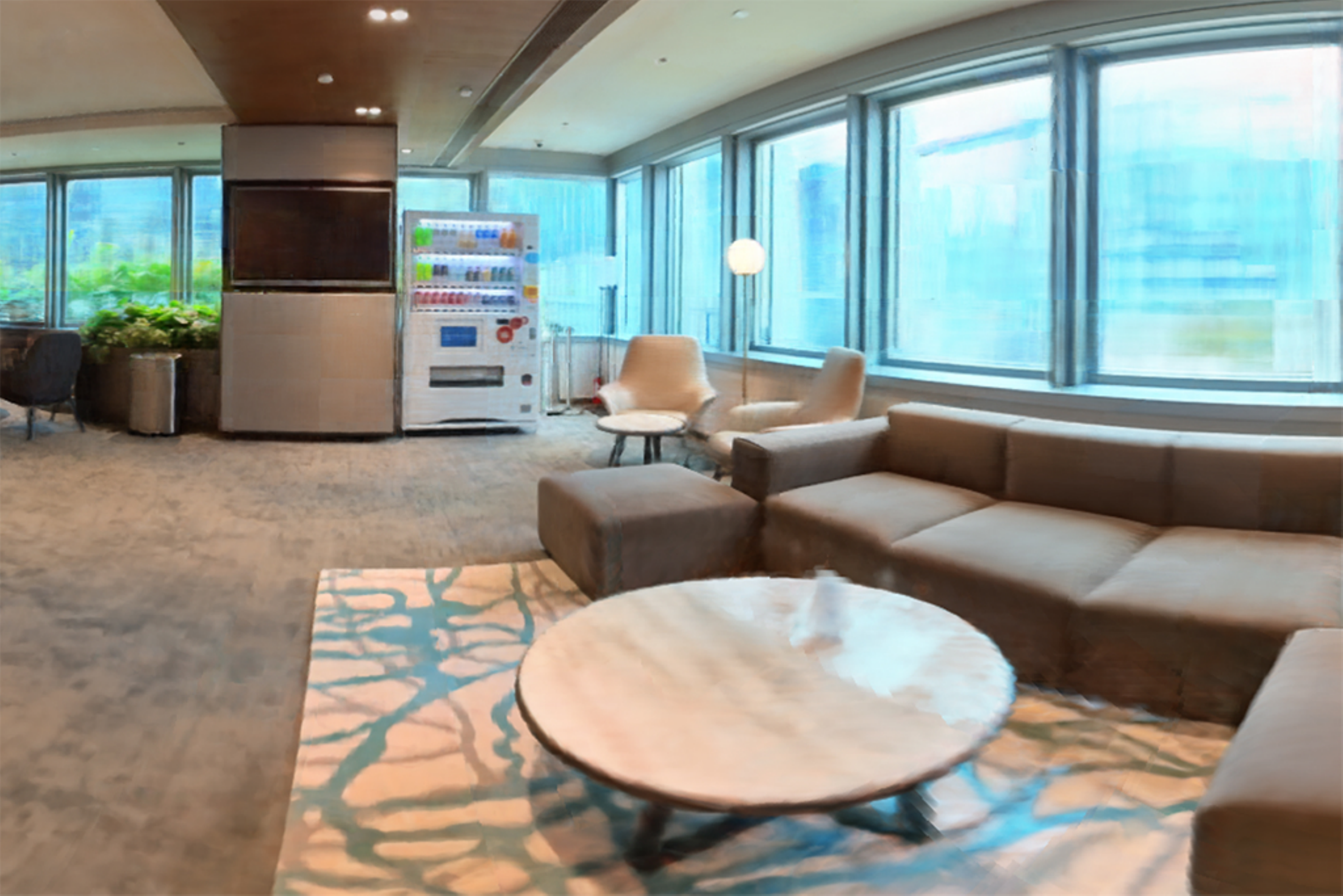}}
    \vspace{-2mm}
    \caption{Comparison with 3D reconstruction and texturing. Left: a perspective result of reconstructed textured mesh. Right: corresponding rendered view from our system. The reconstructed model has obvious geometry and texture artifacts and leads to non-photorealism, while ours is much more photorealistic and suitable for roaming.}
    \label{fig:compare_colmap}
\end{figure}
\begin{figure}[t]
    \def\imgw{0.43}
    \centering
    \subfloat[\small Ground truth]{\includegraphics[width=\imgw\linewidth]{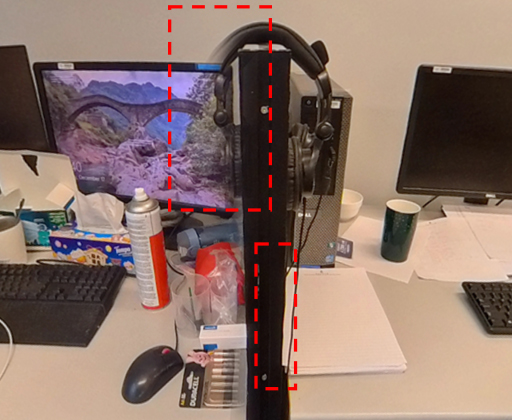}}\quad
    \subfloat[\small Synthetic image]{\includegraphics[width=\imgw\linewidth]{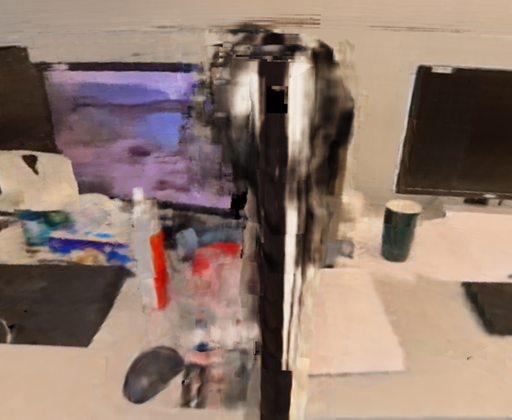}}
    \vskip -0.1cm
    \caption{Stitching artifacts of $360^\circ$ images affect rendering.}
    \label{fig:cameraDefect}
\end{figure}




\section{More Discussions and Limitations} \label{sec:supp_discuss}
We demonstrated how 360Roam effectively extends the conventional NeRF into the geometry-aware 360NeRF to generate high-quality $360^\circ$ images for real-time indoor roaming. We also conduct an additional comparison with a typical method which uses 3D reconstruction~\cite{schoenberger2016sfm, schoenberger2016mvs} and texture mapping~\cite{waechter2014let} techniques to recover meshes for real scene roaming. As COLMAP does not support $360^\circ$ images, we crop the training panoramas into cube maps and then use the same camera poses to run COLMAP. An illustrative example is shown in Fig.~\ref{fig:compare_colmap}. Our approach produces real-time photorealistic renderings and has a dramatic advantage over conventional 3D reconstruction which shows obvious geometry and texture artifacts.  
Our system also inherits some limitations of NeRF and panorama scene understanding. We list the most important issues and potential directions to further improve the system. 

\textit{Camera model.} We use an ideal spherical camera model to describe the projection of the $360^\circ$ camera. However, a consumer-grade $360^\circ$ camera outputs $360^\circ$ images by optimizing the stitching of two fish-eye images. Due to a manufacturing deficiency and the imperfection of factory camera calibration, original images sometimes have obvious stitching artifacts which will affect the rendering quality, as Fig.~\ref{fig:cameraDefect} shows. 
Although we can exploit a professional $360^\circ$ camera for capturing, it is necessary to take distortion into account and optimize camera intrinsic parameters during training.



\textit{Few-shot sampling.}
Most NeRF-based methods require dense scene sampling to collect as much environmental information as possible. Also, most existing datasets for novel view synthesis tasks usually contain tens or even hundreds of images representing an object or small scenes.
However, our dataset only uses a few hundred of images to represent a large-scale scene. Although panoramas having a wider field of view are beneficial for representing large scenes, the rendering quality can still be poor in under-sampled areas. How to provide a more realistic view-dependent effect with few-shot training will be an interesting future direction.

\textit{Region of interest.}
Our current 360Roam focuses on real-time scene navigation and therefore suffers from performance degradation in object-centric scenarios. One promising future direction is to bring in scene understanding such as 3D semantic segmentation to strategically slim the important objects of interest while maintaining good computational performance.

\end{document}